\documentclass[letterpaper, 10 pt, journal, twoside]{IEEEtran}
\usepackage{microtype}
\usepackage{graphicx}
\usepackage{booktabs}
\usepackage{amsmath}
\usepackage{amssymb}
\usepackage{mathtools}
\usepackage{url}
\usepackage{algorithm}
\usepackage{algpseudocode}
\usepackage{listings}
\usepackage[dvipsnames, rgb, table]{xcolor}
\usepackage{tikz}
\usepackage{bm}
\usepackage{pgfplots}
\usepackage{tocloft}   
\usepackage{etoc}
\pgfplotsset{compat=1.18}
\usepackage{pgfplotstable}

\usepgfplotslibrary{fillbetween}
\pgfdeclarelayer{background}
\pgfdeclarelayer{back background}
\pgfsetlayers{back background, background, main}
\usetikzlibrary{calc}
\usetikzlibrary{backgrounds}
\usetikzlibrary{fit}
\usetikzlibrary{bayesnet}
\usetikzlibrary{arrows}
\usetikzlibrary{arrows.meta}
\usetikzlibrary{positioning}
\usetikzlibrary{shapes}
\usetikzlibrary{shapes.geometric}
\usetikzlibrary{shadows}
\usetikzlibrary{patterns}
\usetikzlibrary{fadings}
\usetikzlibrary{decorations.pathmorphing}
\usetikzlibrary{decorations.pathreplacing}
\usetikzlibrary{decorations.text}
\definecolor{ffblue}{RGB}{097, 108, 140}
\definecolor{ffdarkgreen}{RGB}{086, 140, 135}
\definecolor{fflightgreen}{RGB}{178, 213, 155}
\definecolor{ffyellow}{RGB}{242, 222, 121}
\definecolor{ffred}{RGB}{217, 095, 024}

\definecolor{ffred_pv}{RGB}{202, 074, 046}
\definecolor{fforange_pv}{RGB}{232, 141, 047}
\definecolor{ffgreen_pv}{RGB}{059, 165, 149}
\definecolor{ffgreendark_pv}{RGB}{032, 117, 106}
\definecolor{nature_tab_gray1}{HTML}{D8D6C2}
\definecolor{nature_tab_gray2}{HTML}{ECEADF}
\ifCLASSINFOpdf
\else
\fi
\begin{document}
%
\title{Robotic Ultra-Long-Horizon Manipulation Skills via Human-guided Lifelong Code Generation}
%
%
%

\author{
Yuan Meng$^{1,*}$, 
Zhenguo Sun$^{1,2,3,*,\dagger}$, 
Max Fest$^{1}$, 
Xiangtong Yao$^{1}$, 
Xukun Li$^{1,2,3}$, \\
Zhenshan Bing$^{4,\#}$, 
and Alois Knoll$^{1}$~\IEEEmembership{Fellow,~IEEE}
\thanks{Manuscript received: May 4, 2026; Revised July 27, 2026; Accepted September 29, 2026.}
\thanks{This paper was recommended for publication by
Editor Aniket Bera upon evaluation of the Associate Editor and Reviewers
comments.} 
\thanks{*Equal contribution. $^{\dagger}$Project leader. $^{\#}$Corresponding author: {\tt\footnotesize bing@nju.edu.cn}.}
\thanks{$^{1}$Yuan Meng, Zhenguo Sun, Max Fest, Xiangtong Yao, Xukun Li, and Alois Knoll are with School of Computation, Information and Technology, Technical University of Munich, 85748 Munich, Germany.
(email: {\tt\footnotesize y.meng@tum.de; hitsunzhenguo@gmail.com; max.fest@tum.de; xiangtong.yao@tum.de; xukunli0221@gamil.com; k@tum.de})
}%
\thanks{$^{2}$Zhenguo Sun and Xukun Li are also with Beijing Academy of Artificial Intelligence (BAAI), 100084 Beijing, China.}%
\thanks{$^{3}$Zhenguo Sun and Xukun Li are also with XYZ Embodied AI, 100080 Beijing, China.}%
\thanks{$^{4}$Zhenshan Bing is with State Key Laboratory for Novel Software Technology, School of Intelligence Science and Technology, Nanjing University, 215163 Suzhou, China.
(email: {\tt\footnotesize bing@nju.edu.cn})
}%
\thanks{Code is available at: \protect\url{https://ghiara.github.io/LYRA/}.}
\thanks{Digital Object Identifier (DOI): see top of this page.}
}
%
%

\markboth{IEEE Robotics and Automation Letters. Preprint Version. Accepted September 2026}
{Meng \MakeLowercase{\textit{et al.}}: Robotic Ultra-Long-Horizon Manipulation Skills via Human-guided Lifelong Code Generation} 

%



\maketitle

\begin{abstract}
Large language models (LLMs) can translate natural-language instructions for robotic manipulation into executable code, but ambiguity, noisy generations, and limited context windows make ultra-long-horizon tasks unreliable.
Closed-loop approaches that rely only on LLM feedback also struggle because LLMs have limited robotic reasoning, even when task errors are obvious to humans.
Feedback is often stored in representations that generalize poorly to unseen tasks and can cause catastrophic forgetting as new corrections accumulate.
We propose LYRA, a human-guided lifelong skill learning and code generation framework that distills human feedback into modular, reusable skills and incrementally extends their functionality across successive interactions while preserving previously learned behavior.
External memory stores learned skills and execution examples; retrieval-augmented generation selects relevant knowledge, while user hints guide reuse when retrieval is insufficient, supporting ultra-long-horizon execution.
Experiments on Ravens, Franka Kitchen, LIBERO-long, MetaWorld, and real-world tasks show a 0.93 success rate, up to 27\% higher than baselines, and a 42\% improvement in correction efficiency.
LYRA also robustly solves ``build a house'', which requires planning over 20 primitives.
\end{abstract}


%
\IEEEpeerreviewmaketitle

\section{INTRODUCTION}
\begin{figure*}[ht!]
    \centering
    \resizebox{.95\textwidth}{!}{
    \begin{tikzpicture}
        \coordinate (A) at (0,0);
        \coordinate (B) at (-2.6cm,-4.5cm);
        \coordinate (C) at ( 2.6cm,-4.5cm);
        \coordinate (AB) at ($ (A)!0.5!(B) $);
        \coordinate (AC) at ($ (A)!0.5!(C) $);
        \coordinate (BC) at ($ (B)!0.5!(C) $);
        \begin{scope}[on background layer]
            \coordinate (G) at (barycentric cs:A=1,B=1,C=1);
            \def\padscale{1.8} 
            \coordinate (Aexp) at ($ (G)!\padscale!(A) $);
            \coordinate (Bexp) at ($ (G)!\padscale!(B) $);
            \coordinate (Cexp) at ($ (G)!\padscale!(C) $);
            \filldraw[rounded corners=2cm, draw=white, shading=axis, top color=fforange_pv!30, bottom color=fflightgreen!30] 
            ([yshift=-.4cm]Aexp) -- ([yshift=-.4cm]Bexp) -- ([yshift=-.4cm]Cexp) -- cycle;
        \end{scope}
        \draw[draw=white, shading=axis, top color=fforange_pv!30, bottom color=fflightgreen!30, rounded corners=4pt](15.3cm, 1.2cm) -- (24.1cm, 1.2cm) -- (24.1cm, -6.6cm) -- (15.3cm, -6.6cm) -- cycle;
        \node[circle, clip, minimum width=.5em, inner sep=-.1em, label={[label distance=-1pt]-90: \textsf{User}}](user)at(A){\includegraphics[width=2em]{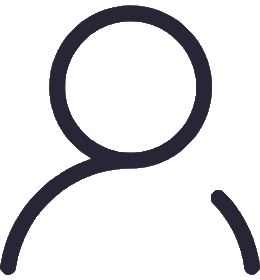}};
        \node[draw=fforange_pv, fill=none, circle, line width=2pt, minimum width=3em]at(user.center){};
        \node[draw, circle, clip, minimum width=1.5em, inner sep=-.45em, label={[label distance=-1pt]-90: \textsf{Agent}}, below left](agent)at(B){\includegraphics[width=3em]{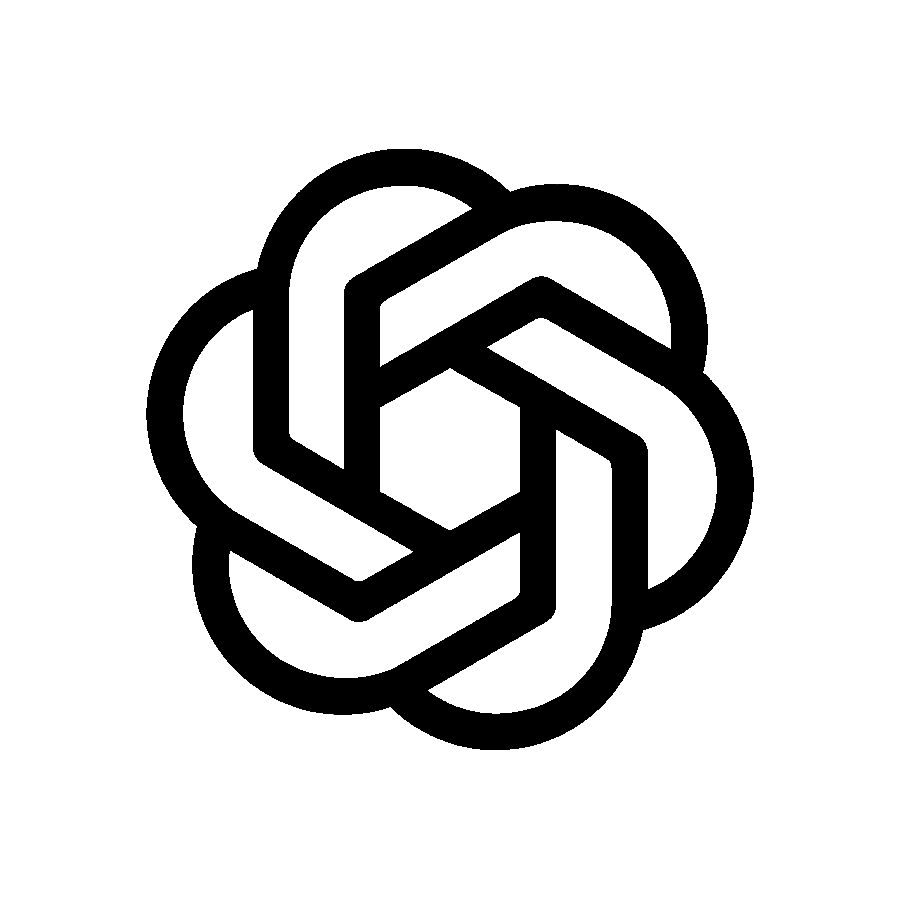}};
        \node[draw=fflightgreen, fill=none, circle, line width=2pt, minimum width=3em]at(agent.center){};
        \node[draw, circle, clip, minimum width=1.5em, inner sep=-.39em, label={[label distance=-1pt]-90: \textsf{Memory}}, below right](memory)at(C){\includegraphics[width=2.8em]{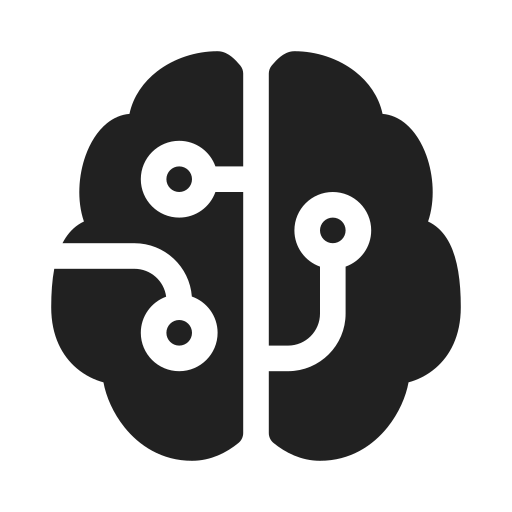}};
        \node[draw=ffgreen_pv, fill=none, circle, line width=2pt, minimum width=3em]at(memory.center){};
        \node[](lyra)at(0, -3cm){\includegraphics[width=2cm]{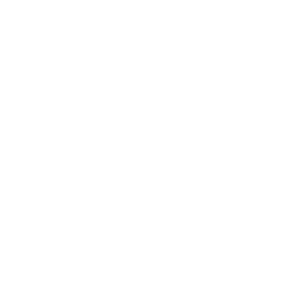}};
        \node[double arrow,minimum width=1cm, minimum height=3.5cm,
            shading=axis,left color=fflightgreen!60,right color=fforange_pv!60,
            rotate=240] at ([xshift=-.2cm, yshift=-.3cm]AB) {};
        \node[double arrow,minimum width=1cm, minimum height=3.5cm,
            shading=axis,left color=fforange_pv!60,right color=ffgreen_pv!60,
            rotate=300] at ([xshift=.2cm, yshift=-.3cm]AC) {};  
        \node[double arrow,minimum width=1cm, minimum height=3.5cm,
            shading=axis,left color=fflightgreen!60,right color=ffgreen_pv!60,
            rotate=0] at ([yshift=-.3cm]BC) {};
        \node[rotate=60]at([xshift=-.2cm, yshift=-.3cm]AB){\textsf{Skill learning}};
        \node[]at([yshift=-.3cm]BC){\textsf{Skill storage}};
        \node[rotate=-60]at([xshift=.2cm, yshift=-.3cm]AC){\textsf{Hint}};
        \node[]at([yshift=-3cm]A){\Large \textsf{\textbf{LYRA}}};
        \coordinate (D) at ([xshift=9cm, yshift=-3cm]user);
        \node[inner sep=0, opacity=.5](img1) at([xshift=-3cm, yshift=1.25cm]D){\includegraphics[width=3cm]{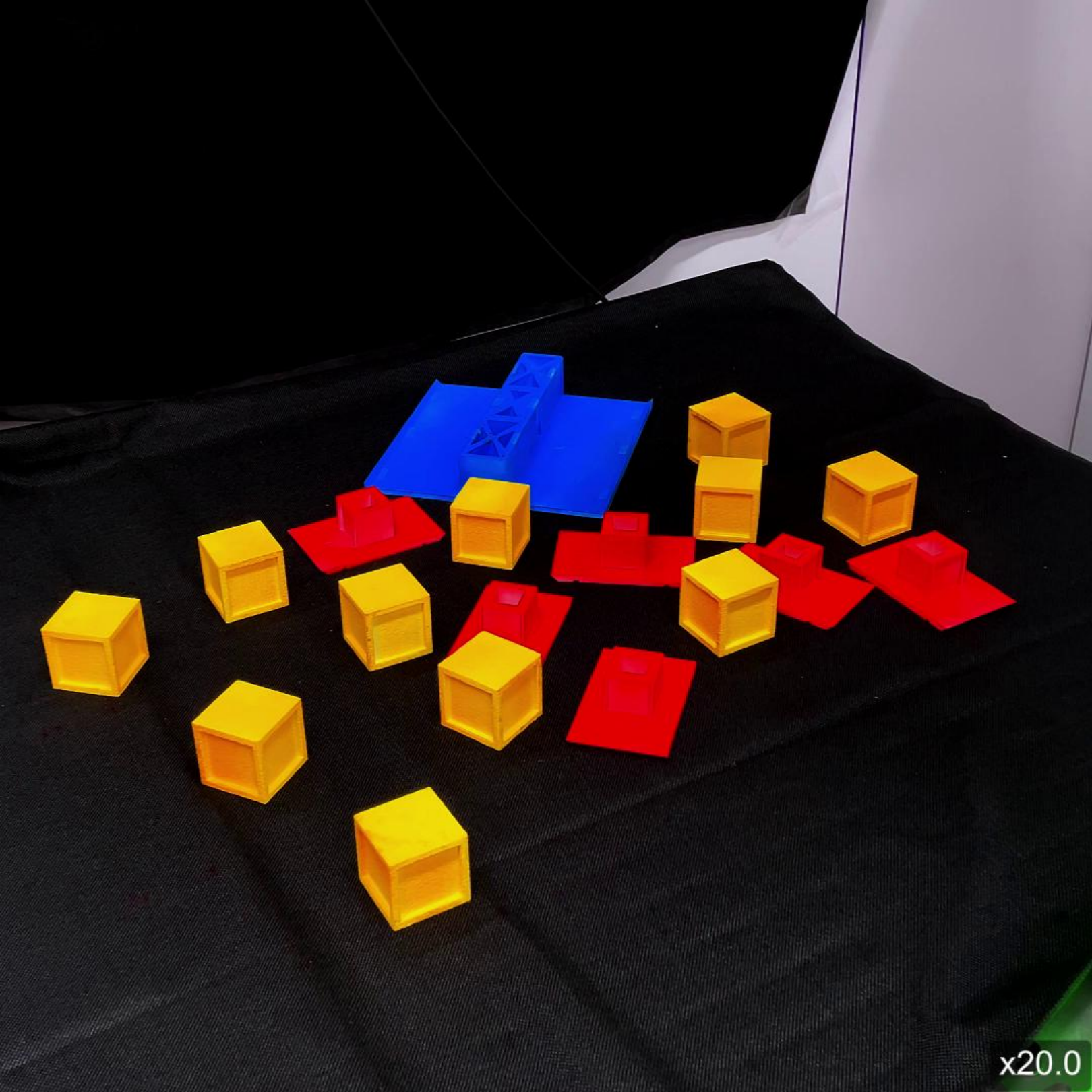}};
        \node[inner sep=-1pt](img2) at([xshift=.25cm, yshift=-.25cm]img1){\includegraphics[width=3cm]{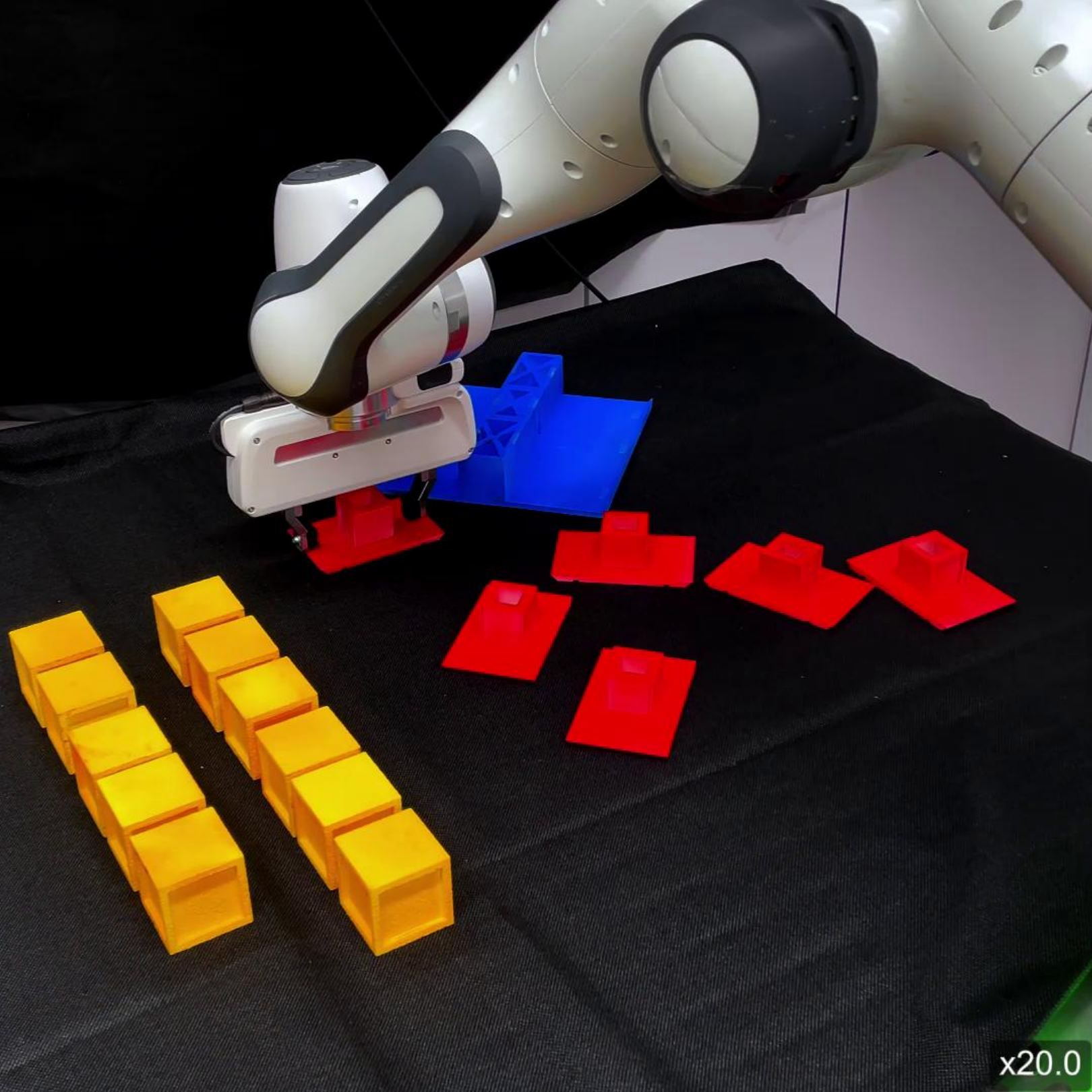}};
        \node[inner sep=0, opacity=.5](img3) at([xshift=3cm, yshift=-1.25cm]D){\includegraphics[width=3cm]{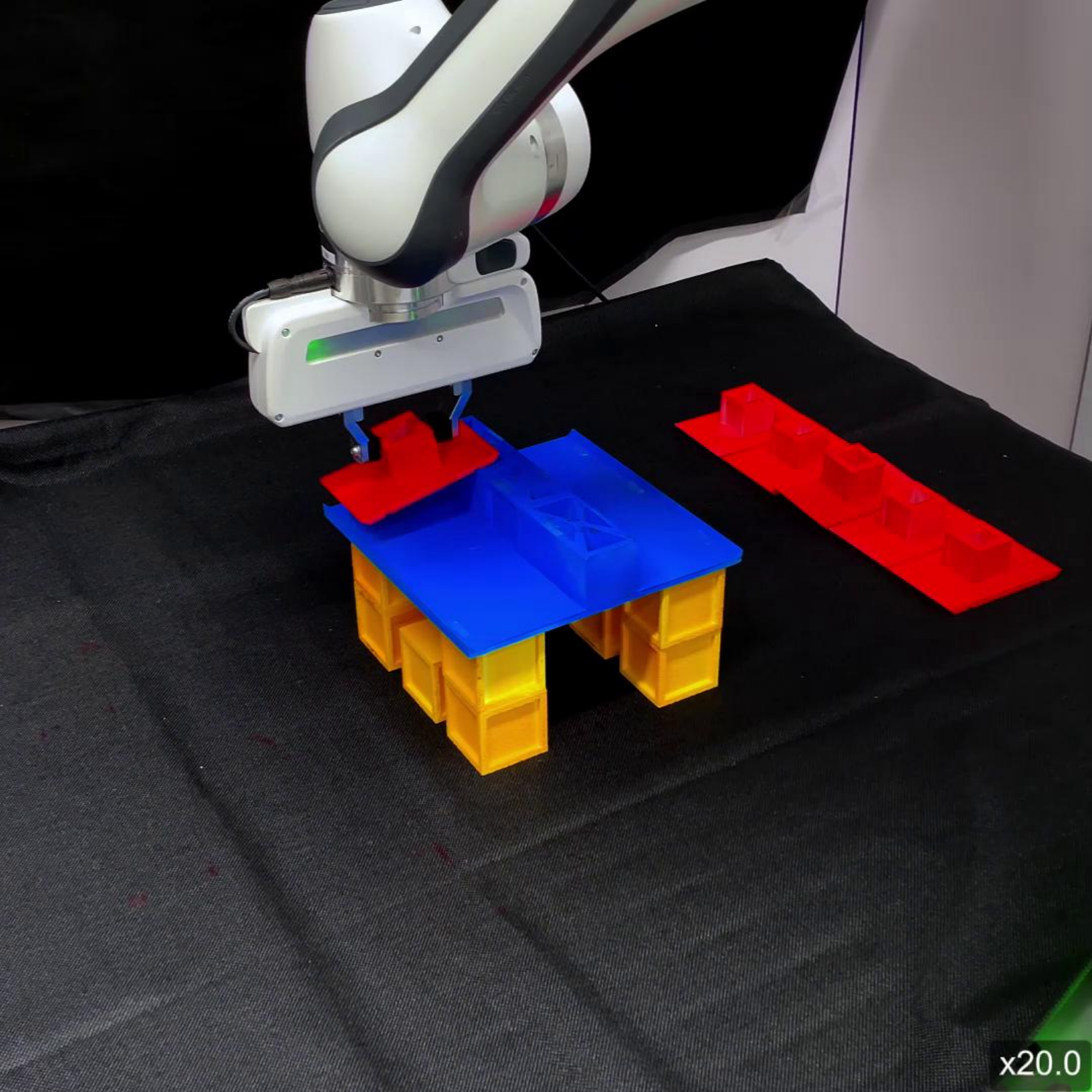}};
        \node[inner sep=0,](img4) at([xshift=-.25cm, yshift=.25cm]img3){\includegraphics[width=3cm]{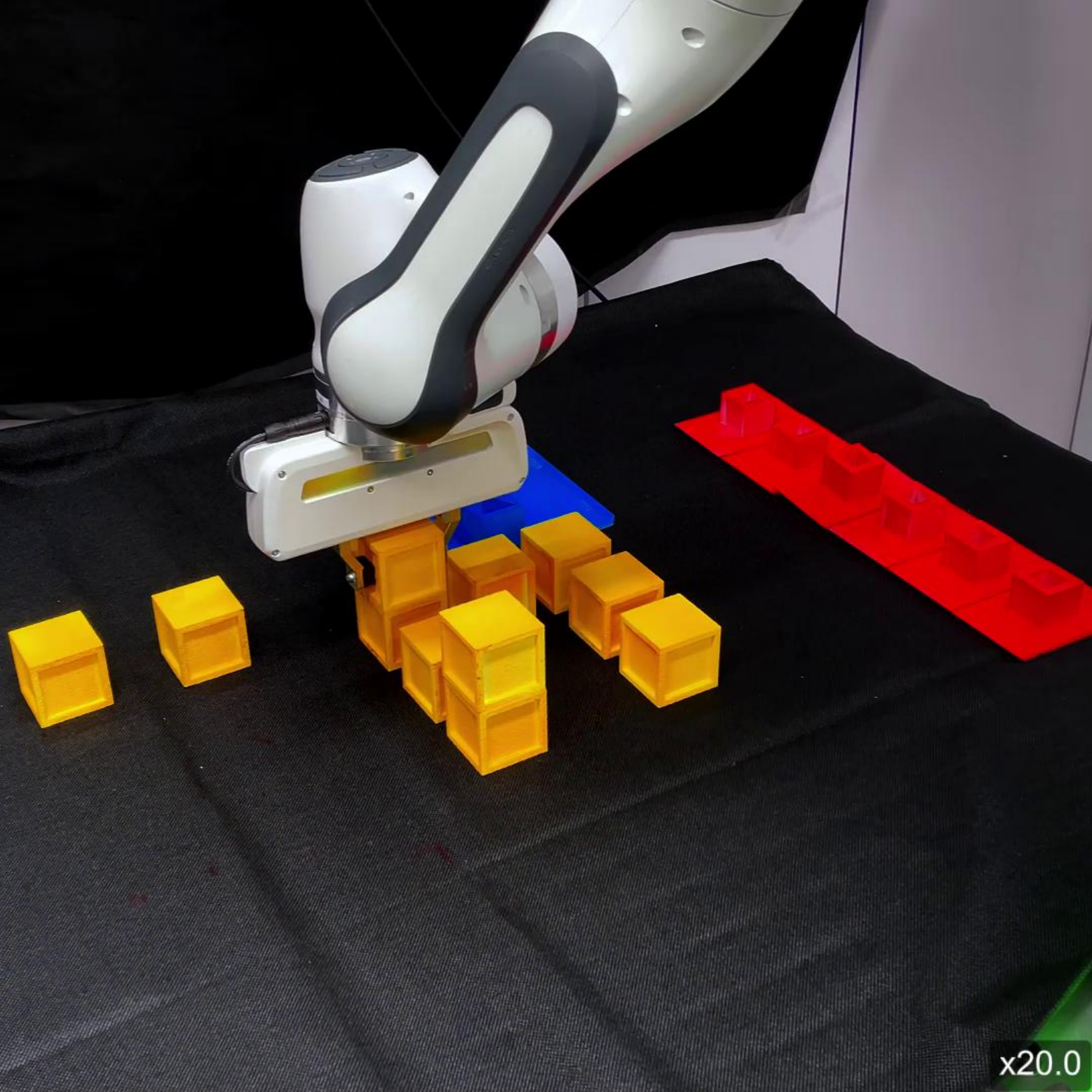}};
        \node[inner sep=0](cover)at(D) {\includegraphics[width=4.5cm]{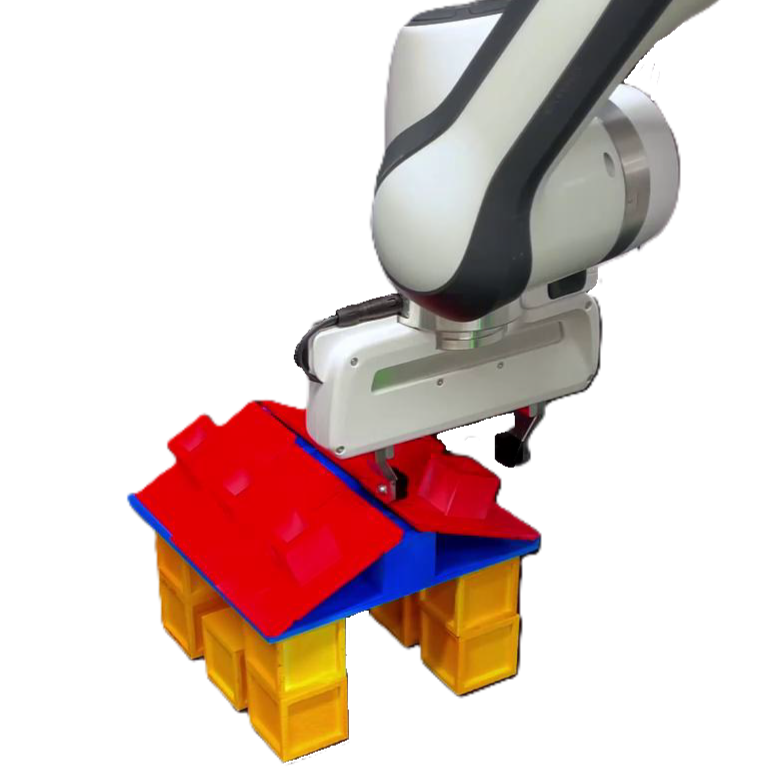}};
        \node[inner sep=0, below right](code)at([xshift=.1cm, yshift=-.2cm]cover.north west){\includegraphics[width=1cm]{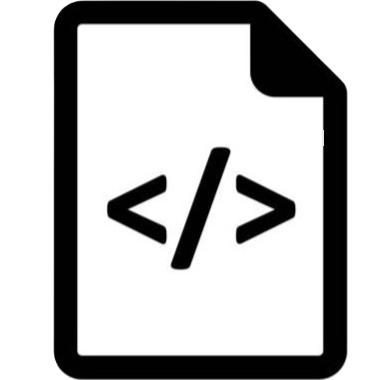}};
        \node[inner sep=0pt, right](img1)at([xshift=4.25cm]cover.east){\includegraphics[width=2cm]{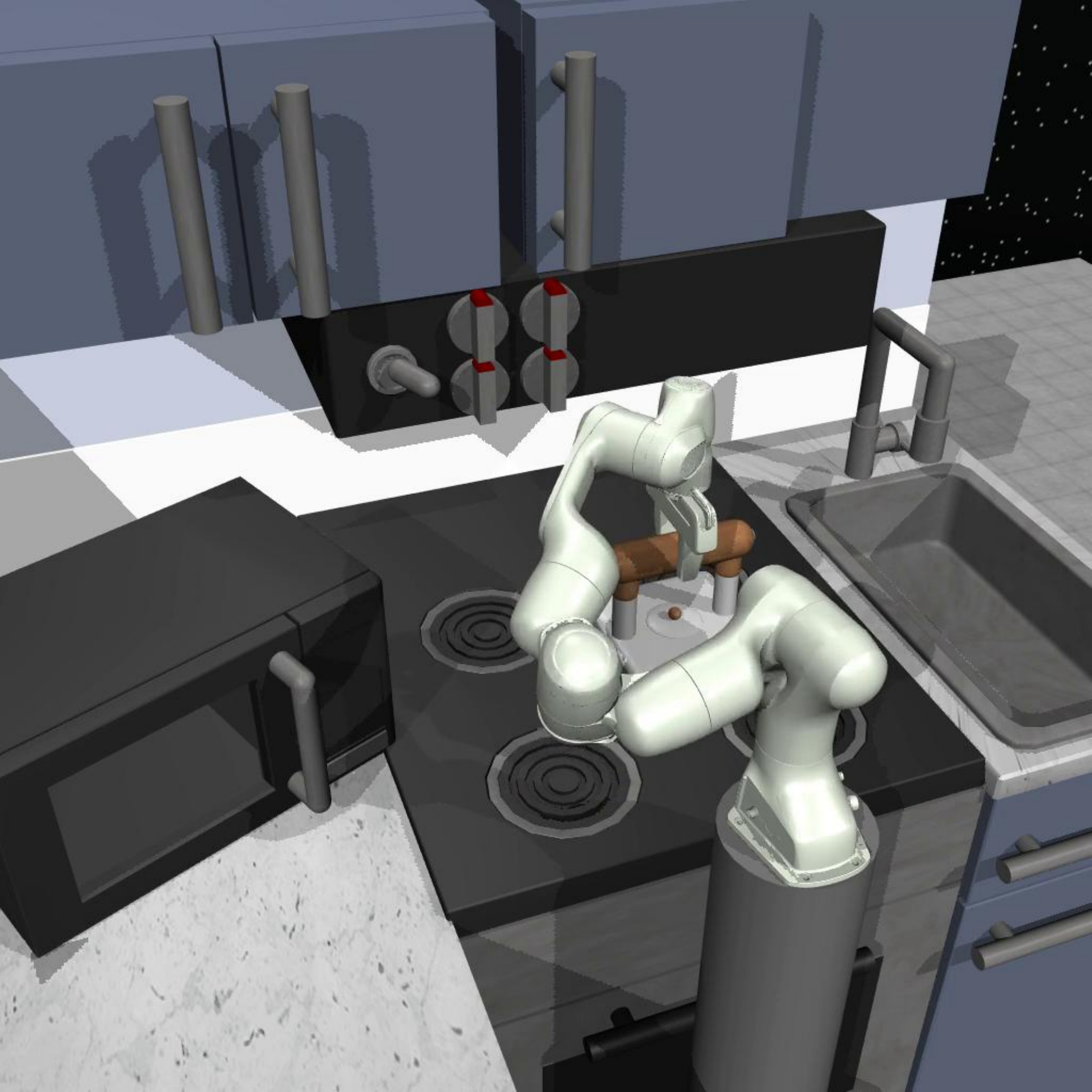}};
        \node[inner sep=0pt, right](img2)at([xshift=.1cm]img1.east){\includegraphics[width=2cm]{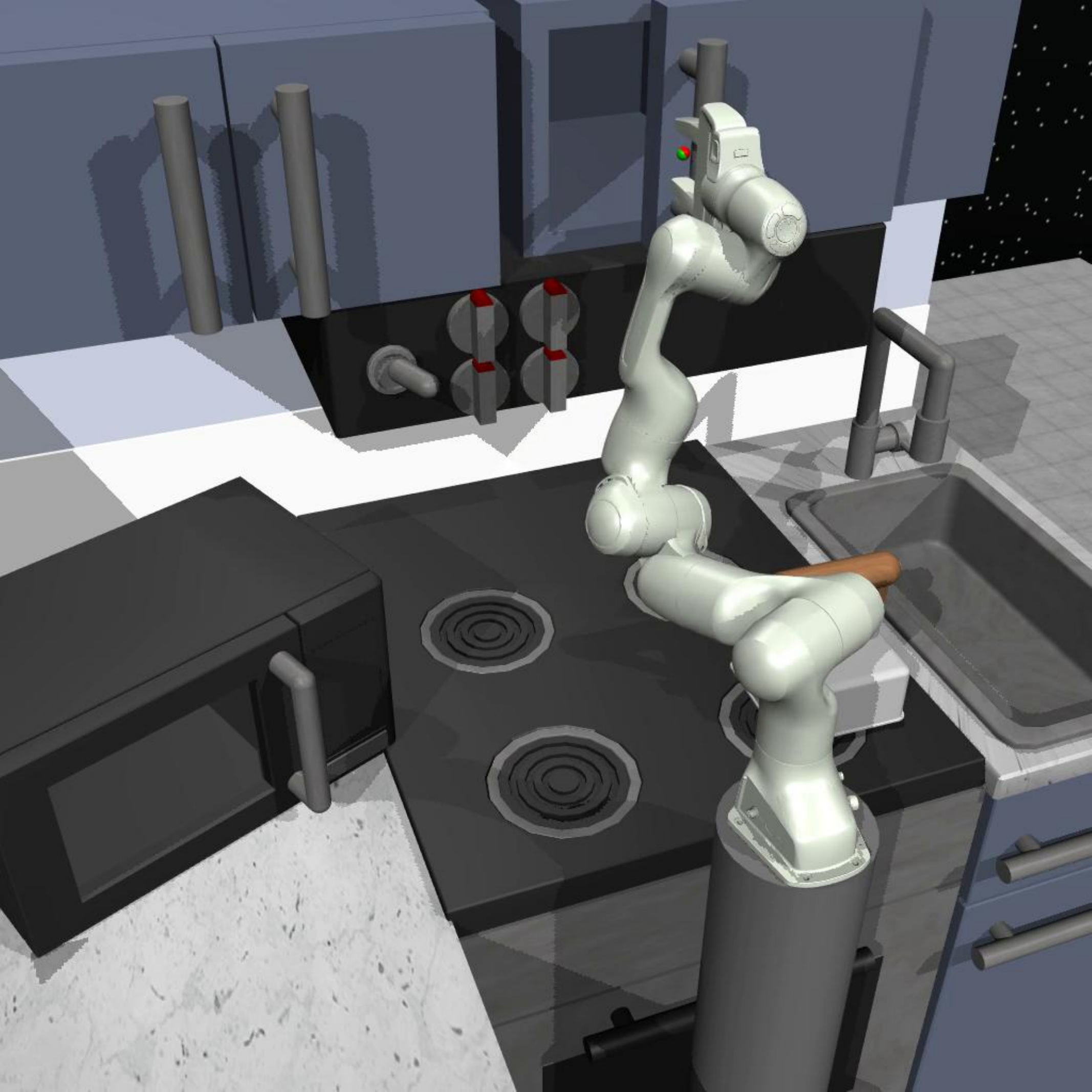}};
        \node[inner sep=0pt, right](img3)at([xshift=.1cm]img2.east){\includegraphics[width=2cm]{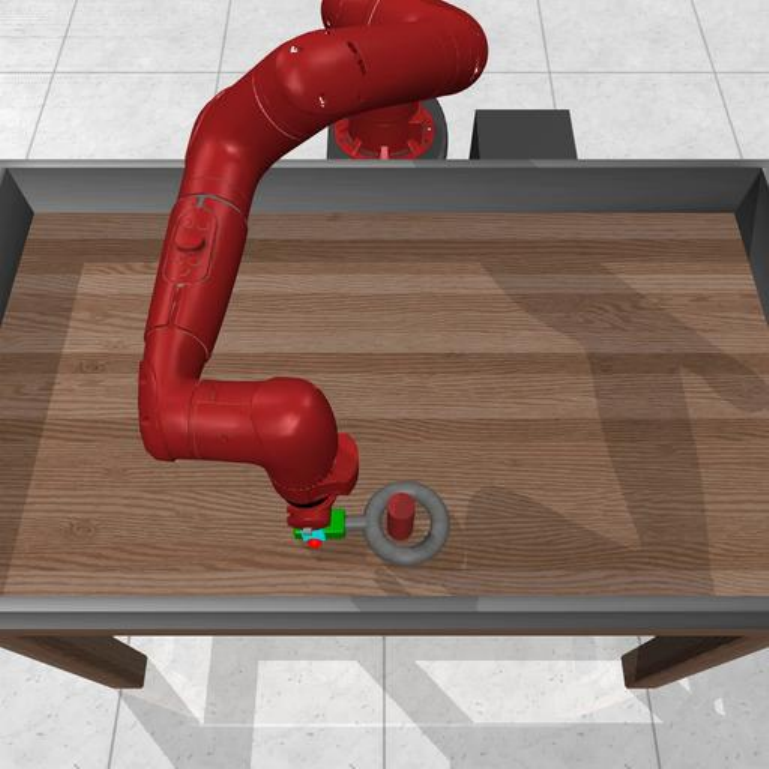}};
        \node[inner sep=0pt, right](img4)at([xshift=.1cm]img3.east){\includegraphics[width=2cm]{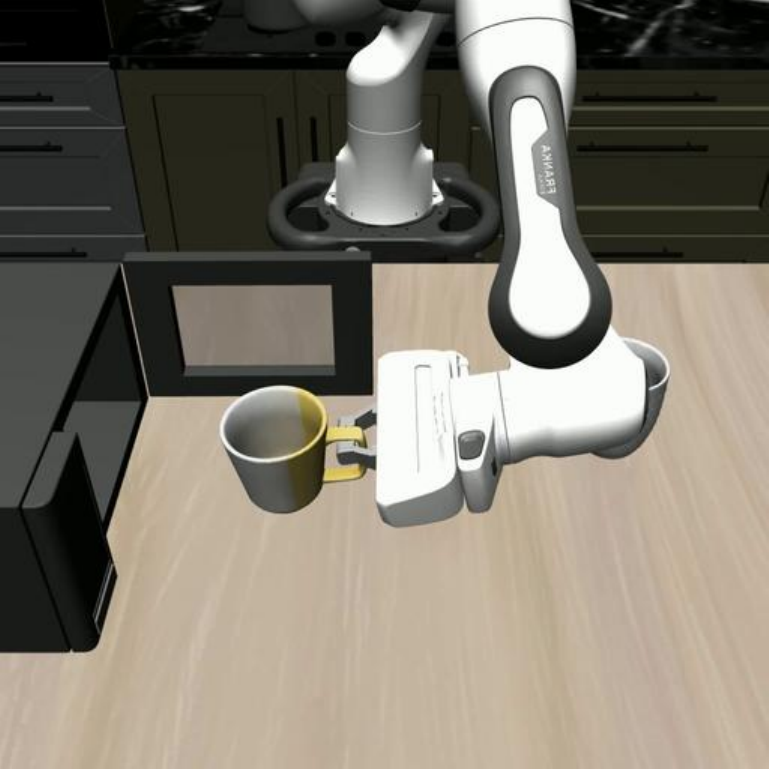}};
        
        \node[inner sep=0, above](img5) at([yshift=.1cm]img1.north){\includegraphics[width=2cm]{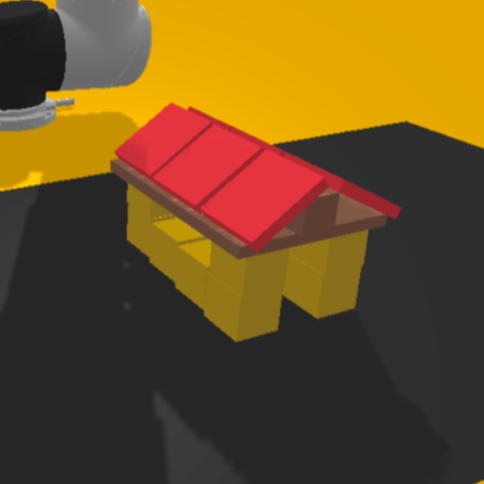}};
        \node[inner sep=0, right](img6) at([xshift=.1cm]img5.east){\includegraphics[width=2cm]{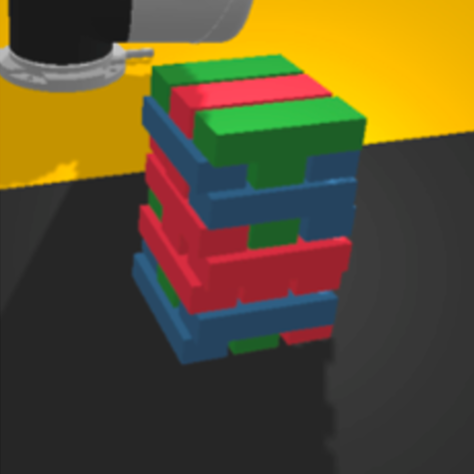}};
        \node[inner sep=0, right](img7) at([xshift=.1cm]img6.east){\includegraphics[width=2cm]{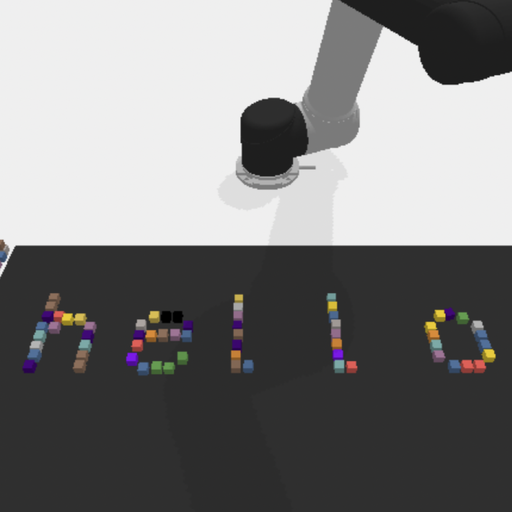}};
        \node[inner sep=0, right](img8) at([xshift=.1cm]img7.east){\includegraphics[width=2cm]{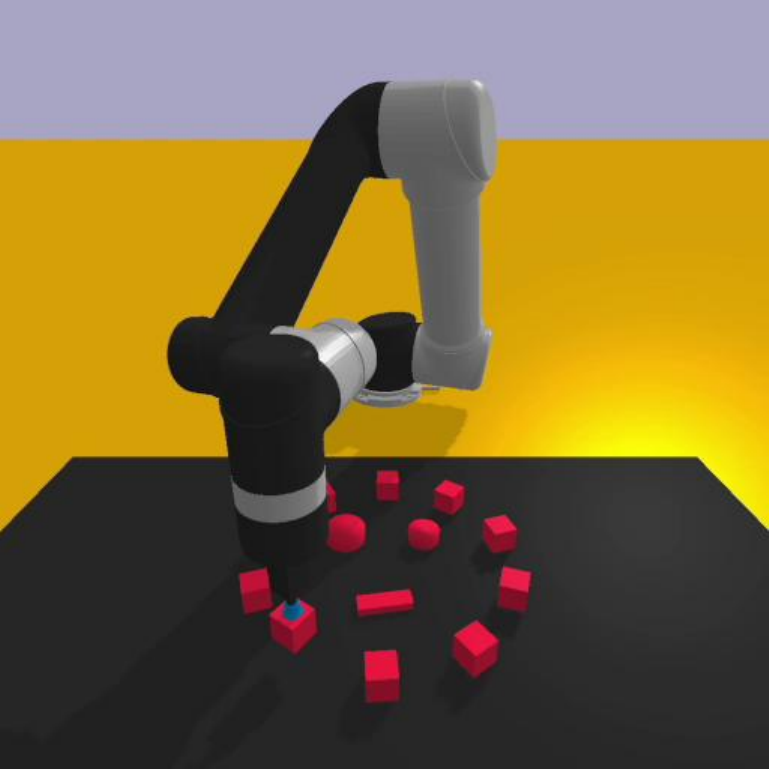}};
        
        \node[inner sep=0, below](img9) at([yshift=-.1cm]img1.south){\includegraphics[width=2cm]{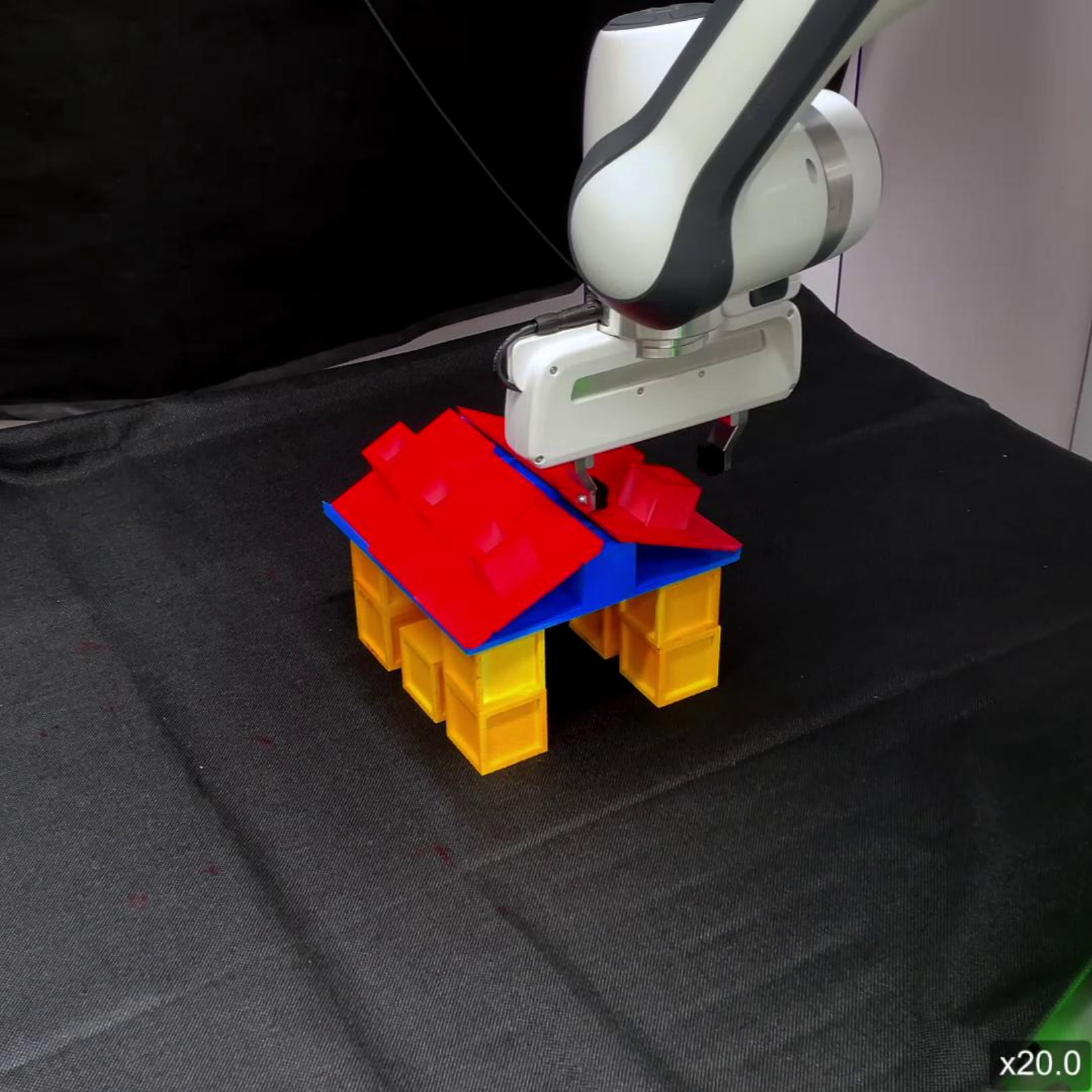}};
        \node[inner sep=0, right](img10) at([xshift=.1cm]img9.east){\includegraphics[width=2cm]{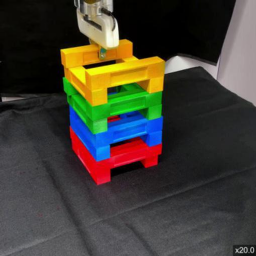}};
        \node[inner sep=0, right](img11) at([xshift=.1cm]img10.east){\includegraphics[width=2cm]{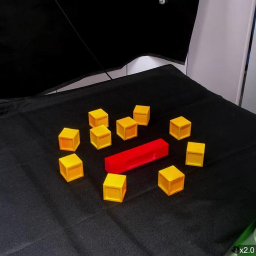}};
        \node[inner sep=0, right](img12) at([xshift=.1cm]img11.east){\includegraphics[width=2cm]{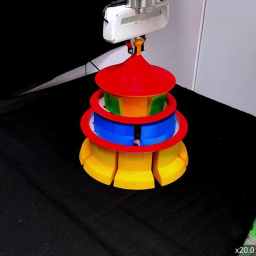}};
        \draw[-{Triangle Cap []. Fast Triangle[]}, draw=fflightgreen!80, line width=5pt, rounded corners=4pt] ([yshift=-.5cm]agent.south) --++ (0, -.5cm) -| (cover.south);
        \draw[-{Triangle Cap []. Fast Triangle[]}, draw=ffgreen_pv!80, line width=5pt, rounded corners=4pt] ([yshift=-.5cm]memory.south) --++ (0, -.5cm) -| (cover.south);
        \draw[-{Triangle Cap []. Fast Triangle[]}, draw=ffyellow!80, line width=5pt, rounded corners=4pt] ([yshift=-.75cm]cover.north east) --++ (3cm, 0) |- (img5.west);
        \draw[-{Triangle Cap []. Fast Triangle[]}, draw=fforange_pv!80, line width=5pt, rounded corners=4pt] ([yshift=-1cm]cover.north east) --++ (3cm, 0) |- (img1.west);
        \draw[-{Triangle Cap []. Fast Triangle[]}, draw=ffred_pv!80, line width=5pt, rounded corners=4pt] ([yshift=-1.25cm]cover.north east) --++ (2.75cm, 0) |- (img9.west);
        \node[rectangle, draw=fforange_pv!80, line width=2pt, rounded corners=4pt, text width=9cm, align=center, right](reply)at([xshift=1cm,yshift=.5cm]user.east){\textsf{You should try to preserve the previous functionality.}};
        \node[above]at([xshift=.3cm, yshift=0.2cm]cover.north east){\textbf{\textsf{Task-specific plan deployment}}};
        \node[above]at([xshift=-.5cm, yshift=-1.0cm]cover.south west){\textbf{\textsf{Skills \& examples retrieval}}};
        \node[above]at([xshift=0cm, yshift=0cm]cover.south){\textbf{\textsf{Task code plan generation}}};
        \node[above]at([yshift=.2cm]img6.north east){\large \textsf{\textbf{Long Horizon Manipulation Tasks}}};
        \node[below]at([xshift=.1cm]img10.south east){\large $\bm{\cdots}$};
        
    \end{tikzpicture}
    }
    \vskip -.1in
    \caption{Framework overview: human-guided lifelong skill code generation and ultra-long-horizon task deployment.}
    \label{fig:overview}
\end{figure*}
%
\IEEEPARstart{L}{arge} language models (LLMs) and vision-language models (VLMs) bring commonsense knowledge and reasoning to robotic manipulation.
Early approaches \cite{yao2023bridging,meng2025preserving,meng2025pretrained} used language embeddings in reinforcement or imitation learning to align robot actions with human commands, but had limited data efficiency and generalization.
Later systems use LLMs such as GPT for task decomposition and language grounding \cite{zhang2023lohoravens,huang2023inner}: an LLM splits a complex task into sub-tasks and calls a pre-trained language-conditioned policy for low-level primitives.
This separates high-level planning from control, but assumes that the policy can execute every requested motion precisely despite environmental perturbations and imperfect policy design.
Another direction uses LLM or VLM backbones to pretrain end-to-end Vision–Language–Action (VLA) models on robotic data \cite{black2024pi_0}.
Robotic data are much scarcer than data for computer vision or natural language, making large-scale pretraining difficult.
These models often have at least 7B parameters, require substantial compute, and are difficult to deploy on edge devices.
These constraints have slowed progress.
Well-pretrained LLMs can instead follow instructions and directly generate code for robot control.

LLM-generated code for robotic manipulation \cite{zhang2025generative} offers another route: code itself represents the robot's actions and can draw on an LLM's instruction-following ability.
Code-as-Policies (CaP) \cite{liang2023code,fu2026cap}, for example, translates human instructions into executable Python code using fixed perception and control primitives, allowing rapid deployment of new behaviors.
Generated code can then be tested and revised during interaction.
Yet three limitations remain:
(1) LLM outputs are noisy and error-prone in long-horizon planning; 
(2) ambiguous language can obscure the user's intent; 
(3) pre-defined primitives and handcrafted prompts restrict the agent, while limited context windows constrain the number of examples. 
Together, these limitations hinder complex long-horizon tasks: errors compound as plans lengthen, while a fixed primitive set and a finite prompt limit adaptation.
Feedback-based approaches \cite{liang2024learning,zhi2025closed,meng2025data} improve robustness, but verifiers must repeat the same corrections even for previously seen tasks, so the system has not retained what it learned.
Dynamic prompt adjustment can support LLM in-context adaptation, but useful corrections must first be represented and retrieved.
This raises two questions.
First, \textbf{how should feedback knowledge be stored and reused?}
A correction should remain useful on related tasks without asking a verifier to repeat it.
Direct prompt updates \cite{arenas2024prompt} can cause catastrophic forgetting, sharply reducing performance on earlier tasks as new instructions are added.
Cross-task insights \cite{zhao2024expel,zha2024distilling} often remain ambiguous and task-specific, limiting generalization to unseen tasks.
Flat code is easy to modify, but repeated edits introduce noise and errors that destabilize long-horizon generation.
Encapsulating temporally extended behaviors as \textbf{skill functions} offers a more stable representation: parameters can be adjusted for new tasks without repeatedly rewriting the behavior, providing flexibility and generalization while preserving stability.
Second, \textbf{which feedback is most efficient during interaction?}
Language is a convenient interface, but feedback generated solely by LLMs \cite{arenas2024prompt,wang2023voyager,meng2025data} is often unreliable: it may diverge from human preferences or fail on ultra-long-horizon reasoning \cite{wang2023voyager}.
Humans can usually identify, evaluate, and correct robotic task errors that an LLM may miss.
In robotic code generation, those corrections can be tested and verified within seconds, whereas training conventional deep learning models often takes hours or days.
This short feedback cycle makes \textbf{human-guided robotic skill learning} both reliable and efficient.

We therefore propose a human-guided framework that turns user feedback into reusable skills and extends their functionality through a user-designed curriculum, enabling preference-aligned lifelong learning (Fig. \ref{fig:overview}).
Across interactions, the curriculum expands what a skill can do while preserving the functionality already learned.
External memory stores learned skills and execution examples to prevent catastrophic forgetting and support reuse across tasks.
Retrieval-Augmented Generation (RAG) selects relevant skills and examples for dynamic in-context learning during long-horizon planning, without placing the entire accumulated experience in a limited prompt.
When retrieval is insufficient, user hints steer the agent toward the skills needed for the current task, improving efficiency and reducing interference from irrelevant data.
Our system-level contribution is to show, theoretically and empirically, that \textbf{human feedback in robot control code generation can become persistent, generalizable, and compositional skill knowledge} rather than repeated local corrections.
The framework connects feedback, skill storage, and code generation so that corrections made during one task can inform later tasks.
Preserving, reusing, and accumulating this knowledge supports lifelong learning and ultra-long-horizon manipulation beyond fully automated LLM-based systems.
We evaluate the framework on Ravens, Franka-Kitchen, LIBERO-long, MetaWorld, and real-world long-horizon tasks with a Franka FR3.
It achieves a 0.93 success rate, up to 27\% higher than baselines, and reduces correction rounds by 42\% relative to LLM-based closed-loop methods.
It also robustly solves ``build a house'', which requires planning over 20 primitives.

Our contributions are threefold:
(1) \textbf{Persistent skill knowledge}: We establish a human-guided lifelong learning framework that distills feedback into stable, reusable, and compositional skills, retaining corrections across tasks while extending previously learned functions.
(2) \textbf{Memory-augmented code synthesis}: External memory stores skills and execution examples; RAG and user hints retrieve, reuse, and selectively compose them for dynamic in-context learning and long-horizon planning while limiting interference from irrelevant experience.
(3) \textbf{Ultra-long-horizon validation}: Experiments across Ravens, Franka-Kitchen, LIBERO-long, MetaWorld, and the real-world Franka FR3 demonstrate the reported success and correction gains, including planning and execution over 20 primitives in ``build a house''.
We call our framework \textbf{LYRA}: a \textbf{L}ifelong learning code s\textbf{Y}nthesis framework for \textbf{R}obotic long-horizon skill \textbf{A}cquisition.

\section{METHOD}

\begin{figure*}[ht!]
    \centering
    \resizebox{\textwidth}{!}{
    \begin{tikzpicture}
        \node[above right](phase1)at(-2cm,1cm){\Large\textcolor{ffblue}{\textbf{\textsf{Phase I: Preference-aligned Skill Acquisition}}}};
        \draw[ffblue, rounded corners=4pt, line width=2pt]
            (-2cm,1cm) -- (27cm,1cm) -- (27cm,-6.cm) -- (-2,-6.cm) -- cycle;
        \node[draw, circle, clip, minimum width=1.5em, inner sep=-.5em, label={[label distance=-1pt]-90:\small \textsf{Agent}}](agent1)at(-.5cm,.3cm){\includegraphics[width=3em]{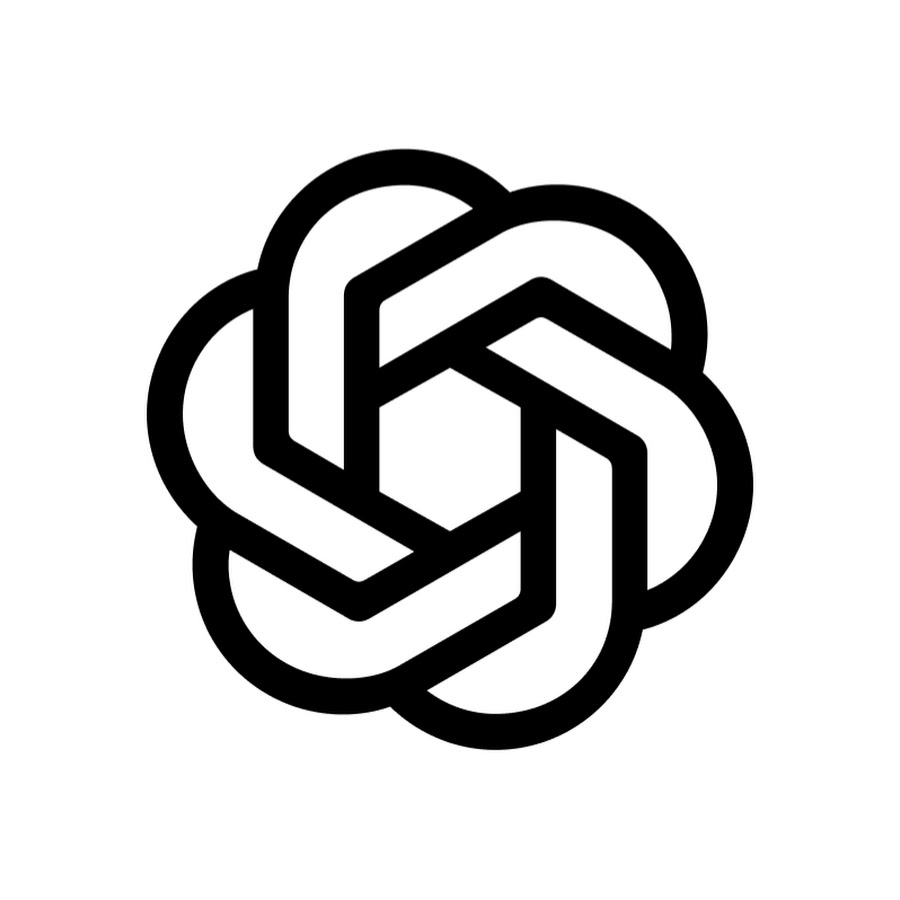}};
        \node[draw=fflightgreen, fill=none, circle, line width=2pt, minimum width=3em]at(agent1.center){};
        \node[draw, circle, clip, minimum width=1.5em, inner sep=-.5em, label={[label distance=-1pt]-90:\small \textsf{Agent}}](agent2)at([xshift=5.5cm]agent1.east){\includegraphics[width=3em]{imgs/openai.jpg}};
        \node[draw=fflightgreen, fill=none, circle, line width=2pt, minimum width=3em]at(agent2.center){};
        \node[draw, circle, clip, minimum width=1.5em, inner sep=-.5em, label={[label distance=-1pt]-90:\small \textsf{Agent}}](agent3)at([xshift=5.5cm]agent2.east){\includegraphics[width=3em]{imgs/openai.jpg}};
        \node[draw=fflightgreen, fill=none, circle, line width=2pt, minimum width=3em]at(agent3.center){};
        \node[draw, circle, clip, minimum width=1.5em, inner sep=-.5em, label={[label distance=-1pt]-90:\small \textsf{Agent}}](agent4)at([xshift=5.5cm]agent3.east){\includegraphics[width=3em]{imgs/openai.jpg}};
        \node[draw=fflightgreen, fill=none, circle, line width=2pt, minimum width=3em]at(agent4.center){};
        \node[draw, circle, clip, minimum width=1.5em, inner sep=-.5em, label={[label distance=-1pt]-90:\small \textsf{Agent}}](agent5)at([xshift=5.5cm]agent4.east){\includegraphics[width=3em]{imgs/openai.jpg}};
        \node[draw=fflightgreen, fill=none, circle, line width=2pt, minimum width=3em]at(agent5.center){};
        \node[below](img1)at([yshift=-2.5em]agent1.south){\includegraphics[width=7em]{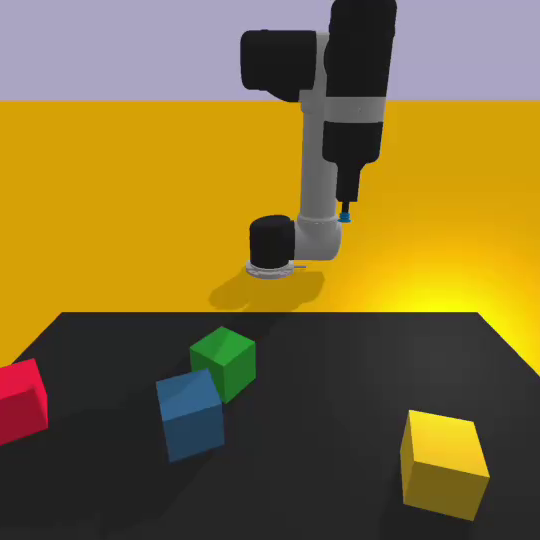}};
        \node[below](img2)at([yshift=-2.5em]agent2.south){\includegraphics[width=7em]{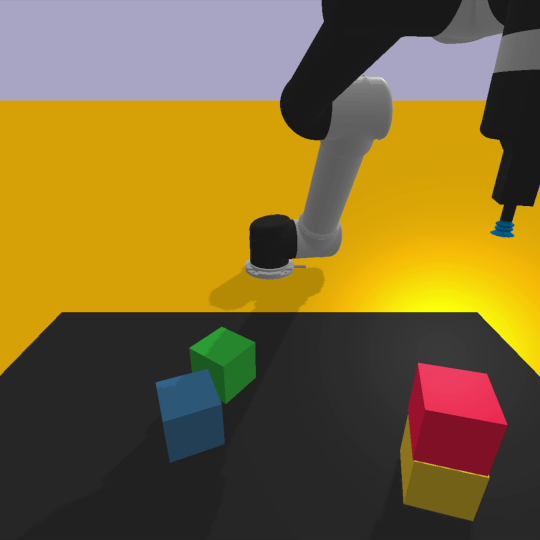}};
        \node[below](img3)at([yshift=-2.5em]agent3.south){\includegraphics[width=7em]{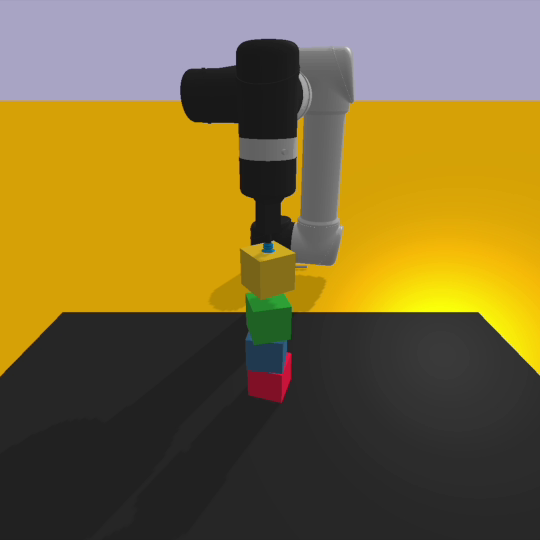}};
        \node[below](img4)at([yshift=-2.5em]agent4.south){\includegraphics[width=7em]{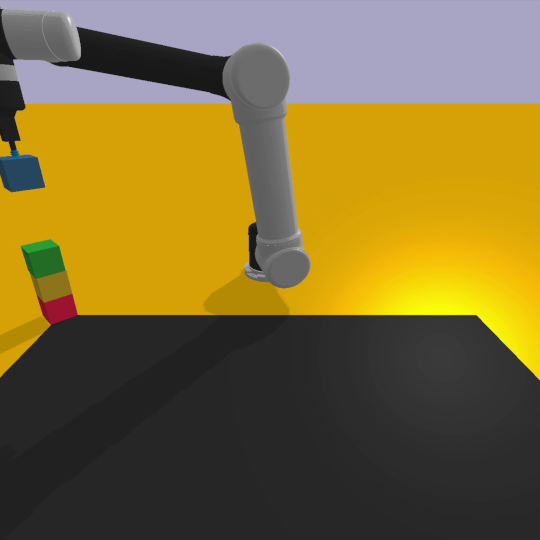}};
        \node[below](img5)at([yshift=-2.5em]agent5.south){\includegraphics[width=7em]{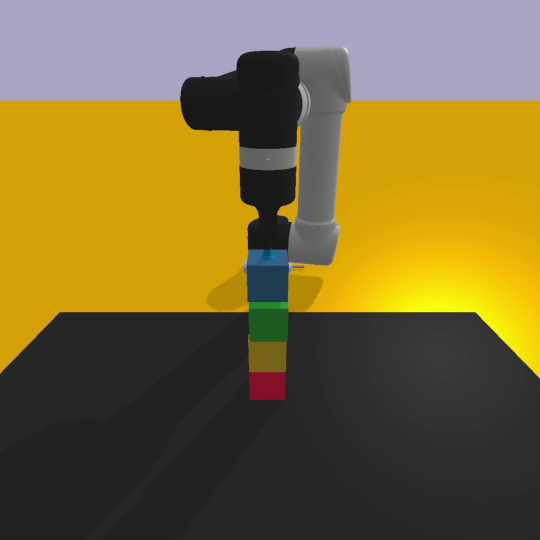}};
        \node[circle, clip, minimum width=.5em, inner sep=-.1em, label={[label distance=-1pt]-90:\small \textsf{User}}](user1)at([yshift=-2em]img1.south){\includegraphics[width=2em]{imgs/user.png}};
        \node[draw=fforange_pv, fill=none, circle, line width=2pt, minimum width=3em]at(user1.center){};
        \node[circle, clip, minimum width=.5em, inner sep=-.1em, label={[label distance=-1pt]-90:\small \textsf{User}}](user2)at([yshift=-2em]img2.south){\includegraphics[width=2em]{imgs/user.png}};
        \node[draw=fforange_pv, fill=none, circle, line width=2pt, minimum width=3em]at(user2.center){};
        \node[circle, clip, minimum width=.5em, inner sep=-.1em, label={[label distance=-1pt]-90:\small \textsf{User}}](user3)at([yshift=-2em]img3.south){\includegraphics[width=2em]{imgs/user.png}};
        \node[draw=fforange_pv, fill=none, circle, line width=2pt, minimum width=3em]at(user3.center){};
        \node[circle, clip, minimum width=.5em, inner sep=-.1em, label={[label distance=-1pt]-90:\small \textsf{User}}](user4)at([yshift=-2em]img4.south){\includegraphics[width=2em]{imgs/user.png}};
        \node[draw=fforange_pv, fill=none, circle, line width=2pt, minimum width=3em]at(user4.center){};        
        \node[circle, clip, minimum width=.5em, inner sep=-.1em, label={[label distance=-1pt]-90:\small \textsf{User}}](user5)at([yshift=-2em]img5.south){\includegraphics[width=2em]{imgs/user.png}};
        \node[draw=fforange_pv, fill=none, circle, line width=2pt, minimum width=3em]at(user5.center){};
        \node[rectangle, draw, fill=white, rounded corners=2pt, text width=3.7cm, below right](code1)at([xshift=1em]agent1.north east)
        {\scriptsize \textsf{\textcolor{RoyalBlue}{def} stack\_blocks(blocks): \\ ~ for block in blocks[1:]: \\ ~~ put\_first\_on\_second( blocks[0].Pose, block.Pose )}};
        \node[rectangle, draw, rounded corners=2pt, text width=3.75cm, below right](reply1)at([xshift=1em]user1.north east)
        {\small\textsf{learn a skill that \textbf{stack the blocks}, use 4 random blocks for initial setup.}};
        
        \node[rectangle, draw, fill=white, rounded corners=2pt, text width=3.7cm, below right](code2)at([xshift=1em]agent2.north east)
        {\scriptsize \textsf{\textcolor{RoyalBlue}{def} stack\_blocks(blocks): \\ $\cdots$ \\ cur\_pose = blocks[0].Pose \\ for block in blocks[1:]: \\ ~ put\_first\_on\_second( block.Pose, cur\_pose ) \\ ~ \textcolor{ffgreen_pv}{cur\_pose.z += block.size[2]} }};
        \node[rectangle, draw, rounded corners=2pt, text width=3.75cm, below right](reply2)at([xshift=1em]user2.north east)
        {\small\textsf{This is not stack. You need to stack ALL the blocks in a tower.}};
        
        \node[rectangle, draw, fill=white, rounded corners=2pt, text width=3.7cm, below right](code3)at([xshift=1em]agent3.north east)
        {\scriptsize \textsf{\textcolor{RoyalBlue}{def} stack\_blocks(blocks): \\ $\cdots$ \\ \textcolor{ffgreen_pv}{cur\_pose.rot = np.deg2rad(45)} \\ ~ cur\_pose = Pose( $\cdots$ \\  \textcolor{ffgreen_pv}{rot=cur\_pose.rot} ) \\ $\cdots$}};
        \node[rectangle, draw, rounded corners=2pt, text width=3.75cm, below right](reply3)at([xshift=1em]user3.north east)
        {\small\textsf{Please stack the blocks corner-to-corner with required angle, e.g., 45 degree.}};
        
        \node[rectangle, draw, fill=white, rounded corners=2pt, text width=3.7cm, below right](code4)at([xshift=1em]agent4.north east)
        {\scriptsize \textsf{\textcolor{RoyalBlue}{def} stack\_blocks(blocks, \textcolor{ffgreen_pv}{start\_pose}): \\ $\cdots$ \\ \textcolor{ffgreen_pv}{cur\_pose = start\_pose} \\ for $\cdots$}};
        \node[rectangle, draw, rounded corners=2pt, text width=3.75cm, below right](reply4)at([xshift=1em]user4.north east)
        {\small\textsf{Please stack blocks at a given place, e.g., table center.}};
        \node[rectangle, draw, rounded corners=2pt, text width=2.2cm, below right](reply5)at([xshift=1em]user5.north east){\small\textsf{Bravo! This is what I want.}};
        \node[fill=white, right](cdot1)at([xshift=1.5cm]img4.east){\Large $\bm\cdots$};
        \draw[->, draw=ffblue!80, line width=2pt](code1.south) to[out=-90, in=180] (img2.west);
        \draw[->, draw=ffblue!80, line width=2pt](code2.south) to[out=-90, in=180] (img3.west);      
        \draw[->, draw=ffblue!80, line width=2pt](code3.south) to[out=-90, in=180] (img4.west);
        \draw[->, draw=ffblue!80, line width=2pt]([xshift=-1cm]code4.south) to[out=-90, in=180] (cdot1.west);

        \node[above right](phase2)at(-2cm, -6.75cm){\Large\textcolor{ffgreen_pv}{\textbf{\textsf{Phase II: Lifelong Learning-oriented Capability Extension}}}};
        \draw[ffgreen_pv, rounded corners=4pt, line width=2pt]
            (-2cm,-6.75cm) -- (27cm,-6.75cm) -- (27cm,-13.75cm) -- (-2,-13.75cm) -- cycle;
        \node[inner sep=0, label={[text width=3cm, align=right, label distance=-2pt]-180:\scriptsize \textsf{stack small2big \\ $(l_1, s_{10})$}}](img6)at(2.5cm, -8cm){\includegraphics[width=4.5em]{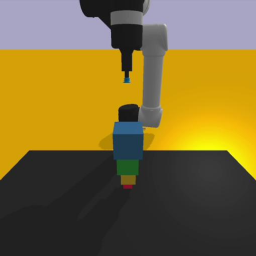}};
        \node[inner sep=0, label={[text width=3cm, align=right, label distance=-2pt]-180:\scriptsize \textsf{stack zigzag tower \\ $(l_2, s_{20})$}}, below](img7)at(img6.south){\includegraphics[width=4.5em]{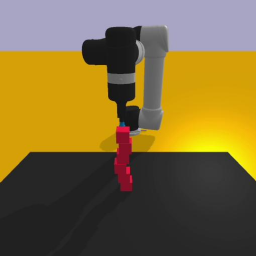}};
        \node[inner sep=0, label={[text width=3cm, align=right, label distance=-2pt]-180:\scriptsize \textsf{stack two towers by color \\ $(l_i, s_{i0})$}}, below](img8)at(img7.south){\includegraphics[width=4.5em]{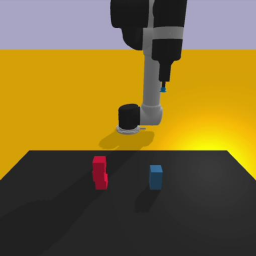}};
        \node[inner sep=0, label={[text width=3cm, align=right, label distance=-2pt]-180:\scriptsize \textsf{make cross around blue block \\ $(l_j, s_{j0})$}}, right](img9)at([xshift=10cm]img6.east){\includegraphics[width=4.5em]{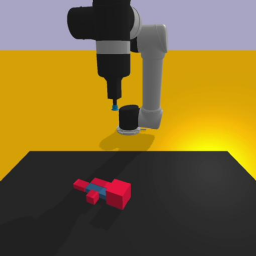}};
        \node[inner sep=0, label={[text width=3cm, align=right, label distance=-2pt]-180:\scriptsize \textsf{build a \{i * k\} pyramid \\ $(l_k, s_{k0})$}}, below](img10)at(img9.south){\includegraphics[width=4.5em]{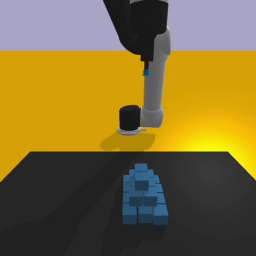}};
        \node[inner sep=0, label={[text width=3cm, align=right, label distance=-2pt]-180:\scriptsize \textsf{arrange in a circle \\ $(l_n, s_{n0})$}}, below](img11-1)at(img10.south){\includegraphics[width=4.5em]{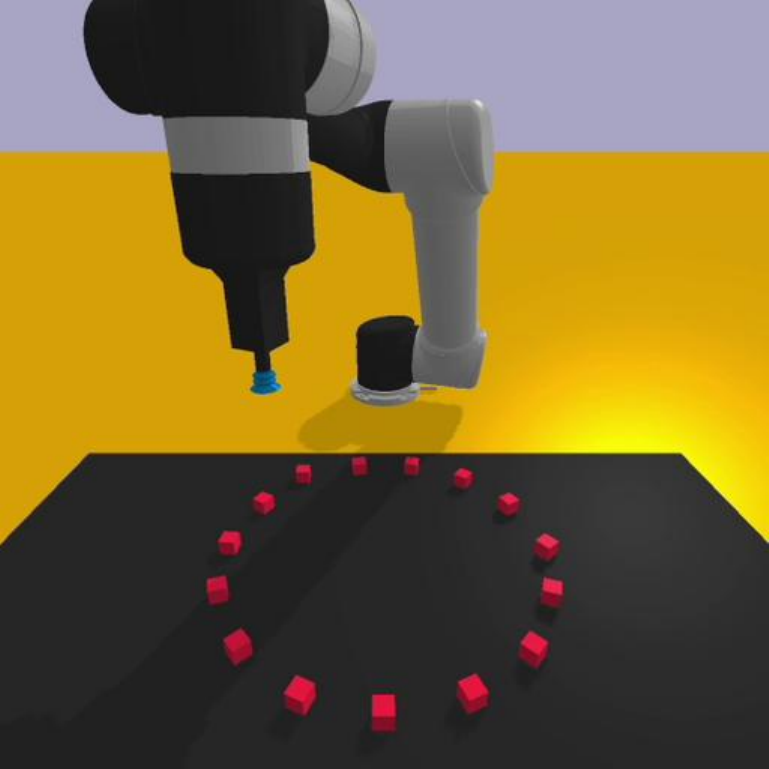}};
        \node[draw, circle, clip, minimum width=1.5em, inner sep=-.5em, label={[label distance=-1pt]-90:\small \textsf{Agent}}, right](agent5)at([xshift=1.5cm]img7.east){\includegraphics[width=3em]{imgs/openai.jpg}};
        \node[draw=fflightgreen, fill=none, circle, line width=2pt, minimum width=3em]at(agent5.center){};
        \node[draw, circle, clip, minimum width=1.5em, inner sep=-.5em, label={[label distance=-1pt]-90:\small \textsf{Agent}}, right](agent6)at([xshift=1.5cm]img10.east){\includegraphics[width=3em]{imgs/openai.jpg}};
        \node[draw=fflightgreen, fill=none, circle, line width=2pt, minimum width=3em]at(agent6.center){};
        \node[circle, clip, minimum width=.5em, inner sep=-.1em, label={[label distance=-1pt]-90:\small \textsf{User}}](user5)at(-1cm, -12.75cm){\includegraphics[width=2em]{imgs/user.png}};
        \node[draw=fforange_pv, fill=none, circle, line width=2pt, minimum width=3em]at(user5.center){};
        \node[circle, clip, minimum width=.5em, inner sep=-.1em, label={[label distance=-1pt]-90:\small \textsf{User}}, right](user6)at([xshift=1cm]agent5.east){\includegraphics[width=2em]{imgs/user.png}};
        \node[draw=fforange_pv, fill=none, circle, line width=2pt, minimum width=3em]at(user6.center){};
        \node[circle, clip, minimum width=.5em, inner sep=-.1em, label={[label distance=-1pt]-90:\small \textsf{User}}, right](user7)at([xshift=1cm]agent6.east){\includegraphics[width=2em]{imgs/user.png}};
        \node[draw=fforange_pv, fill=none, circle, line width=2pt, minimum width=3em]at(user7.center){};
        \node[rounded rectangle, draw=ffblue, rounded rectangle arc length=180, minimum width=4cm, minimum height=2cm, line width=2pt](shape1)at([xshift=3cm, yshift=-.1cm]img7.east){};
        \node[above]at([yshift=-.05cm]shape1.north){\textcolor{ffblue}{\small \textbf{\textsf{User-designed curriculum}}}};
        \node[double arrow, fill=ffblue!80, minimum width=.1cm, minimum height=1cm, line width=1pt]at([xshift=3cm]img7.east){};
        \node[inner sep=0pt, label={[text width=2cm, align=left, label distance=0cm]0:\small \textsf{Base skill:\\ stack\_blocks}}, above right](baseskill1)at([xshift=-.5cm, yshift=.5cm]shape1.north west){\includegraphics[width=1cm]{imgs/code.png}};
        \node[rounded rectangle, draw=ffblue, rounded rectangle arc length=180, minimum width=4cm, minimum height=2cm, line width=2pt](shape2)at([xshift=3cm, yshift=-.1cm]img10.east){};
        \node[above]at([yshift=-.05cm]shape2.north){\textcolor{ffblue}{\small \textbf{\textsf{User-designed curriculum}}}};
        \node[inner sep=0pt, label={[text width=2cm, align=left, label distance=0cm]0:\small \textsf{Base skill:\\ move\_to\_reference}}, above right](baseskill2)at([xshift=-.5cm, yshift=.5cm]shape2.north west){\includegraphics[width=1cm]{imgs/code.png}};
        \node[double arrow, fill=ffblue!80, minimum width=.1cm, minimum height=1cm, line width=1pt]at([xshift=3cm]img10.east){};
        \node[label={[label distance=-.25cm]-90:\small \textsf{Memory}}](memory)at(25cm, -10cm){\includegraphics[width=2.5cm]{imgs/brain_database.png}};
        \node[above left](example0)at([xshift=.2cm, yshift=.9cm]memory.west){\includegraphics[width=1.5cm]{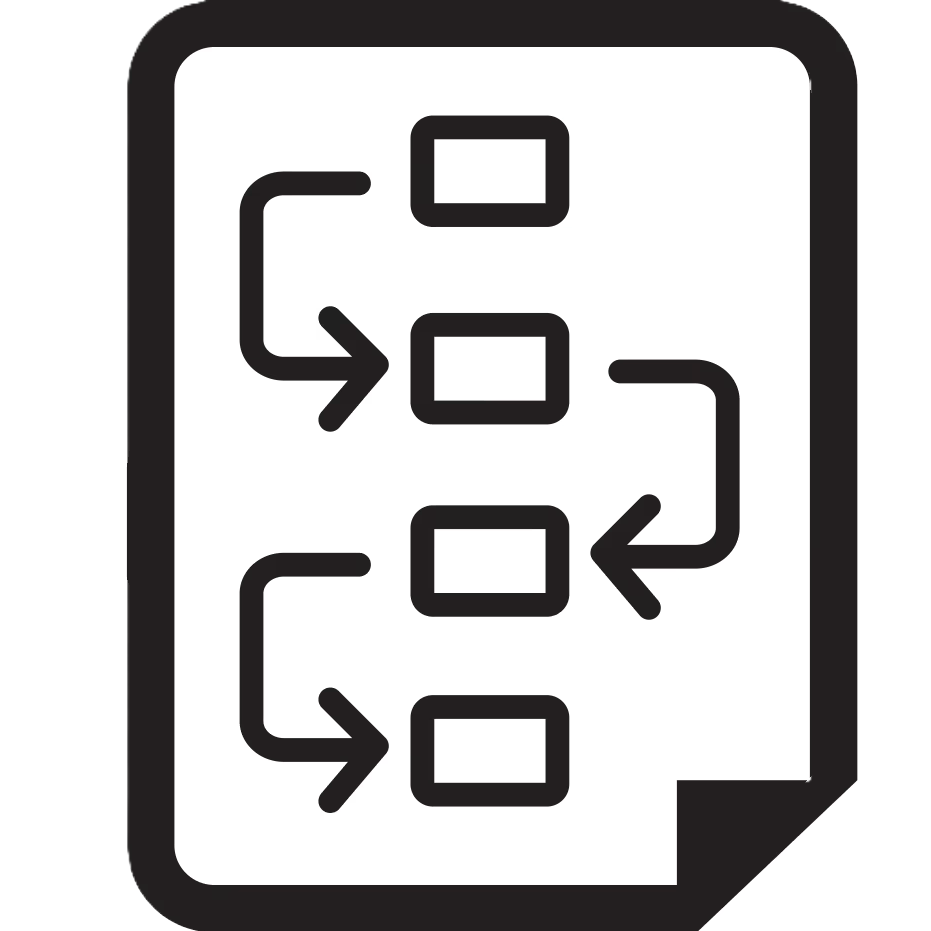}};
        \node[label={[text width=1.5cm, align=center, label distance=0cm]-90:\small \textsf{Examples \\ $(l ,c)\in\mathcal{E}$}}, above left](example)at([yshift=.7cm]memory.west){\includegraphics[width=1.5cm]{imgs/example.png}};
        \node[above left](example00)at([xshift=-.2cm, yshift=.5cm]memory.west){\includegraphics[width=1.5cm]{imgs/example.png}};
        
        \node[below left](code0)at([xshift=.2cm, yshift=-.5cm]memory.west){\includegraphics[width=1.5cm]{imgs/code.png}};
        \node[label={[text width=1.5cm, align=center, label distance=0cm]-90:\small \textsf{Skills \\ $z\in\mathcal{Z}$}}, below left](code)at([yshift=-.7cm]memory.west){\includegraphics[width=1.5cm]{imgs/code.png}};
        \node[below left](code00)at([xshift=-.2cm, yshift=-.9cm]memory.west){\includegraphics[width=1.5cm]{imgs/code.png}};
        \node[above right](prompt)at([xshift=1em]user5.east){\textcolor{fforange_pv}{\textbf{\textsf{Meta-prompt}}}};
        \node[rectangle, draw, rounded corners=2pt, text width=8cm, below right](reply5)at([xshift=1em]user5.east){You should try to preserve the previous functionality.};
        \node[]at([xshift=2cm]user6.east){\Large $\bm{\cdots}$};
        \node[right]at(img7.south east){\Large $\bm{\cdots}$};
        \node[right]at(img10.south east){\Large $\bm{\cdots}$};
        \draw[-{Triangle Cap []. Fast Triangle[]}, draw=ffyellow, line width=5pt, rounded corners=1pt] (example.east) --++ (0.3cm, 0.0cm) --++ (0.4cm, -0.4cm);
        \draw[-{Triangle Cap []. Fast Triangle[]}, draw=fflightgreen!80, line width=5pt, rounded corners=1pt] (code.east) --++ (0.3cm, 0.0cm) --++ (0.4cm, 0.4cm);
        \draw[-{Triangle Cap []. Fast Triangle[]}, draw=ffyellow, line width=5pt, rounded corners=1pt] ([xshift=-.15cm, yshift=2.75cm]memory.north) -- ([xshift=-.15cm, yshift=-.2cm]memory.north);
        \draw[-{Triangle Cap []. Fast Triangle[]}, draw=fflightgreen!80, line width=5pt, rounded corners=1pt] ([xshift=.15cm, yshift=2.75cm]memory.north) -- ([xshift=.15cm, yshift=-.2cm]memory.north);
        \draw[-{Triangle Cap []. Fast Triangle[]}, draw=fforange_pv!80, line width=5pt, rounded corners=1pt] (reply5.north) |- ([xshift=.5cm, yshift=.25cm]reply5.north) -| (shape1.south);
        \draw[-{Triangle Cap []. Fast Triangle[]}, draw=fforange_pv!80, line width=5pt, rounded corners=1pt] (reply5.north) |- ([xshift=.5cm, yshift=.25cm]reply5.north) -| (shape2.south);
        \draw[-, draw=ffyellow, line width=5pt, rounded corners=1pt] ([yshift=.15cm]shape1.east) --++ (1cm, 0.0cm);
        \draw[-, draw=fflightgreen!80, line width=5pt, rounded corners=1pt] ([yshift=-.15cm]shape1.east) --++ (1cm, 0.0cm);
        \draw[-{Triangle Cap []. Fast Triangle[]}, draw=ffyellow, line width=5pt, rounded corners=1pt] ([yshift=.15cm]shape2.east) --++ (.5cm, 0.0cm) -- ([xshift=-.5cm]example.west) -- (example.west);
        \draw[-{Triangle Cap []. Fast Triangle[]}, draw=fflightgreen!80, line width=5pt, rounded corners=1pt] ([yshift=-.15cm]shape2.east) --++ (.5cm, 0.0cm) -- ([xshift=-.5cm]code.west) -- (code.west);
        \draw[-{Triangle Cap []. Fast Triangle[]}, draw=ffblue!80, line width=5pt, rounded corners=1pt] (img6.east) --++ (.5cm, 0.0cm) --++ (.5cm, -.5cm);
        \draw[-{Triangle Cap []. Fast Triangle[]}, draw=ffblue!80, line width=5pt, rounded corners=1pt] (img7.east) --++ (.8cm, 0.0cm);
        \draw[-{Triangle Cap []. Fast Triangle[]}, draw=ffblue!80, line width=5pt, rounded corners=1pt] (img8.east) --++ (.5cm, 0.0cm) --++ (.5cm, .5cm);
        \draw[-{Triangle Cap []. Fast Triangle[]}, draw=ffblue!80, line width=5pt, rounded corners=1pt] (img9.east) --++ (.5cm, 0.0cm) --++ (.5cm, -.5cm);
        \draw[-{Triangle Cap []. Fast Triangle[]}, draw=ffblue!80, line width=5pt, rounded corners=1pt] (img10.east) --++ (.8cm, 0.0cm);
        \draw[-{Triangle Cap []. Fast Triangle[]}, draw=ffblue!80, line width=5pt, rounded corners=1pt] (img11-1.east) --++ (.5cm, 0.0cm) --++ (.5cm, .5cm);

        \node[above right](phase3)at(-2cm, -14.5cm){\Large\textcolor{fforange_pv}{\textbf{\textsf{Phase III: Task-specific-oriented Retrieval and Planning}}}};
        \draw[fforange_pv, rounded corners=4pt, line width=2pt]
            (-2cm,-14.5cm) -- (27cm,-14.5cm) -- (27cm,-21.5cm) -- (-2,-21.5cm) -- cycle;
        \node[draw, circle, clip, minimum width=1.5em, inner sep=-.5em, label={[label distance=-1pt]-90:\small \textsf{Agent}}](agent7)at(2.83cm,-16cm){\includegraphics[width=3em]{imgs/openai.jpg}};
        \node[draw=fflightgreen, fill=none, circle, line width=2pt, minimum width=3em]at(agent7.center){};
        \node[draw, circle, clip, minimum width=1.5em, inner sep=-.5em, label={[label distance=-1pt]-90:\small \textsf{Agent}}](agent8)at(12.5cm,-16cm){\includegraphics[width=3em]{imgs/openai.jpg}};
        \node[draw=fflightgreen, fill=none, circle, line width=2pt, minimum width=3em]at(agent8.center){};
        \node[draw, circle, clip, minimum width=1.5em, inner sep=-.5em, label={[label distance=-1pt]-90:\small \textsf{Agent}}](agent9)at(22.17cm,-16cm){\includegraphics[width=3em]{imgs/openai.jpg}};
        \node[draw=fflightgreen, fill=none, circle, line width=2pt, minimum width=3em]at(agent9.center){};
        \draw[-{Triangle Cap []. Fast Triangle[]}, draw=fflightgreen, line width=5pt, rounded corners=1pt] ([xshift=.15cm, yshift=-.5cm]memory.south) --++ (0cm, -3.3cm) --++ (-22.15cm, 0) -- ([xshift=.2cm]agent7.north);
        \draw[-{Triangle Cap []. Fast Triangle[]}, draw=fflightgreen, line width=5pt, rounded corners=1pt] ([xshift=.15cm, yshift=-.5cm]memory.south) --++ (0cm, -3.3cm) --++ (-12.49cm, 0) -- ([xshift=.2cm]agent8.north);
        \draw[-{Triangle Cap []. Fast Triangle[]}, draw=fflightgreen, line width=5pt, rounded corners=1pt] ([xshift=.15cm, yshift=-.5cm]memory.south) --++ (0cm, -3.3cm) --++ (-2.83cm, 0) -- ([xshift=.2cm]agent9.north);
        \draw[-{Triangle Cap []. Fast Triangle[]}, draw=ffyellow, line width=5pt, rounded corners=1pt] ([xshift=-.15cm, yshift=-.5cm]memory.south) --++ (0cm, -3cm) --++ (-22.2cm, 0) -- ([xshift=-.2cm]agent7.north);
        \draw[-{Triangle Cap []. Fast Triangle[]}, draw=ffyellow, line width=5pt, rounded corners=1pt] ([xshift=-.15cm, yshift=-.5cm]memory.south) --++ (0cm, -3cm) --++ (-12.54cm, 0) -- ([xshift=-.2cm]agent8.north);
        \draw[-{Triangle Cap []. Fast Triangle[]}, draw=ffyellow, line width=5pt, rounded corners=1pt] ([xshift=-.15cm, yshift=-.5cm]memory.south) --++ (0cm, -3cm) --++ (-2.88cm, 0) -- ([xshift=-.2cm]agent9.north);

        \node[fill=white, text width=2cm, align=center, below]at([yshift=-1cm]memory.south){\textcolor{ffgreen_pv}{\textbf{\textsf{Retrival \\ Generation}}}};

        \node[]at([yshift=-4.7cm]agent7.south){\large \textcolor{fforange_pv}{\textbf{\textsf{Challenging ultra long-horizon tasks}}}};
        \node[label={[label distance=-.1cm]-90:\small \textsf{Task Plans}}, below](plan1)at([yshift=-1cm]agent7.south){\includegraphics[width=1.5cm]{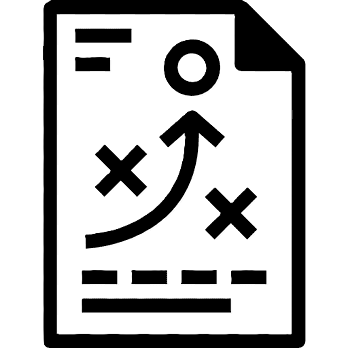}};
        \node[below left, label={[label distance=-5pt]-90:\small \textsf{write ``hello''}}](img11)at([xshift=-.5cm, yshift=.5cm]agent7.west){\includegraphics[width=6em]{imgs/hello-letters2.png}};
        \node[below right, label={[label distance=-5pt]-90:\small \textsf{make smiley face}}] (img12)at([xshift=.5cm, yshift=.5cm]agent7.east) {\includegraphics[width=6em]{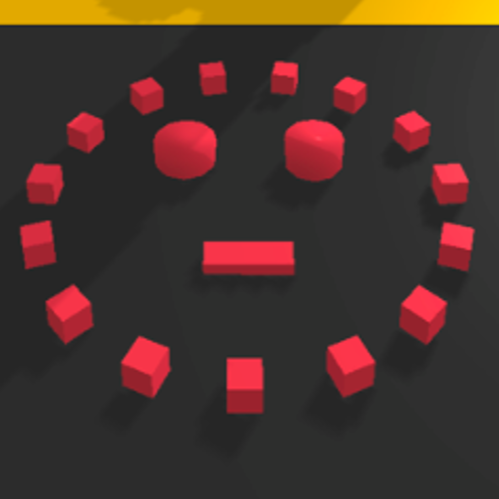}};
        \node[below, label={[label distance=-5pt]-90:\small \textsf{build house}}] (img13)at([yshift=-1em]img11.south) {\includegraphics[width=6em]{imgs/build-house2.png}};
        \node[below, label={[label distance=-5pt]-90:\small \textsf{stack jenga tower}}] (img14)at([yshift=-1em]img12.south) {\includegraphics[width=6em]{imgs/jenga-tower2.png}};

        \node[]at([yshift=-4.7cm]agent8.south){\large \textcolor{fforange_pv}{\textbf{\textsf{Transfering to cross-embodiment}}}};
        \node[label={[label distance=-.1cm]-90:\small \textsf{Task Plans}}, below](plan2)at([yshift=-1cm]agent8.south){\includegraphics[width=1.5cm]{imgs/plan.png}};
        \node[below left, label={[label distance=-5pt]-90:\small \textsf{place the kettle}}](img15)at([xshift=-.5cm, yshift=.5cm]agent8.west){\includegraphics[width=6em]{imgs/kitchen-1.pdf}};
        \node[below right, label={[label distance=-5pt]-90:\small \textsf{use hammer}}] (img16)at([xshift=.5cm, yshift=.5cm]agent8.east) {\includegraphics[width=6em]{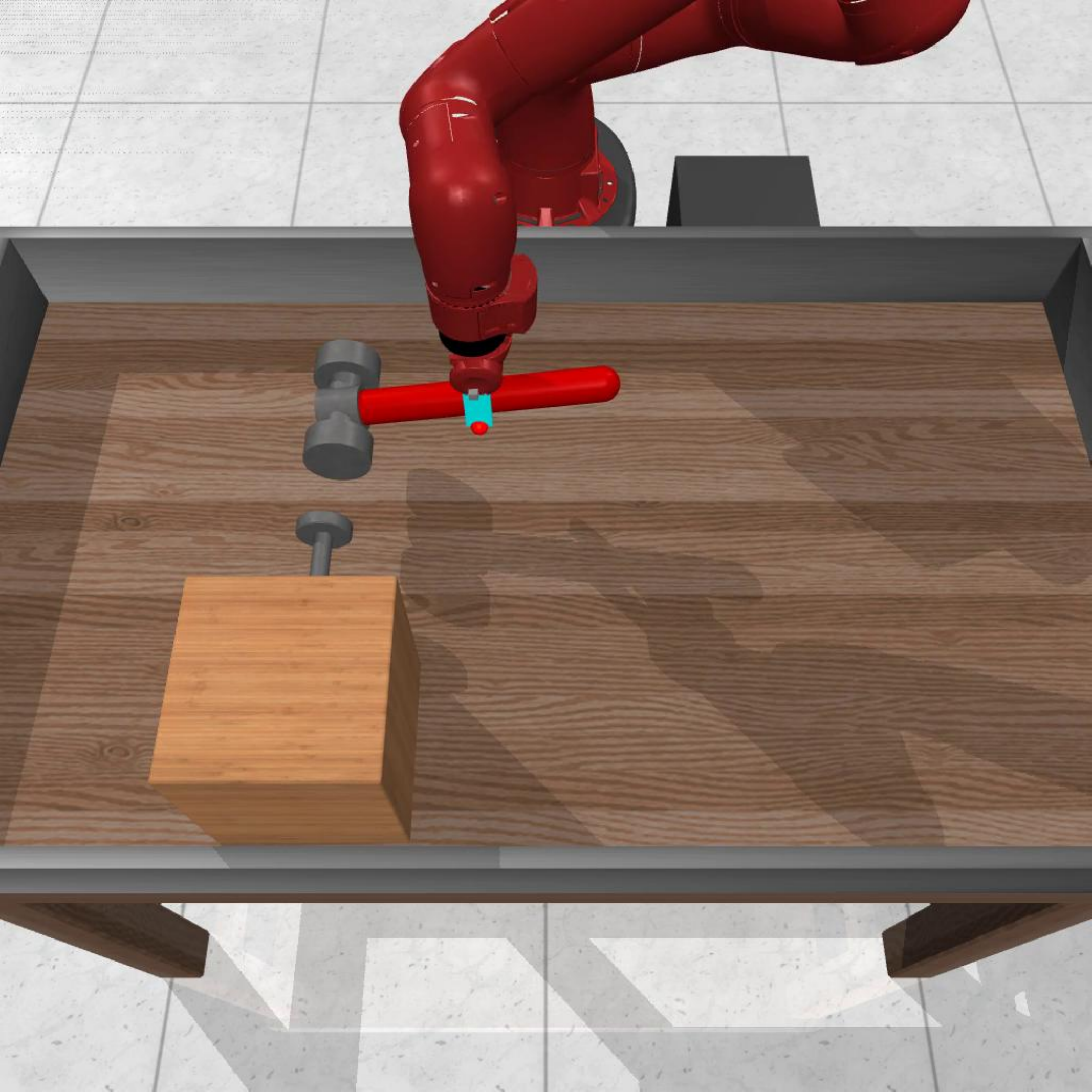}};
        \node[below, label={[label distance=-5pt]-90:\small \textsf{open cabinet}}] (img17)at([yshift=-1em]img15.south) {\includegraphics[width=6em]{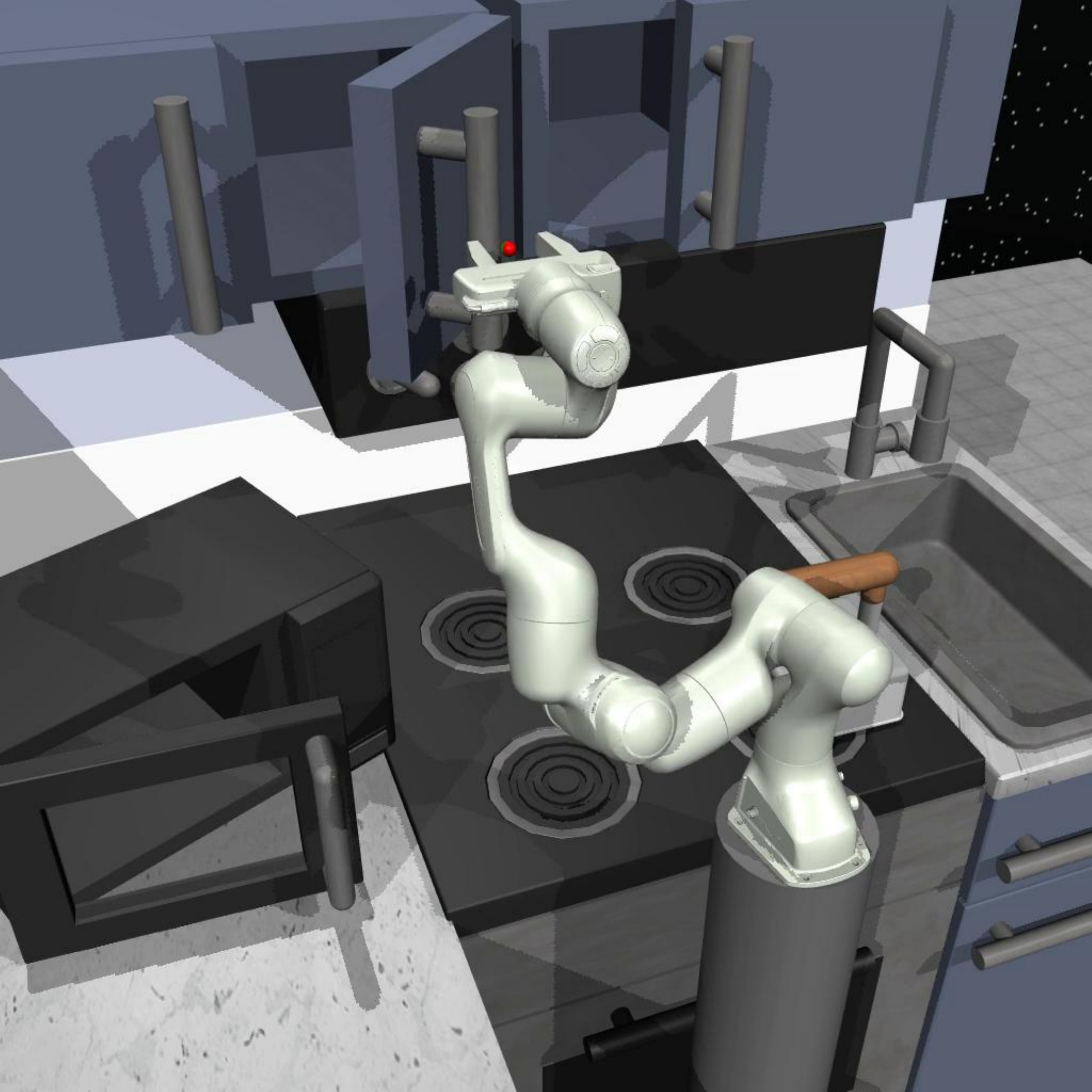}};
        \node[below, label={[label distance=-5pt]-90:\small \textsf{use moka pot}}] (img18)at([yshift=-1em]img16.south) {\includegraphics[width=6em]{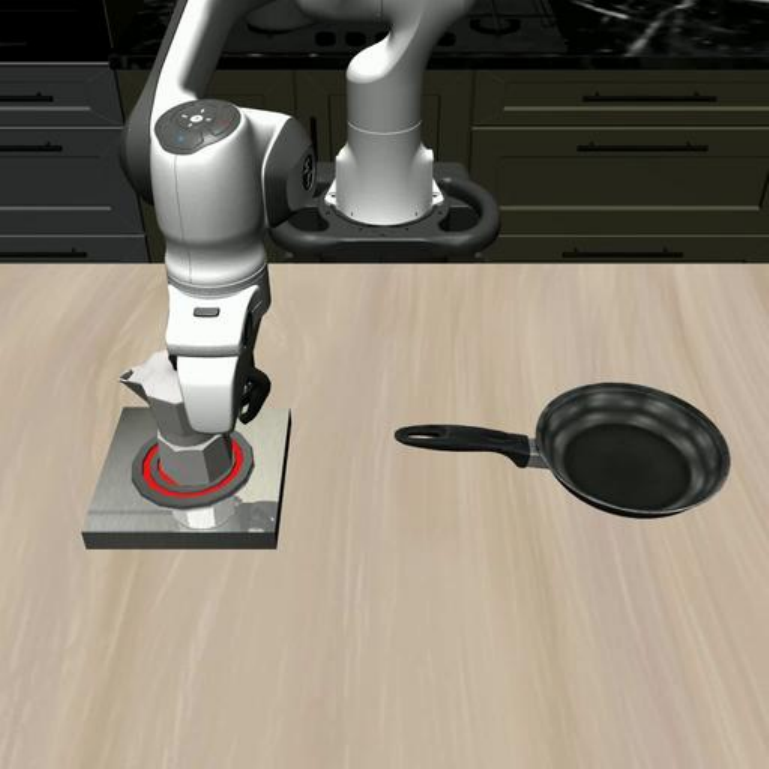}};

        \node[]at([yshift=-4.7cm]agent9.south){\large \textcolor{fforange_pv}{\textbf{\textsf{Real-world scenarios validation}}}};
        \node[label={[label distance=-.1cm]-90:\small \textsf{Task Plans}}, below](plan1)at([yshift=-1cm]agent9.south){\includegraphics[width=1.5cm]{imgs/plan.png}};
        \node[below left, label={[label distance=-5pt]-90:\small \textsf{build house}}](img19)at([xshift=-.5cm, yshift=.5cm]agent9.west){\includegraphics[width=6em]{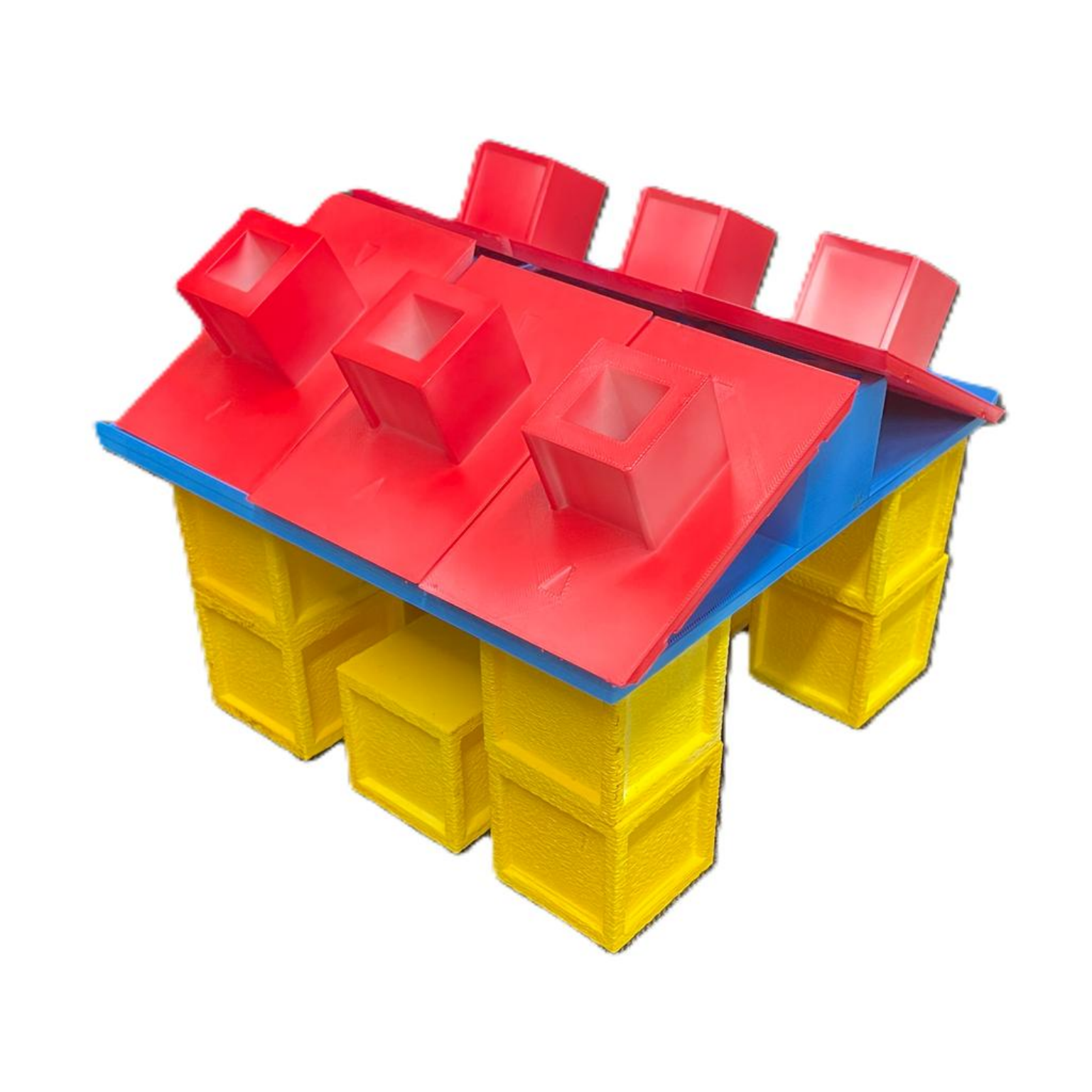}};
        \node[below right, label={[label distance=-5pt]-90:\small \textsf{build temple}}] (img20)at([xshift=.5cm, yshift=.5cm]agent9.east) {\includegraphics[width=6em]{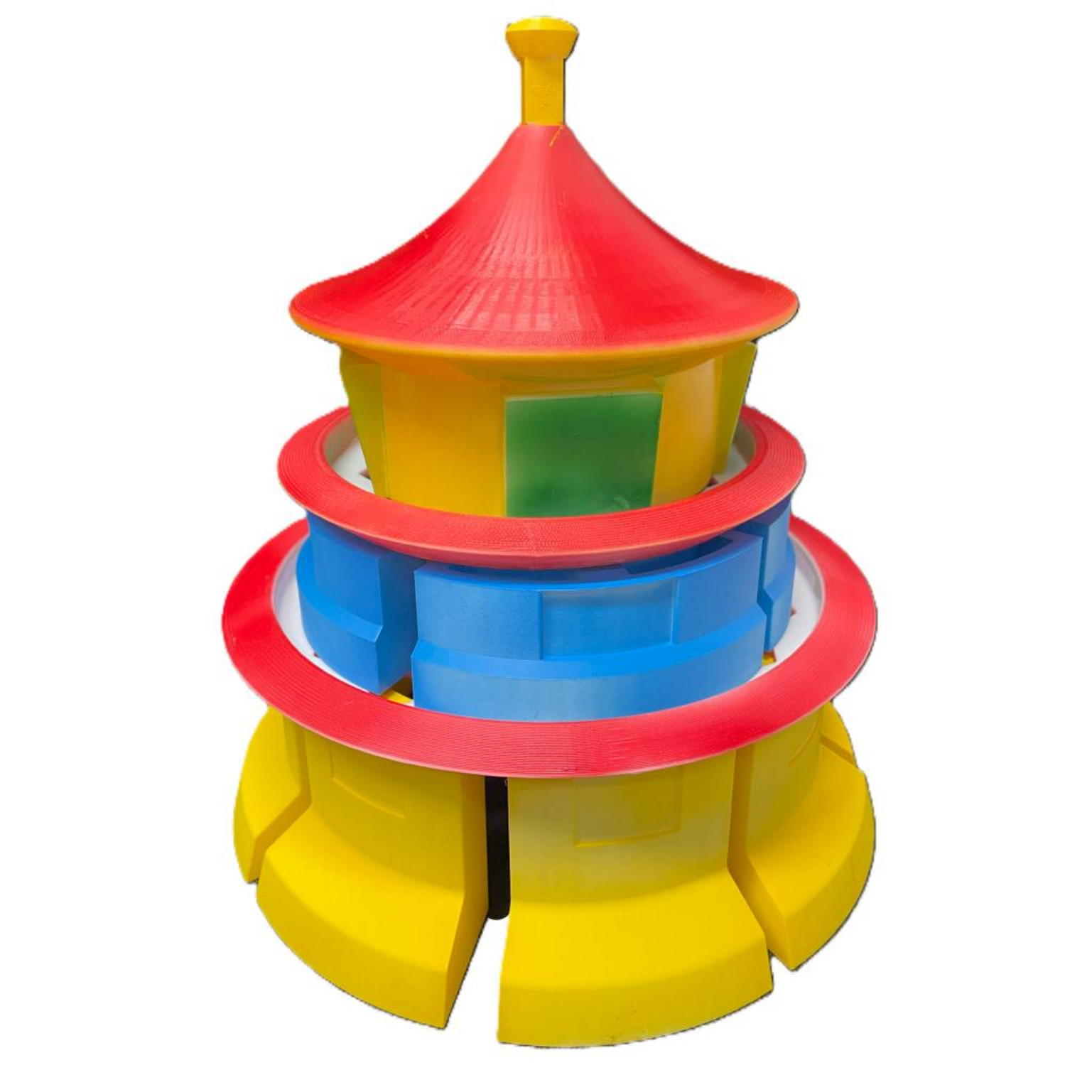}};
        \node[below, label={[label distance=-5pt]-90:\small \textsf{stack jenga tower}}] (img21)at([yshift=-1em]img19.south) {\includegraphics[width=6em]{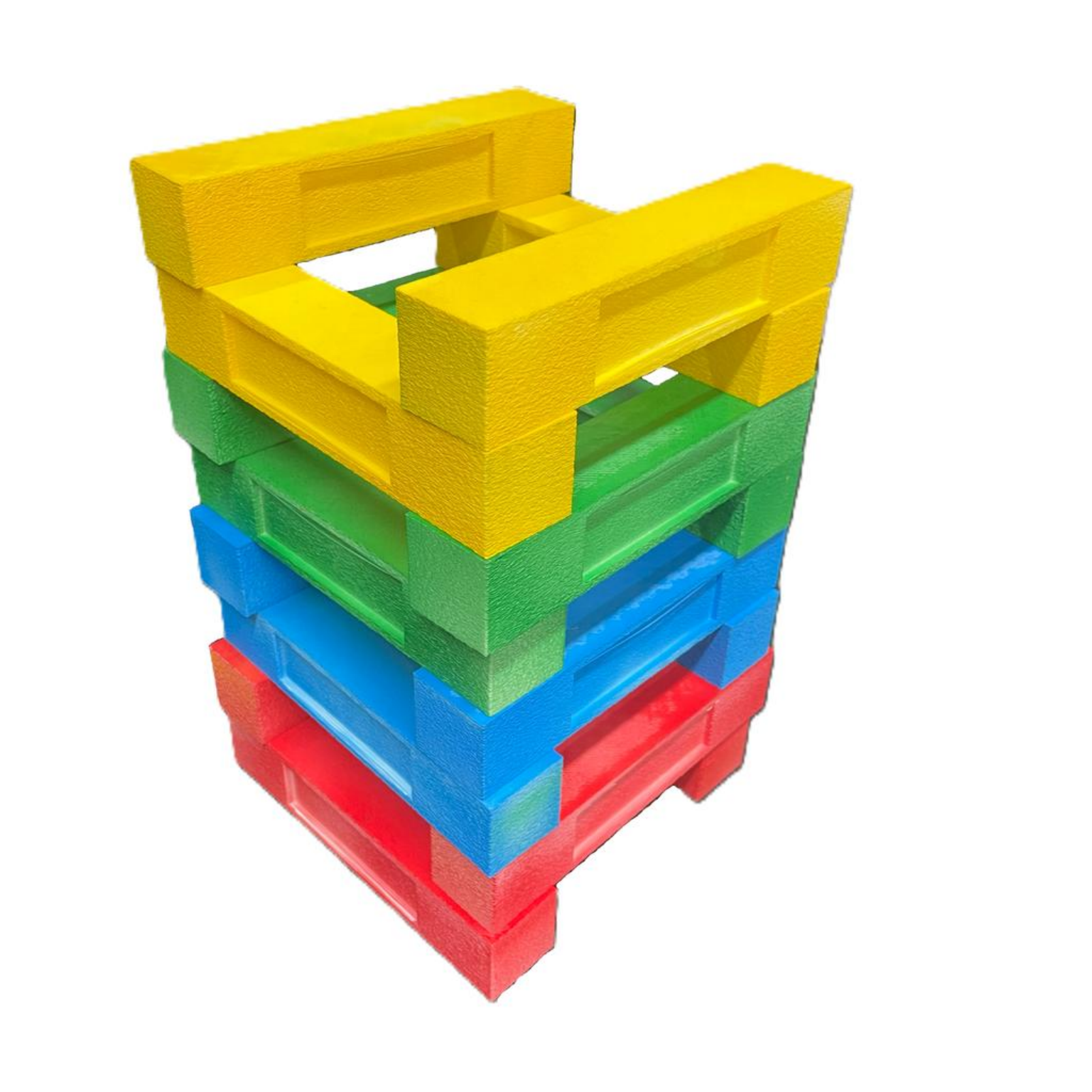}};
        \node[below, label={[label distance=-5pt]-90:\small \textsf{build human face}}] (img22)at([yshift=-1em]img20.south) {\includegraphics[width=6em]{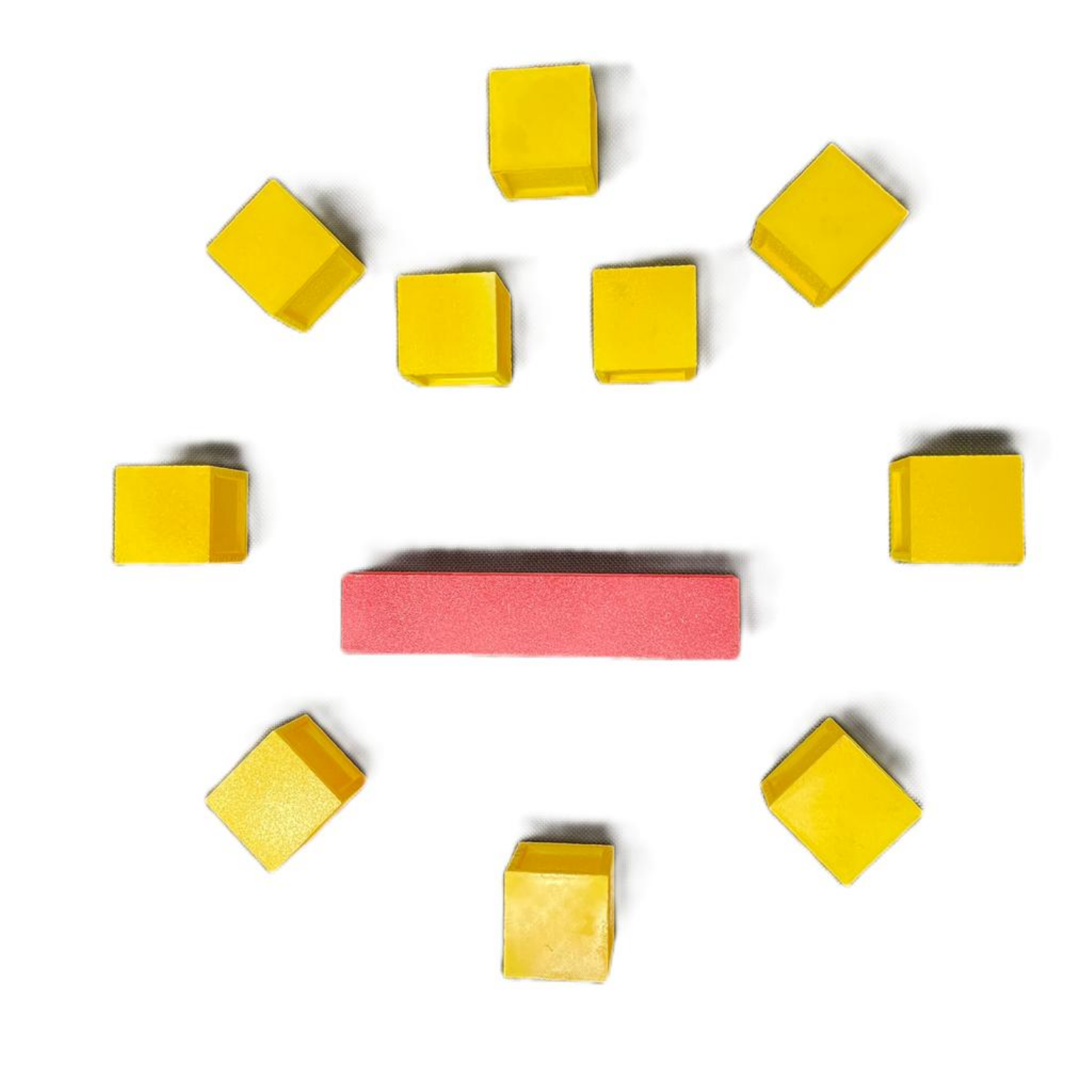}};
        
    \end{tikzpicture}
    }
    \vskip -.1in
    \caption{Structure of the proposed human-in-the-loop lifelong skill learning pipeline.}
    \label{fig:pipeline}
\end{figure*}

\subsection{Preliminaries} 

Code generation for robotic manipulation can be modelled as using an LLM $f$ to map a natural language instruction $l$ to a robot control code $c$, i.e., $f(l) = c$, where $f$ is termed a Language Model Program (LMP) in prior work \cite{liang2023code}. 
Each code $c$ is achieved by multiple behaviours $b$; thus, $f$ can also be viewed as a mapping from $l$ to $b$. 
A behaviour $b \in \mathcal{B}$ denotes an evaluable semantic motion or outcome produced during robot execution, such as picking up an object.
In our work, we argue that behaviours $b$ requiring no variation can be encapsulated as \textbf{\textit{skills}} that reliably elicit these behaviours. 
We denote the skill space as $\mathcal{Z}$ and state parameters as $s$. 
Importantly, a skill $z \in \mathcal{Z}$ may also build on other skills:

\textbf{Definition 1}: A \textbf{\textit{skill}} $z \in \mathcal{Z}$ is an executable parameterized function. Given state-dependent parameters $s$, its execution realizes a target behaviour $b \in \mathcal{B}$, denoted as $b = \mathrm{Exec}(z,s)$.

It is worth noting that although the task code $c$ generated by the agent represents a sequence of behaviours to accomplish a task, it is not equivalent to a skill. 
We define a \textbf{\textit{task}} $(l, s_0)$ by a natural language instruction $l$ and an initial environment state $s_0$, which together form the main interface between the user and the agent. 
When a task is successfully completed with user feedback, we extend this tuple to $(l, s_0, c, s_T)$, where $c$ is the task-specific code (often flat, process-oriented) and $s_T$ is the achieved final state. In practice, $c$ may call skills $z$ to complete the task, and its execution should produce an evaluable task result.
To align the skill output with the user’s desired behaviour $b$, the agent adjusts the function details under user guidance. 
We call this process \textbf{\textit{learning the skill}}. 
Importantly, ``learning'' does not refer to deep learning, instead follows a broader concept in computer science \cite{mitchell2006discipline}: 

\textbf{Definition 2}: An agent is said to \textbf{\textit{learn}} if it produces better behavioural responses $b$ to an instruction $l$, as evaluated by a human.

Based on these definitions, the learning pipeline should answer:
(1) How to encode human-provided feedback into skills that align with user preferences?
(2) How to extend skills through lifelong learning while enabling dynamic in-context adaptation?
(3) How to reuse and transfer learned skills to unseen long-horizon tasks?

\subsection{Preference-aligned Skill Acquisition}

Based on the above questions, we design a three-phase pipeline to preserve and utilize human-guided feedback for preference-aligned lifelong skill learning in long-horizon manipulation (Fig.~\ref{fig:pipeline}). The phases are: (1) preference-aligned skill acquisition, (2) lifelong capability extension, and (3) task-specific retrieval and planning.  

To learn a new skill, three elements must be clarified: (1) what skill to learn, (2) the initial task environment, and (3) the expected functionality or behavior.
Asking an LLM to infer all of this at once is impractical due to language ambiguity, so we adopt multi-turn interaction. 
The agent is initiated with a fixed set of skills $z \in \mathcal{Z}_0$, along with examples $e=(l, c) \in \mathcal{E}_0$ that demonstrate their use, enabling basic environment interaction.
The user first describes the skill in natural language, and the LLM invokes the ``skill\_parse'' API to generate a Python function header with suggested parameters (e.g., ``def stack\_blocks(blocks, start\_pose):'').
The user can accept or refine this skill definition. 
Once accepted, the LLM requests a base environment setup, and generates a corresponding task scenario. 
The complexity of the task is user-controlled, but overly complex tasks (over 6–8 steps) hinder skill acquisition when the agent’s ability is still limited. 
We therefore recommend two principles: (1) tasks should be simple, ideally completed within 1–4 primitives, and (2) tasks should allow flexibility so that one skill can adapt to diverse requirements. 
After the setup, the user and the LLM agent will work together to ``learn'' the skill. 
The agent generates candidate code, deploys it in simulation, and displays results in real time. 
The user then accepts, rejects, or provides free-form feedback to refine the skill. 
User tests multiple requirements within the same scenario to check if the skill and its parameters generalize. 
For example, in Fig.~\ref{fig:pipeline} Phase I, the user guides the agent to stack blocks under different conditions such as edge alignment, tower positioning, or rotation. During this phase, \textbf{the agent can freely modify implementation details under the reserved skill name}, with the objective of producing a functional, preference-aligned skill. 
This forms the foundation for later phases of lifelong skill learning and capability extension.

\subsection{Lifelong Learning-oriented Capability Extension}

Learning skills from a single task scenario is insufficient for complex ultra-long-horizon tasks, for two main reasons:  
(1) Simple skills cannot handle variations beyond their learning distribution. 
For example, a skill like ``stack\_blocks'', which stacks four blocks at a fixed pose, cannot generalize to tasks such as building a $\{i \times j \times k\}$ structure or stacking multiple towers by color or size.  
(2) Directly inserting learned skills into prompts risks hallucination, where LLMs may rewrite or fabricate code during long-horizon interactions, leading to catastrophic forgetting of existing skills.

Inspired by lifelong learning, we extend the pipeline with a user-designed curriculum that enables \textbf{bottom-up skill functionality expansion} (Fig.~\ref{fig:pipeline} Phase II). 
In this phase, the agent must: (1) solve more complex user-defined task variations, (2) preserve and extend skill functionality to unseen cases, and (3) store skills permanently for reuse. 
The learning process mirrors Phase I, with an added \textbf{meta-prompt that explicitly asks the agent to preserve prior functionality while adapting to new tasks}.  
Users should design tasks outside the original distribution and aligned with long-term goals, guiding the agent to acquire foundational skills for future complex tasks. 
The agent can either (1) create a new named skill that calls existing ones as nested functions, or (2) extend the current skill using modularized code (e.g., if-else or match-case) to avoid overwriting. 
After adapting to new tasks, the agent is re-evaluated on prior tasks; if performance is preserved, we can say that lifelong learning is achieved. 
To further prevent forgetting and hallucination, both learned skills $z$ and explored successful examples (including task instruction and task plan) $(l, c) \in \mathcal{E}$ are stored in \textbf{external memory} and retrieved during planning.

\newcommand{\D}{6}         
\newcommand{\U}{10} 
\newdimen\R
\R=3cm                   
\newdimen\L
\L=4cm                   
\newcommand{\A}{360/\D}    
\newcommand{\Start}{90}    
\newcommand{\radar}[2]{({\Start + (#1-1)*\A}:{#2*\the\R})}
\begin{figure*}[t!]
\vskip 0.05in
\centering
\resizebox{\textwidth}{!}{
\begin{tikzpicture}
    \path (0cm,0cm) coordinate (O);
    
    \foreach \X in {1,...,\D}{
    \draw (\Start+\X*\A:0) -- (\Start+\X*\A:\R);
    }
    \foreach \t in {0.1,0.2,...,1}{
    \draw[opacity=0.2] (\Start:\t*\R)
      \foreach \i in {1,...,\D}{
        -- (\Start+\i*\A:\t*\R)
      } -- cycle;
    }
    \foreach \t in {0.1,0.2,...,1}{
    \foreach \i in {1,...,\D}{
      \fill[opacity=0.5] (\Start+\i*\A:\t*\R) circle (1pt);
    }
    }

    \path (\Start+0*\A:\L) node[inner sep=0pt] (img-label1){\includegraphics[width=1cm]{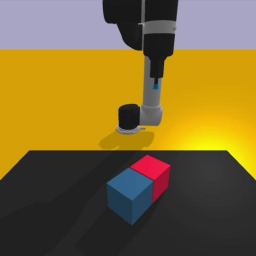}};
    \node[below=0cm of img-label1] {\footnotesize next to reference};
    \path (\Start+1*\A:\L) node[inner sep=0pt] (img-label2){\includegraphics[width=1cm]{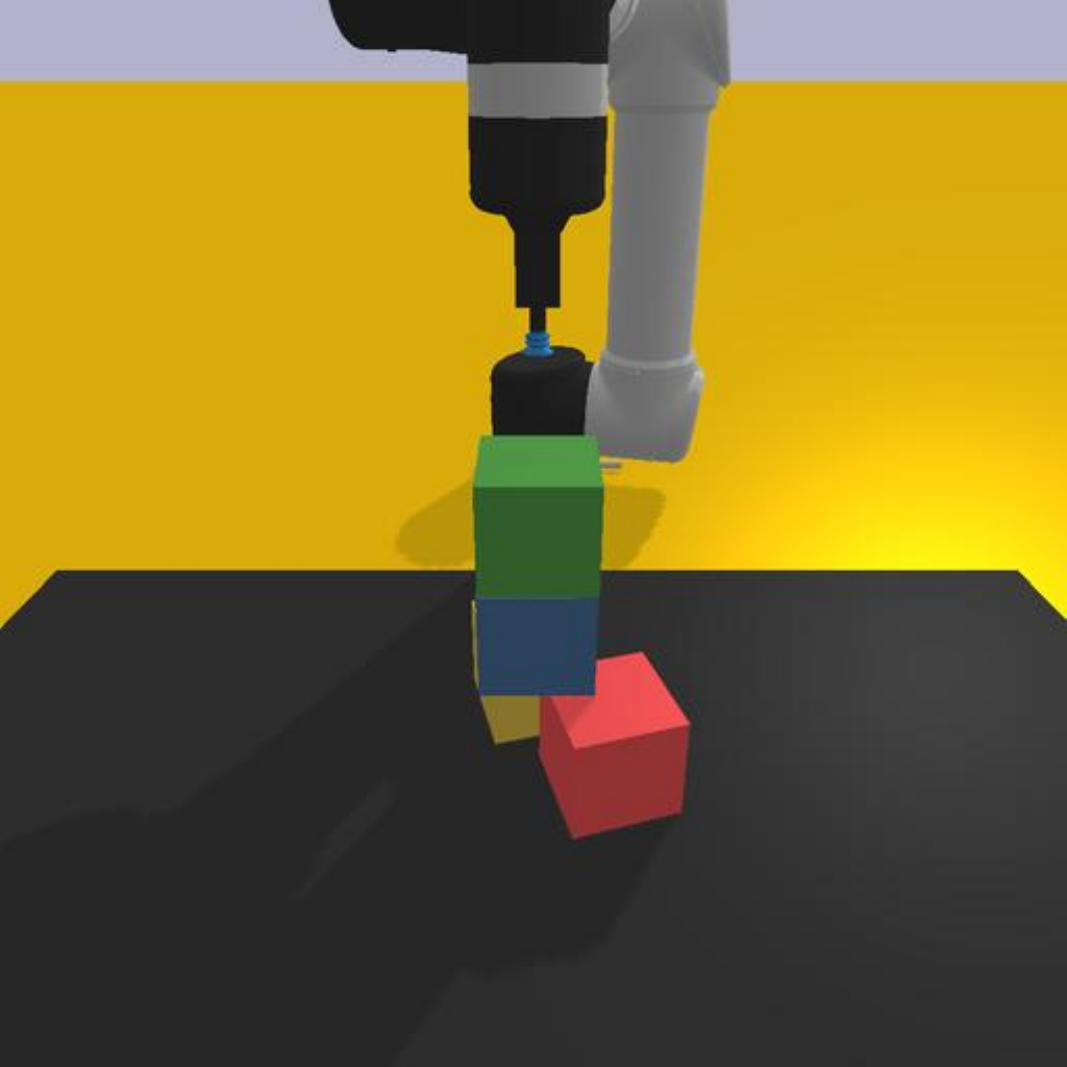}};
    \node[below=0cm of img-label2] {\footnotesize stack blocks};
    \path (\Start+2*\A:\L) node[inner sep=0pt] (img-label3){\includegraphics[width=1cm]{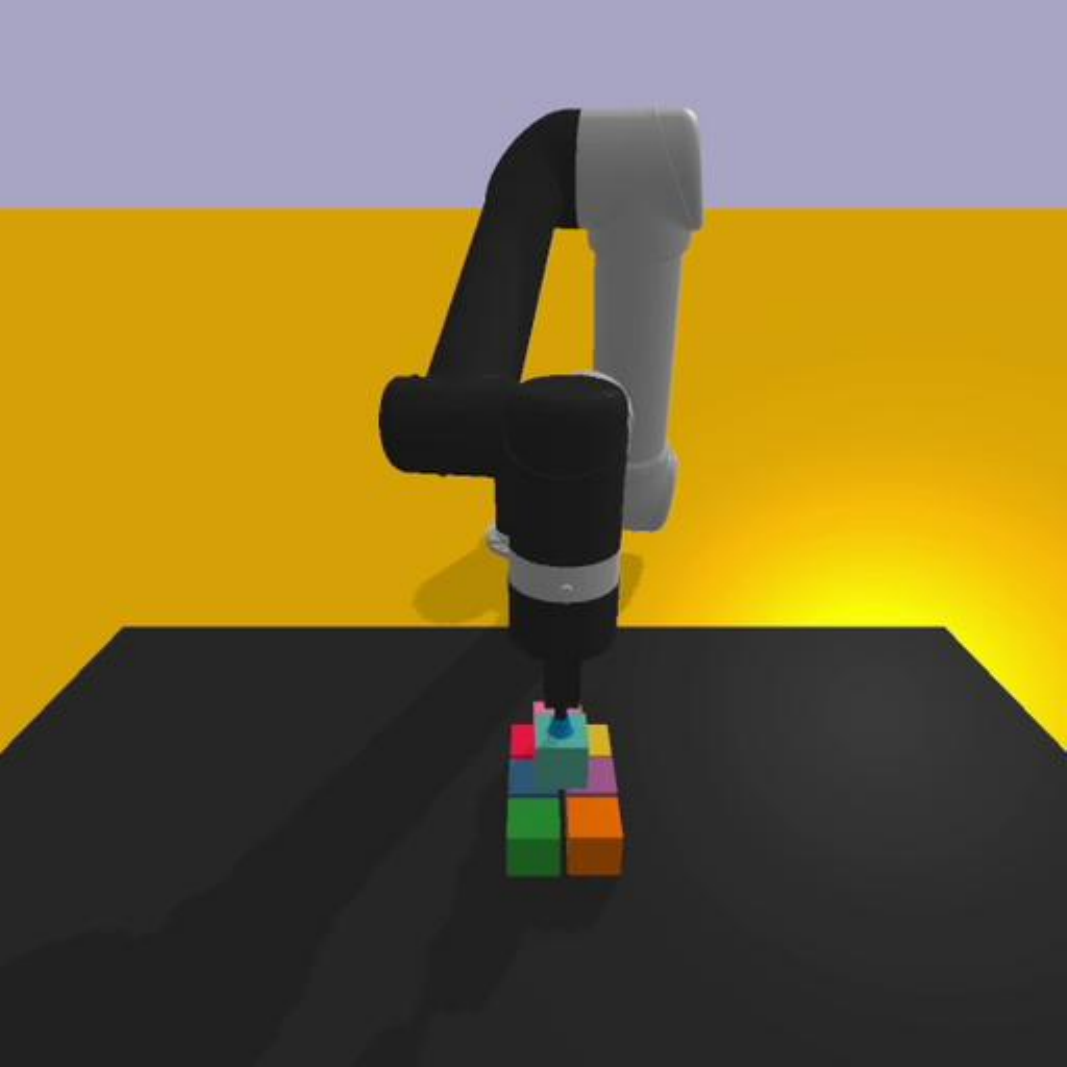}};
    \node[below=0cm of img-label3] {\footnotesize build \{i * j * k\} \{structure\}};
    \path (\Start+3*\A:\L) node[inner sep=0pt] (img-label4){\includegraphics[width=1cm]{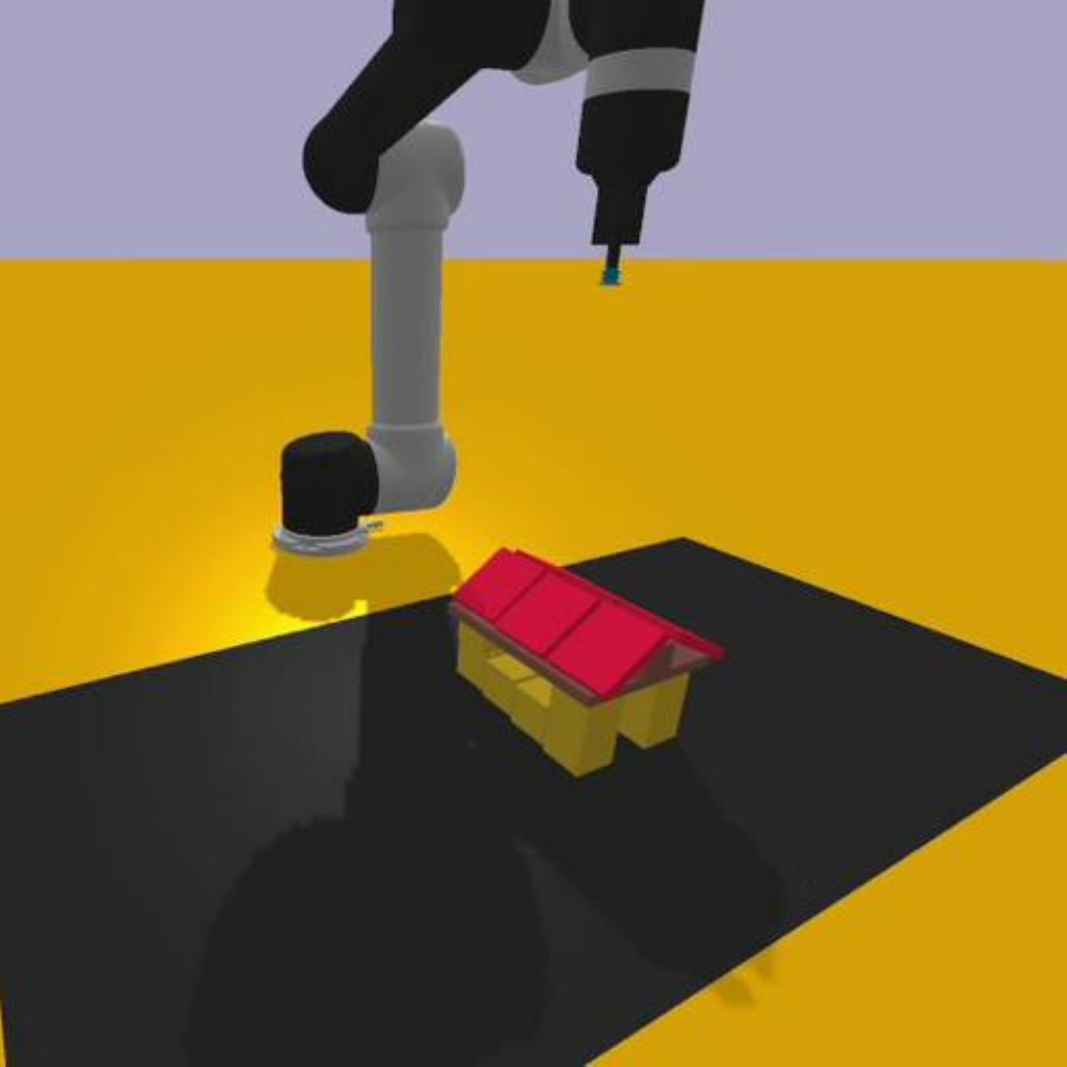}};
    \node[below=0cm of img-label4] {\footnotesize build a house};
    \path (\Start+4*\A:\L) node[inner sep=0pt] (img-label5){\includegraphics[width=1cm]{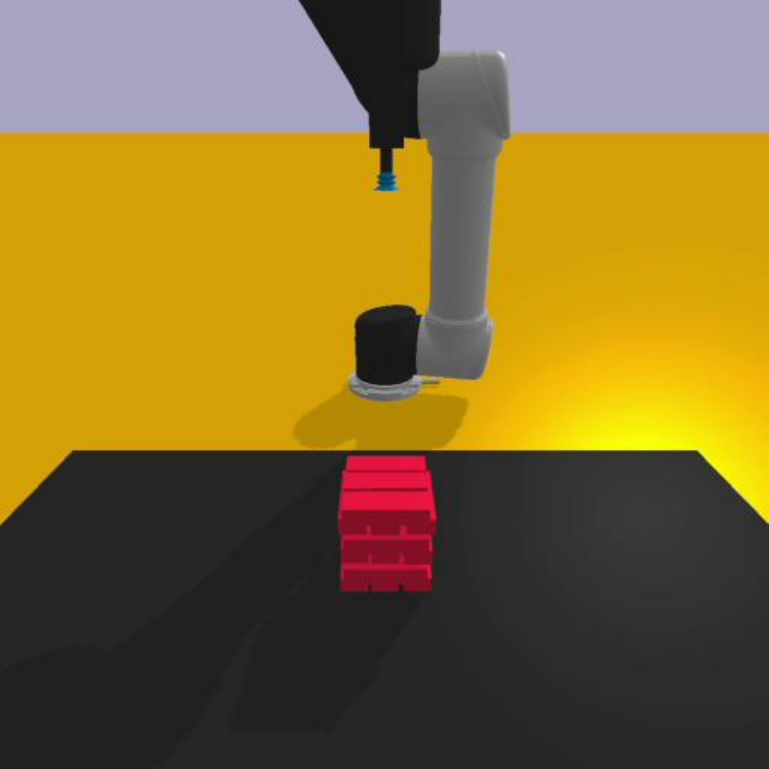}};
    \node[below=0cm of img-label5] {\footnotesize jenga tower};
    \path (\Start+5*\A:\L) node[inner sep=0pt] (img-label6){\includegraphics[width=1cm]{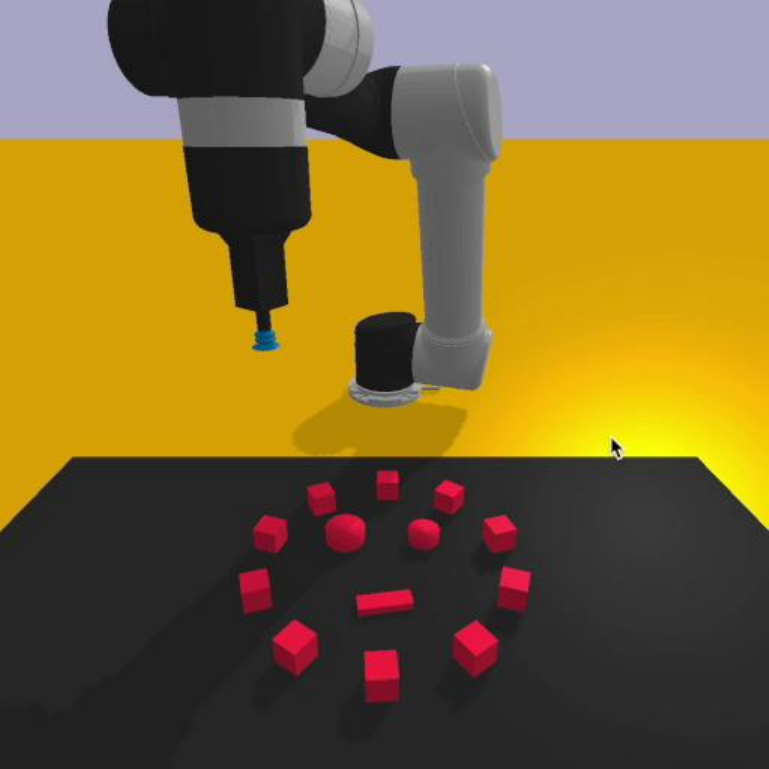}};
    \node[below=0cm of img-label6] {\footnotesize make a face};
    \draw[line width=2pt, draw=ffgreen_pv!80, draw opacity=1.0, fill=ffgreen_pv!80, fill opacity=0.1]
        \radar{1}{1.00} --
        \radar{2}{1.00} --
        \radar{3}{0.85} --
        \radar{4}{0.80} --
        \radar{5}{0.95} --
        \radar{6}{0.95} -- cycle;
    \draw[line width=2pt, draw=fflightgreen!80, draw opacity=1.0, fill=fflightgreen!80, fill opacity=0.1]
        \radar{1}{0.95} --
        \radar{2}{0.90} --
        \radar{3}{0.65} --
        \radar{4}{0.55} --
        \radar{5}{0.50} --
        \radar{6}{0.40} -- cycle;
    \draw[line width=2pt, draw=ffyellow!80, draw opacity=1.0, fill=yellow!80, fill opacity=0.1]
        \radar{1}{0.90} --
        \radar{2}{1.00} --
        \radar{3}{0.75} --
        \radar{4}{0.45} --
        \radar{5}{0.80} --
        \radar{6}{0.75} -- cycle;
    \draw[line width=2pt, draw=fforange_pv!80, draw opacity=1.0, fill=fforange_pv!80, fill opacity=0.1]
        \radar{1}{0.90} --
        \radar{2}{1.00} --
        \radar{3}{0.55} --
        \radar{4}{0.00} --
        \radar{5}{0.20} --
        \radar{6}{0.35} -- cycle;
    \draw[line width=2pt, draw=ffred!80, draw opacity=1.0, fill=ffred!80, fill opacity=0.1]
        \radar{1}{0.90} --
        \radar{2}{0.75} --
        \radar{3}{0.50} --
        \radar{4}{0.00} --
        \radar{5}{0.05} --
        \radar{6}{0.10} -- cycle;
    \draw[line width=2pt, draw=ffblue!80, draw opacity=1.0, fill=ffblue!80, fill opacity=0.1]
        \radar{1}{0.75} --
        \radar{2}{0.65} --
        \radar{3}{0.45} --
        \radar{4}{0.00} --
        \radar{5}{0.00} --
        \radar{6}{0.00} -- cycle;

    \node[fill=ffgreen_pv!80, label={[label distance=0pt]0:\small \textsf{\textbf{LYRA (Ours)}}}](1)at(5cm, 3cm){};
    \node[fill=fflightgreen!80, label={[label distance=0pt]0:\small \textsf{\textbf{LYRA (Ours) w/o memory}}}, below](2)at([yshift=-1cm]1.south){};
    \node[fill=ffyellow!80, label={[label distance=0pt]0:\small \textsf{\textbf{LYRA (Ours) w/ LLM feedback}}}, below](3)at([yshift=-1cm]2.south){};
    \node[fill=fforange_pv!80, label={[label distance=0pt]0:\small \textsf{\textbf{DAHLIA (GPT-4o)}}}, below](4)at([yshift=-1cm]3.south){};
    \node[fill=ffred_pv!80, label={[label distance=0pt]0:\small \textsf{\textbf{LoHoRavens (GPT-4o)}}}, below](5)at([yshift=-1cm]4.south){};
    \node[fill=ffblue!80, label={[label distance=0pt]0:\small \textsf{\textbf{CaP (GPT-4o)}}}, below](6)at([yshift=-1cm]5.south){};

    \node[inner sep=0pt](img1)at(12cm, 3cm){\includegraphics[width=3cm]{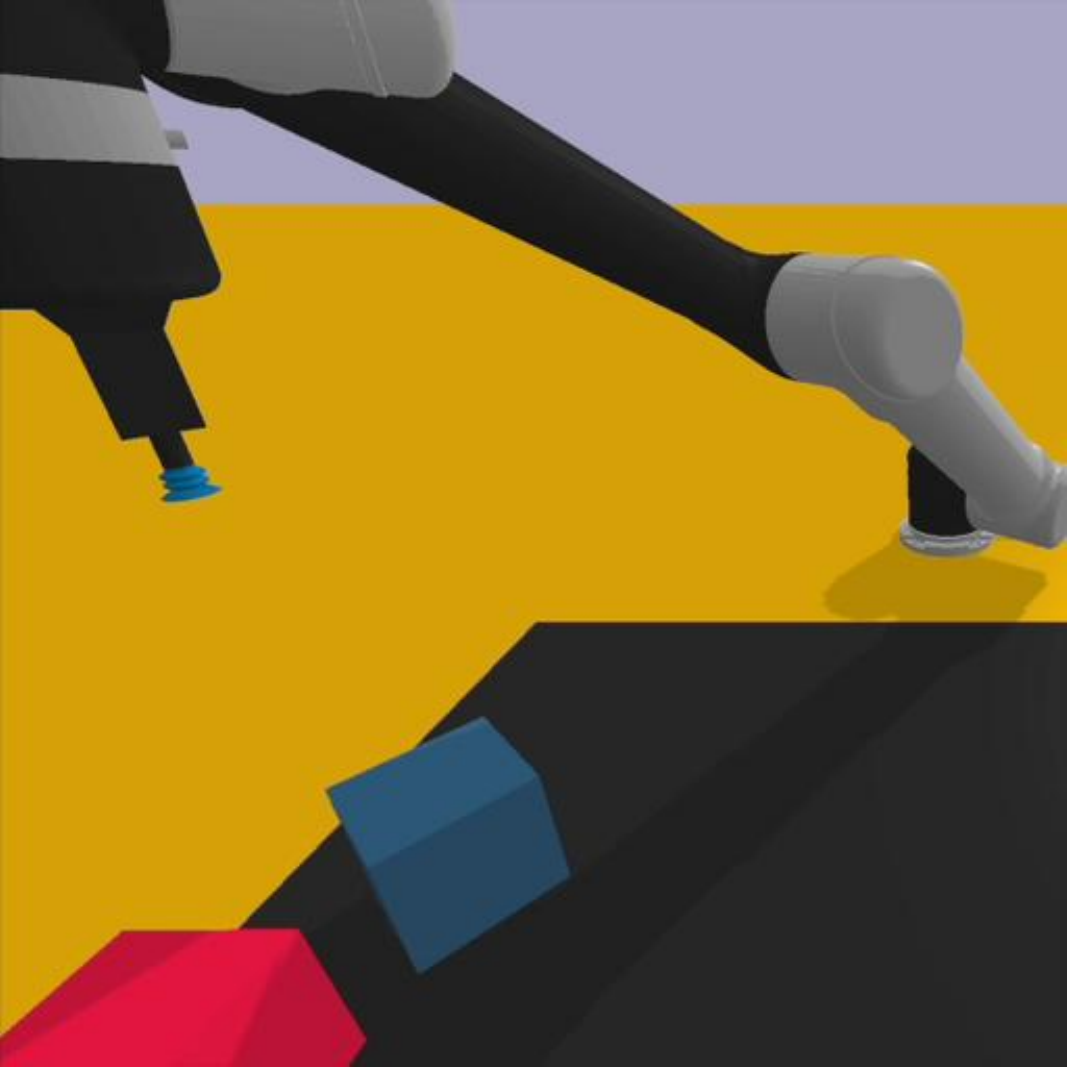}};
    \node[inner sep=0pt, right](img2)at([xshift=.1cm]img1.east){\includegraphics[width=3cm]{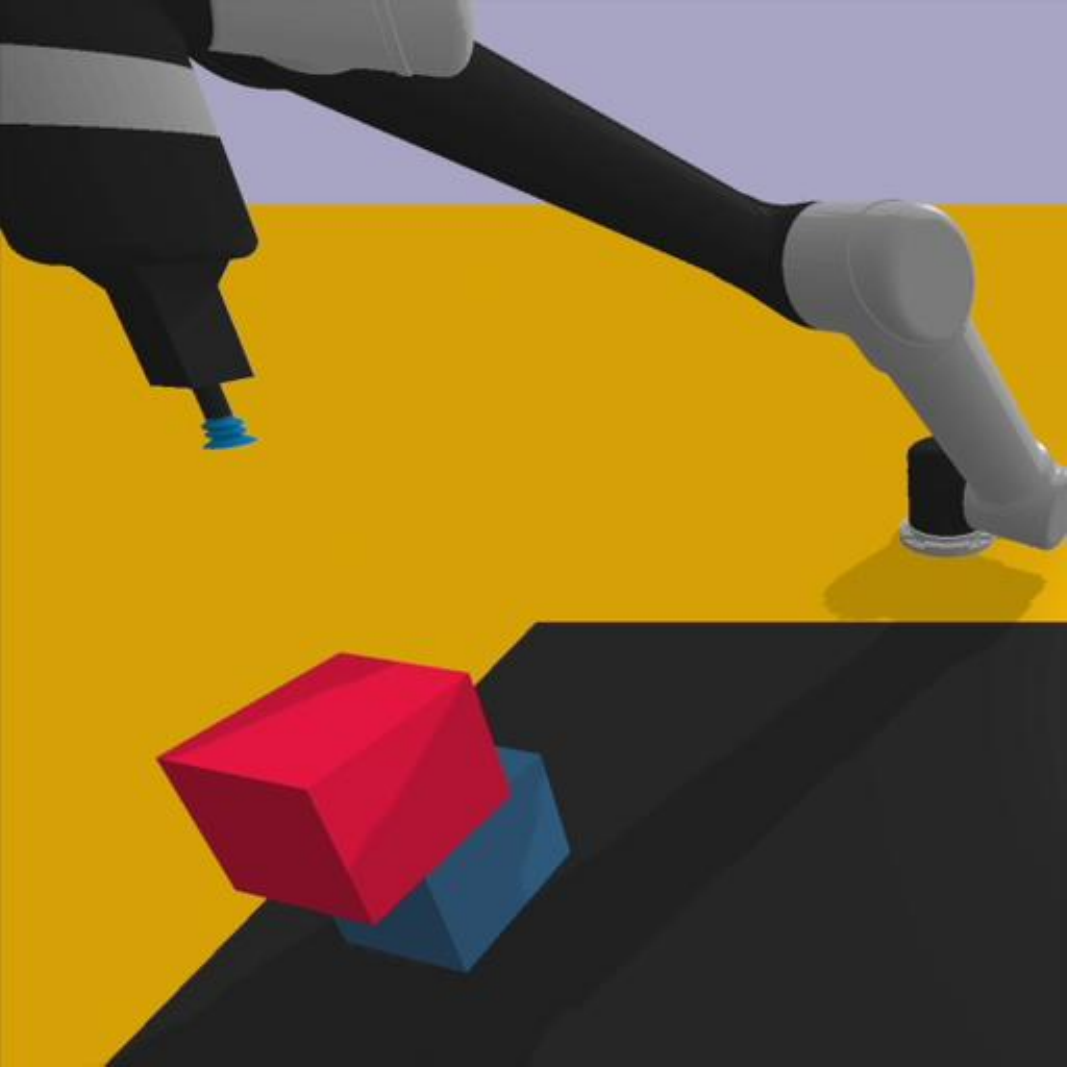}};
    \node[inner sep=0pt, right](img3)at([xshift=1cm]img2.east){\includegraphics[width=3cm]{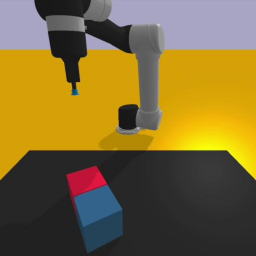}};
    \node[inner sep=0pt, right](img4)at([xshift=.1cm]img3.east){\includegraphics[width=3cm]{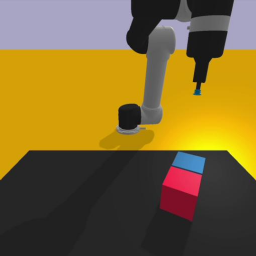}};
    
    \node[inner sep=0pt, below](img5)at([yshift=-.1cm]img1.south){\includegraphics[width=3cm]{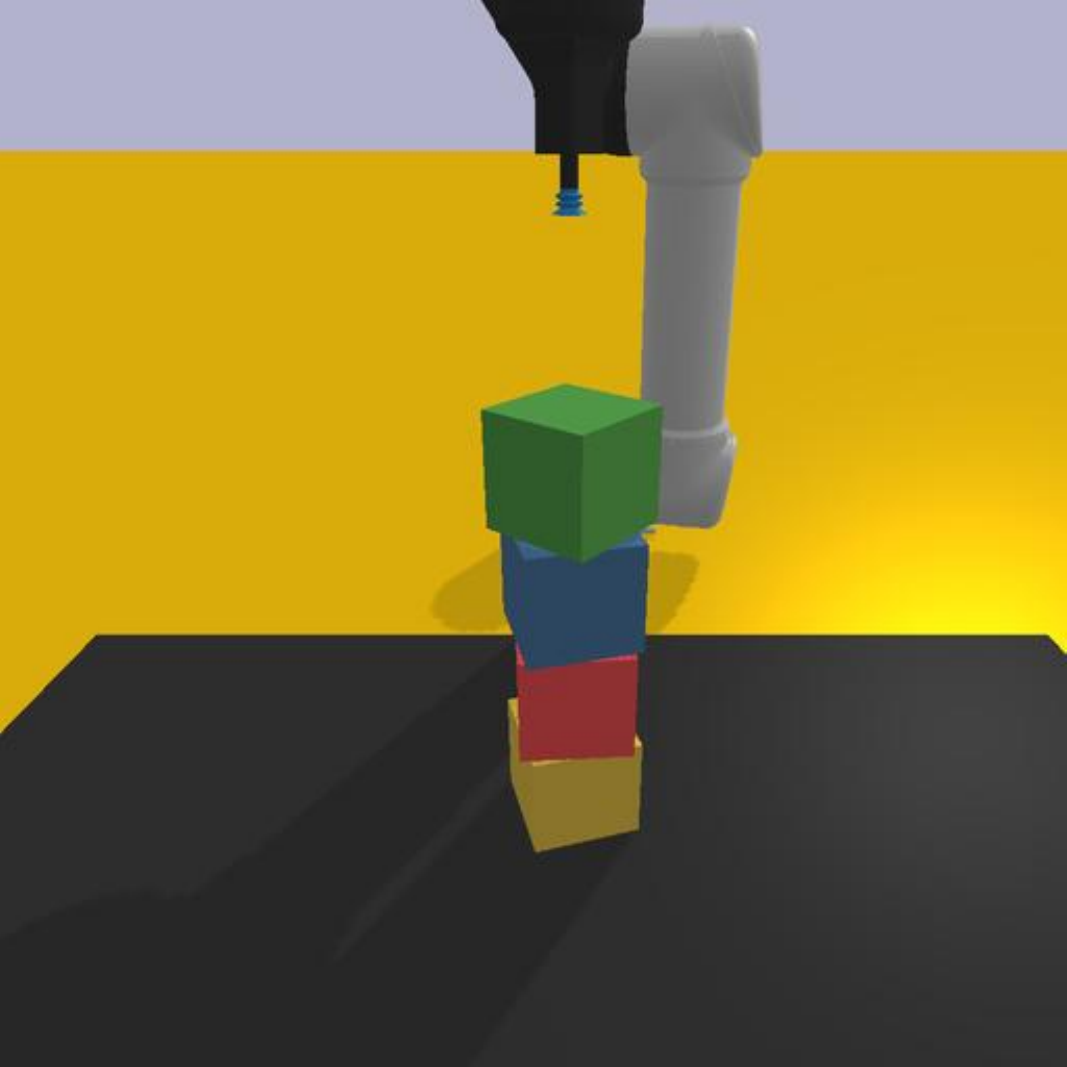}};
    \node[inner sep=0pt, right](img6)at([xshift=.1cm]img5.east){\includegraphics[width=3cm]{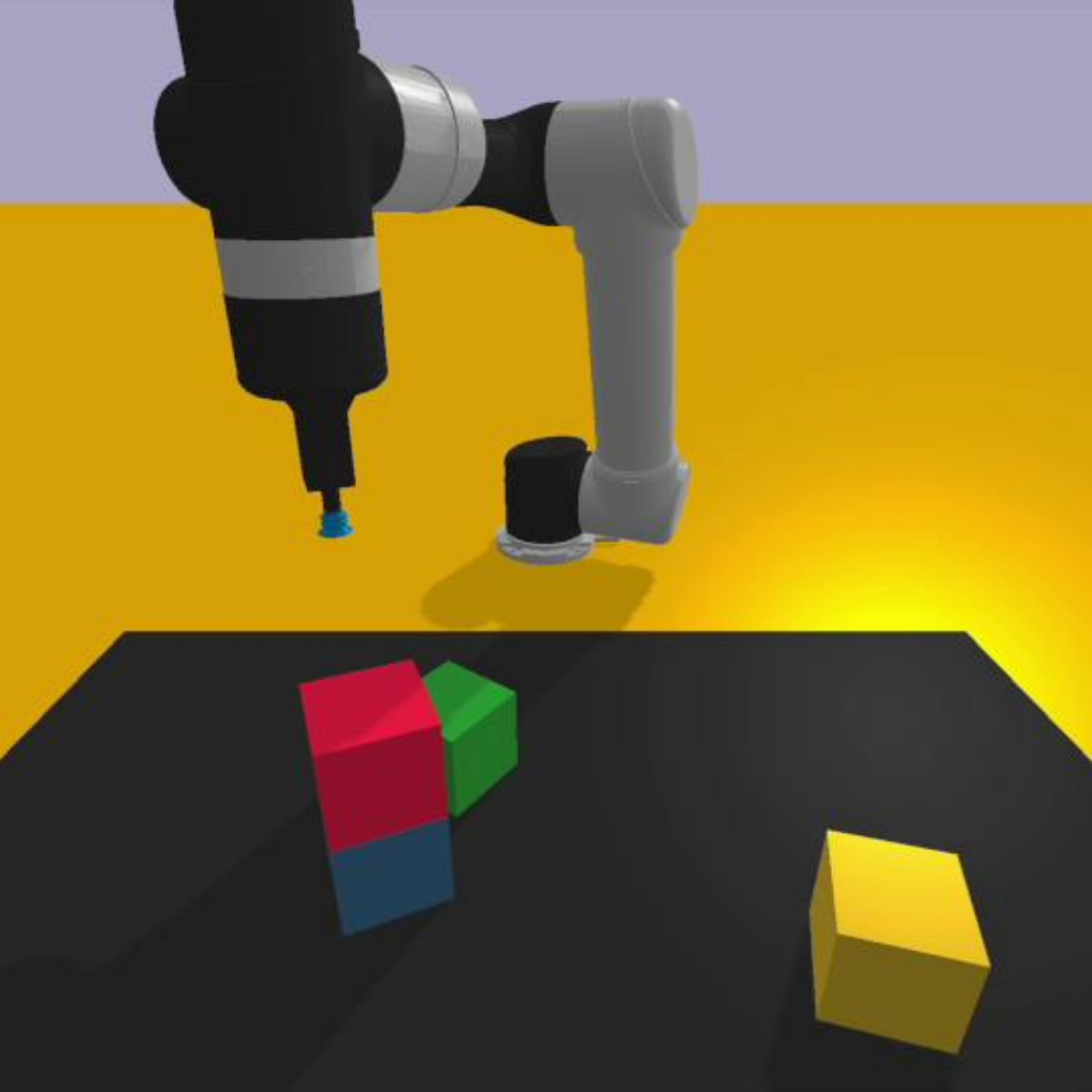}};
    \node[inner sep=0pt, right](img7)at([xshift=1cm]img6.east){\includegraphics[width=3cm]{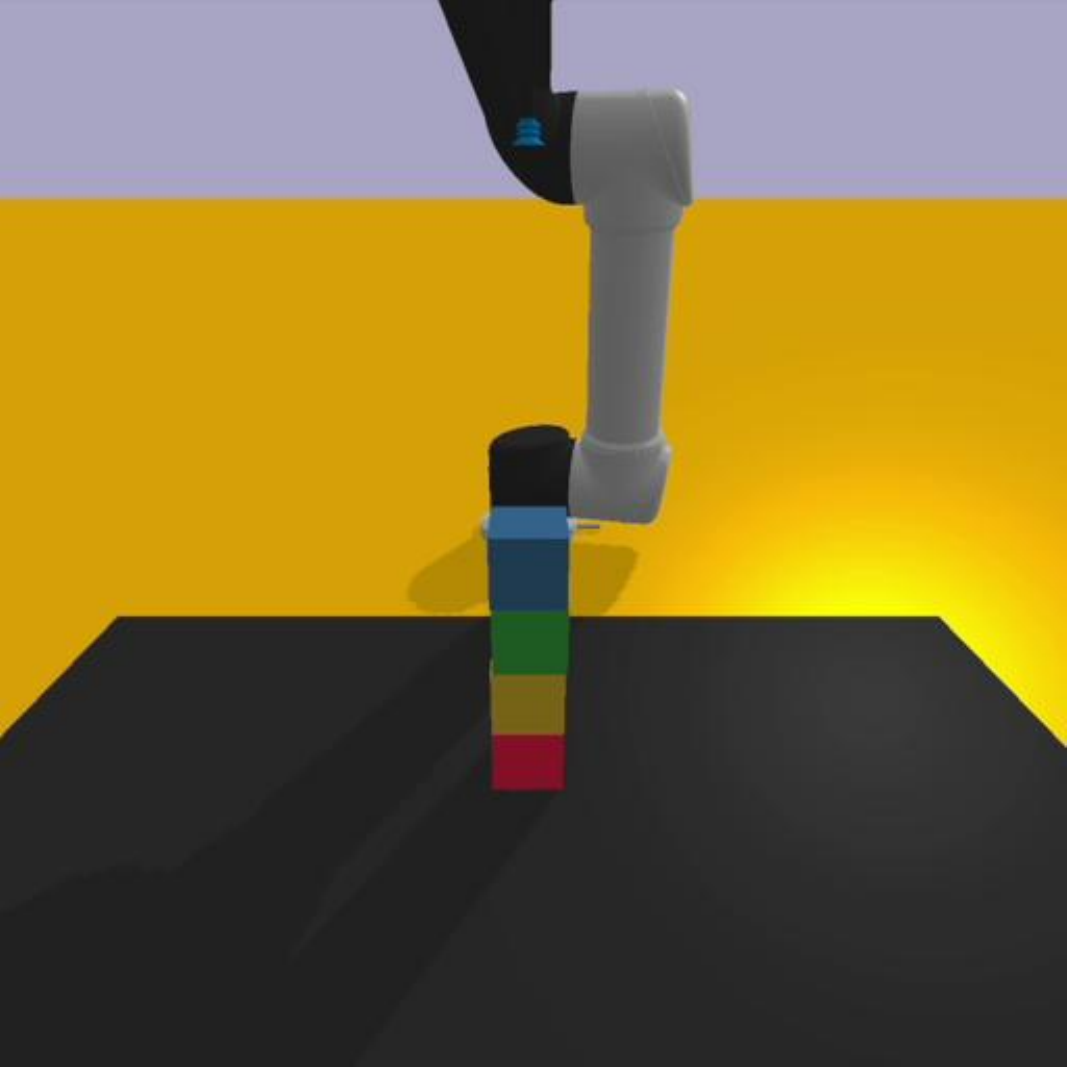}};
    \node[inner sep=0pt, right](img8)at([xshift=.1cm]img7.east){\includegraphics[width=3cm]{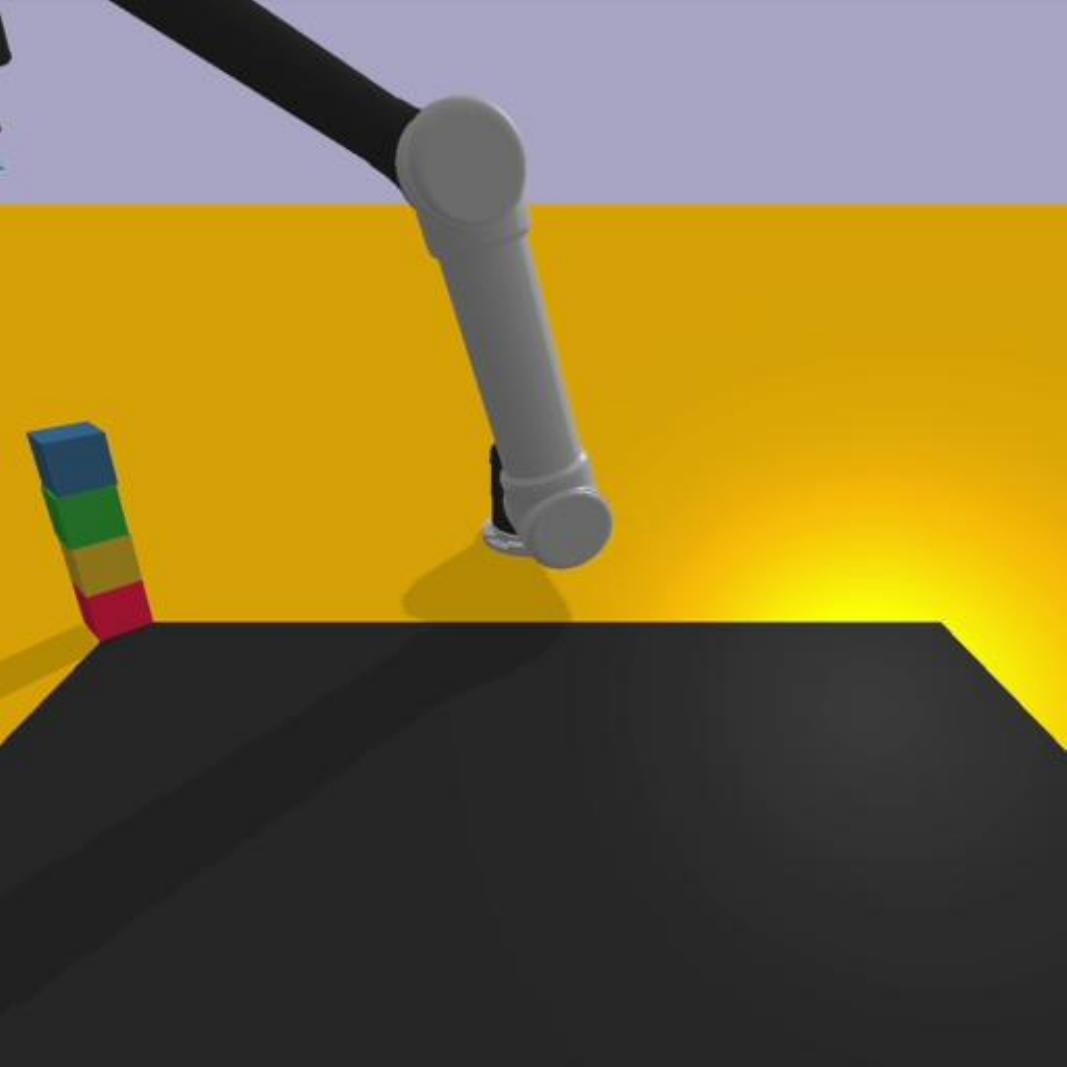}};
    
    \node[inner sep=0pt, below](img9)at([yshift=-.1cm]img5.south){\includegraphics[width=3cm]{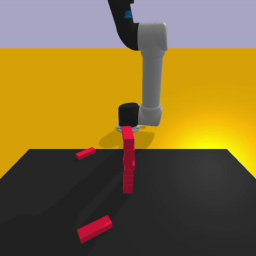}};
    \node[inner sep=0pt, right](img10)at([xshift=.1cm]img9.east){\includegraphics[width=3cm]{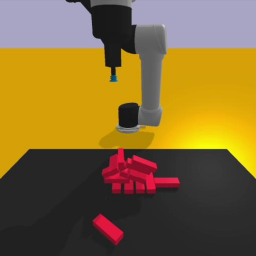}};
    \node[inner sep=0pt, right](img11)at([xshift=1cm]img10.east){\includegraphics[width=3cm]{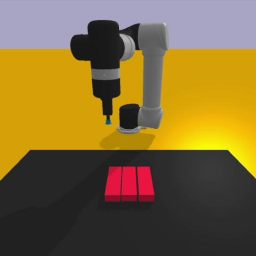}};
    \node[inner sep=0pt, right](img12)at([xshift=.1cm]img11.east){\includegraphics[width=3cm]{imgs/jenga-tower-lyra.pdf}};

    \node[above]at([yshift=.05cm]img3.north east){\textsf{\textbf{LYRA (Ours)}}};
    \node[above]at([yshift=.05cm]img1.north east){\textsf{\textbf{Baselines}}};
    
    \node[]at(-4.5cm, 4.5cm){\textsf{\textbf{a}}};
    \node[below right]at(img1.north west){\textcolor{white}{\textbf{\textsf{b}}}};
    \node[below right]at(img2.north west){\textcolor{white}{\textbf{\textsf{c}}}};
    \node[below right]at(img3.north west){\textcolor{white}{\textbf{\textsf{d}}}};
    \node[below right]at(img4.north west){\textcolor{white}{\textbf{\textsf{e}}}};
    \node[below right]at(img5.north west){\textcolor{white}{\textbf{\textsf{f}}}};
    \node[below right]at(img6.north west){\textcolor{white}{\textbf{\textsf{g}}}};
    \node[below right]at(img7.north west){\textcolor{white}{\textbf{\textsf{h}}}};
    \node[below right]at(img8.north west){\textcolor{white}{\textbf{\textsf{i}}}};
    \node[below right]at(img9.north west){\textcolor{white}{\textbf{\textsf{j}}}};
    \node[below right]at(img10.north west){\textcolor{white}{\textbf{\textsf{k}}}};
    \node[below right]at(img11.north west){\textcolor{white}{\textbf{\textsf{l}}}};
    \node[below right]at(img12.north west){\textcolor{white}{\textbf{\textsf{m}}}};
    \node[above]at(img1.south){\textcolor{white}{\textbf{\textsf{DAHLIA}}}};
    \node[above]at(img2.south){\textcolor{white}{\textbf{\textsf{CaP}}}};
    \node[above]at(img5.south){\textcolor{white}{\textbf{\textsf{DAHLIA}}}};
    \node[above]at(img6.south){\textcolor{white}{\textbf{\textsf{CaP}}}};
    \node[above]at(img9.south){\textcolor{white}{\textbf{\textsf{DAHLIA}}}};
    \node[above]at(img10.south){\textcolor{white}{\textbf{\textsf{CaP}}}};
    \node[below]at(img1.north){\textcolor{white}{\small \textbf{\textsf{next to ref.}}}};
    \node[below]at(img5.north){\textcolor{white}{\small \textbf{\textsf{stack blocks}}}};
    \node[below]at(img9.north){\textcolor{white}{\small \textbf{\textsf{jenga tower}}}};
    \node[right]at([xshift=.2cm]img2.east){\textbf{\textsf{vs}}};
    \node[right]at([xshift=.2cm]img6.east){\textbf{\textsf{vs}}};
    \node[right]at([xshift=.2cm]img10.east){\textbf{\textsf{vs}}};

\end{tikzpicture}
}
\vskip -.1in
\caption{Empirical analysis of skill learning. \textbf{(a)} Case study: Baseline comparison on Ravens long-horizon tasks, reported as the average success rate over 20 attempts. \textbf{(b)}–\textbf{(m)} Snapshots comparison: why HITL is essential for ultra long-horizon manipulation skills.}
\vskip -.1in
\label{fig:empirical-analysis}
\end{figure*}

\subsection{Task-specific-oriented Retrieval and Planning}
In Fig.~\ref{fig:pipeline} Phase III, we enable \textbf{long-horizon planning} through in-context learning, adapting general-purpose LLMs to specific tasks using two main prompt inputs: 
(1) few-shot task-specific \textbf{examples} $\mathcal{E} = \{(l_1, c_1), ..., (l_M, c_M)\}$ that show mappings from instructions to task-specific code plan, and  
(2) a set of task-agnostic \textbf{skills} $\mathcal{Z} = \{z_1, ..., z_N\}$ that the agent can call to compose the plan $c$.  
Due to context window limits, both $|\mathcal{E}| = M$ and $|\mathcal{Z}| = N$ are fixed.
The goal is that the examples and skills are representative enough for the agent to generalize to unseen pairs $(l', c')$, assuming $l'$ reflects the human’s intent and the LLM can produce the correct $c'$ from $\mathcal{E}$ and $\mathcal{Z}$.

Key to our approach is \textbf{dynamically managing the context window}. 
As the number of examples grows, appending all of them to the prompt is infeasible, and many are irrelevant. 
For example, when solving ``press the button'', examples about “open the drawer” add noise and overwhelm the prompt. 
To address this, we use an external memory module with two vector databases: one for few-shot examples $(l, c)$ indexed by their instructions, and one for skills $z$ indexed by their docstrings. 
We implement this using ChromaDB and compute embeddings with OpenAI’s text-embeddings-3 \cite{neelakantan2022text}. 
For a new instruction $l'$, we retrieve the $K=10$ most similar examples based on cosine similarity and append them to the prompt. 
We also retrieve the relevant skill headers from memory and include them in the prompt.
Additionally, our framework introduces a simple yet efficient mechanism for guiding the agent with a shared free-form language format: \textbf{\textit{hints}}. 
Hints allow users to explicitly indicate which previously learned skill may assist with the current task, effectively triggering retrieval from the library of known behaviors. Because our skill library is developed bottom-up, while long-horizon tasks require top-down decomposition, automatic similarity-based retrieval may occasionally overlook a relevant skill, making user-provided hints a simple yet highly effective way to guide the agent toward the correct skills.
As the number of skills $|\mathcal{Z}|$ and behaviours $|\mathcal{E}|$ grows, most are irrelevant, and hints provide a way for the user to direct the agent toward the correct subset. 
If a required sub-behaviour has not yet been learned and cannot be resolved with a hint, the failure signals the user to pause the current skill and teach the missing sub-behaviour first.

With this pipeline, our framework can tackle challenging ultra-long-horizon tasks such as ``build a house'' and ``stack a jenga tower'', which require more than 20 planning steps (Fig.~\ref{fig:pipeline} Phase III left). 
We further validate our method on widely used benchmarks, including Metaworld and Franka Kitchen, covering over 20 task variations where the framework consistently solves all tasks (Fig.~\ref{fig:pipeline} Phase III middle). 
Finally, we deploy the framework in real-world settings on a franka FR3, successfully completing diverse tasks such as building a house, and generalizing to the novel task ``build a temple'' (Fig.~\ref{fig:pipeline} Phase III right). 
It is worth noting that although we present the learning pipeline in three phases, \textbf{the process is not strictly linear}. 
Whenever the agent cannot handle the current task, the user can roll back to the skill learning or extension phase and iterate until the agent meets the requirements.

\section{EXPERIMENTS}
\subsection{Experiment Setup}
Our work focuses on tabletop long-horizon manipulation tasks. 
For fair comparison, we include several code and language generation baselines in our experiments.
Recent work, Voyager \cite{wang2023voyager}, explores skill learning with LLM-based automatic feedback, but it is designed for the Java-based game ``Minecraft'' with simple action primitives. 
Adapting it for Python-based robotic manipulation is costly and impractical. 
Thus, we design a variation of our framework that mirrors their idea, where an LLM provides feedback instead of humans for final task plan generation. 
We also include a w/o memory version of our framework for ablation that simulates the human-guided updates at the prompt level \cite{arenas2024prompt}:

\textbf{Code as Policy (CaP)} \cite{liang2023code}: A representative baseline for LLM-based open-loop code generation without correction or retrieval.

\textbf{LoHoRavens (GPT-4o)} \cite{zhang2023lohoravens}: A language-generation baseline with explicit LLM feedback, using a pretrained RL policy CLIPort as action expert. 

\textbf{DAHLIA} \cite{meng2025data}: A state-of-the-art code generation framework with LLM-based closed-loop control and incremental examples for in-context learning. 

\textbf{LYRA (Ours) w/ LLM feedback}: A variation of our framework that relies on LLM feedback, where RGB-D scenes before and after execution are given to LLM to determine task success and provide feedback. This variation has access to our well-learned memory database.

\textbf{LYRA (Ours) w/o memory}: A variant without the retrieval module, where 25 out of 49 learned skills and 25 out of 86 explored examples in Ravens are randomly sampled and appended until the context window is filled. For Franka Kitchen and MetaWorld, we sample half of the learned skills and examples at random to evaluate how irrelevant data affects feedback efficiency.

\textbf{Vision-Language-Action (VLA) Models}: To enable a comprehensive comparison, we select $\pi_0$ \cite{black2024pi_0} and OpenVLA \cite{kim2024openvla} as VLA baselines. For each task, we collect over 30 trajectories leveraging our LYRA output for finetuning. However, our empirical analysis finds that recent VLA models struggle on the ultra-long-horizon tasks.

For simulation, we build customized tasks upon the \textbf{Ravens} benchmark \cite{zeng2021transporter,zhang2023lohoravens}, where a UE5 robot manipulates multiple tabletop objects (Fig. \ref{fig:empirical-analysis}), including using blocks to build a house, constructing a Jenga tower, etc.
Following prior work \cite{zhang2023lohoravens}, we allow LLMs to set up tasks so scenarios can be modified by user intention. 
To test scalability, we also evaluate on \textbf{Franka Kitchen} \cite{gupta2020relay}, \textbf{MetaWorld} \cite{yu2020meta}, and \textbf{LIBERO-long} \cite{NEURIPS2023_8c3c6668}.
In the real world, we deploy agents on a Franka FR3 with a RealSense camera at the table edge and validate performance on challenging long-horizon tasks.
To bridge the sim-to-real gap, we adopt state-of-the-art vision foundation models \cite{ren2024grounded} to obtain open-world object information.

This work involved human participants in a minimal-risk human--robot interaction evaluation and was exempt from institutional review board approval. 
The study evaluated robot-side performance and interaction efficiency rather than participant characteristics. 
All participants were university volunteers who provided informed consent. 
No sensitive demographic attributes were analyzed, and all interaction records were anonymized before analysis.

\subsection{Human-guided Skill Learning and Reutilization}
\textbf{Ultra-Long-Horizon Performance}: To analyze how human-guided skill learning improves performance, we evaluate all models on the customized Ravens with six long-horizon tasks for a case study 
(Fig. \ref{fig:empirical-analysis}(a)). 
For models with feedback, we report task plan results after up to five iterations. 
For CaP with open-loop control, we report the outcome of each fresh generation. 
All models are tested with 20 attempts, and we report the average success rate (SR) for comparison.
As shown in the figure, CaP performs well only on tasks included in the prompt examples. 
For tasks outside the prompts, such as ``build a 4 x 3 x 2 pyramid'' in 3 of 5 unseen structure building tasks, it fails to generalize, achieving only a 0.45 SR. 
Compared with open-loop CaP, LoHoRavens and DAHLIA improve performance by using LLMs for closed-loop feedback.  
However, relying solely on LLM feedback can cause the agent to get stuck in long-horizon planning.
Our framework variation, LYRA w/ LLM feedback, further underscores this issue.  
Although the agent has access to all learned skills and examples in the database, the LLM’s limited reasoning makes it uncertain about which skills to apply or what the agent can actually perform.  
As a result, this causes a drop in performance, and the agent may get stuck on ultra long-horizon tasks, with an average SR of only 0.77.
Moreover, LYRA w/o memory simulates methods that extend data directly in the prompt.  
With well-learned skills and examples, this variant represents the upper bound of such methods.
As shown in the figure, its performance drops substantially due to the limited prompt length, and unrelated examples in the prompt further interfere with reasoning, resulting in an average SR of 0.66.  
By comparison, our framework achieves a high SR of 0.93, with up to \textbf{27\% SR improvement} over the variant models. 
For a comprehensive evaluation, we also compare against recent pure learning-based end-to-end VLA models.
Our framework achieves an average SR of 0.95 on LIBERO-Long, outperforming VLA models such as $\pi_0$ and OpenVLA (average 0.65).
Notably, current VLAs fail on ultra-long-horizon tasks such as 
\begin{table}[t!]
\centering
\caption{Comparison between OpenVLA, $\pi_0$, and LYRA on LIBERO-long, and ultra-long-horizon Ravens tasks.}
\vskip -.1in
\resizebox{\linewidth}{!}{
\begin{tabular}{l | cccc}
\toprule
\rowcolor{nature_tab_gray1}
 & \shortstack{\textbf{LIBERO}\\\textbf{-- long}}
 & \shortstack{\textbf{Build a}\\\textbf{Jenga tower}}
 & \shortstack{\textbf{Construct a}\\\textbf{4$\times$4$\times$3 pyramid}}
 & \shortstack{\textbf{Build a}\\\textbf{House}} \\
\midrule
\textbf{OpenVLA} 
& 0.65 & 0.00 & 0.00 & 0.00 \\
\midrule
\rowcolor{nature_tab_gray2}
$\bm{\pi_{0}}$ 
& 0.60 & 0.00 & 0.00 & 0.00 \\
\midrule
\textbf{LYRA (Ours)} 
& 0.95 & 0.95 & 0.85 & 0.80 \\
\bottomrule
\end{tabular}}
\label{tab:vla_comparison}
\vskip -.2in
\end{table}
constructing a Jenga tower and building a house (with SR 0.0), underscoring our approach as an effective alternative for ultra-long-horizon manipulation.

\textbf{Preference Alignment and Robustness}: Relying only on success rate is limited, as it cannot capture alignment with user preferences.  
A key contribution of our approach is representing human preferences from language corrections in a form that can be preserved and reused through skill inheritance.  
Snapshots in Fig.~\ref{fig:empirical-analysis}(b)–(m) show how user feedback improves task performance beyond what SR reflects.  
For example, in (b) vs. (d), (e), and (f) vs. (h), (i), the baseline places blocks at the target positions, but their poses and distances do not match user expectations.  
Unlike LLM-only feedback, which often labels sub-optimal results as successful, user interventions guide the agent to acquire precise skills (e.g., stacking corner-to-corner or aligning blocks with fixed spacing) that lead to more reliable and preference-aligned outcomes.  
Fig.~\ref{fig:empirical-analysis}(j), (k) vs. (l), (m) further highlights the reliability of human-in-the-loop skill learning. 
In this example case, we observe \textbf{false positives} from LLM evaluation, where incomplete tasks are incorrectly marked as successful.  
For instance, in Fig.~\ref{fig:empirical-analysis}(j), DAHLIA stacks long blocks in the same direction for a Jenga tower.  
Although clearly incorrect to a human, the LLM still labels it as correct, raising doubts about evaluation reliability.  
In contrast, our framework reuses the ``build jenga layer'' skill, which was learned earlier with human guidance in a simpler scene, to stack a stable Jenga tower.
Full results see supplementary video.  

\begin{figure}[h]
    \centering
    \resizebox{\linewidth}{!}{
    \begin{tikzpicture}
        \node[inner sep=0pt, 
        ](img1)at(0,0){\includegraphics[width=2cm]{imgs/kitchen-1.pdf}};
        \node[fill=black, opacity=.6, text opacity=1, above]at(img1.south){\scriptsize \textcolor{white}{kettle}};
        \node[inner sep=0pt, right, 
        ](img2)at([xshift=.1cm]img1.east){\includegraphics[width=2cm]{imgs/kitchen-2.pdf}};
        \node[fill=black, opacity=.6, text opacity=1, above]at(img2.south){\scriptsize \textcolor{white}{slide cabinet}};
        \node[inner sep=0pt, right, 
        ](img3)at([xshift=.1cm]img2.east){\includegraphics[width=2cm]{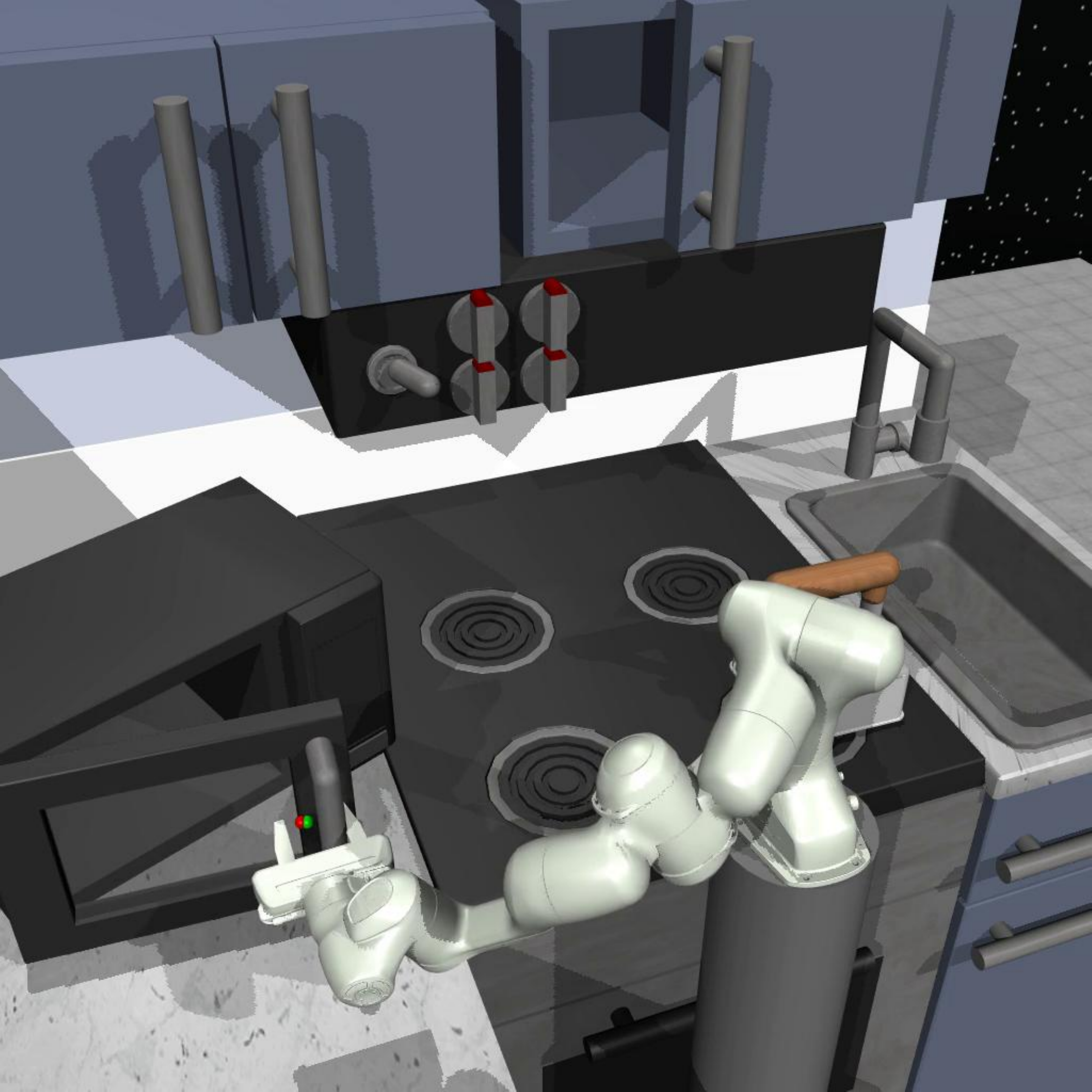}};
        \node[fill=black, opacity=.6, text opacity=1, above]at(img3.south){\scriptsize \textcolor{white}{microwave}};
        \node[inner sep=0pt, right, 
        ](img4)at([xshift=.1cm]img3.east){\includegraphics[width=2cm]{imgs/kitchen-4.pdf}};
        \node[fill=black, opacity=.6, text opacity=1, above]at(img4.south){\scriptsize \textcolor{white}{hinge cabinet}};
        \node[inner sep=0pt, below, 
        ](img5)at([yshift=-.1cm]img1.south){\includegraphics[width=2cm]{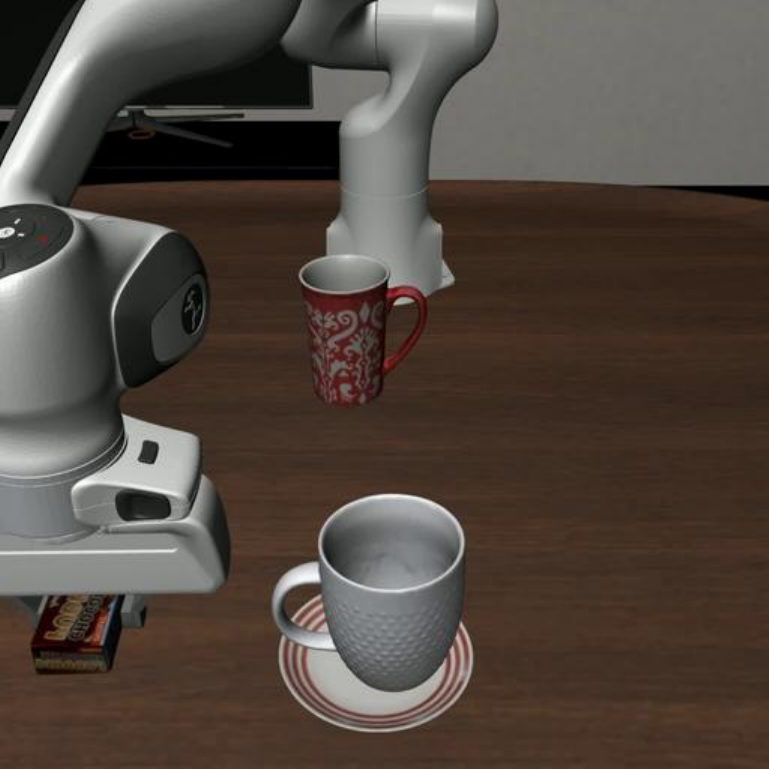}};
        \node[fill=black, opacity=.6, text opacity=1, above]at(img5.south){\scriptsize \textcolor{white}{tea set}};
        \node[inner sep=0pt, right, 
        ](img6)at([xshift=.1cm]img5.east){\includegraphics[width=2cm]{imgs/libero-task9_frame_000099.pdf}};
        \node[fill=black, opacity=.6, text opacity=1, above]at(img6.south){\scriptsize \textcolor{white}{coffee heat}};
        \node[inner sep=0pt, right, 
        ](img7)at([xshift=.1cm]img6.east){\includegraphics[width=2cm]{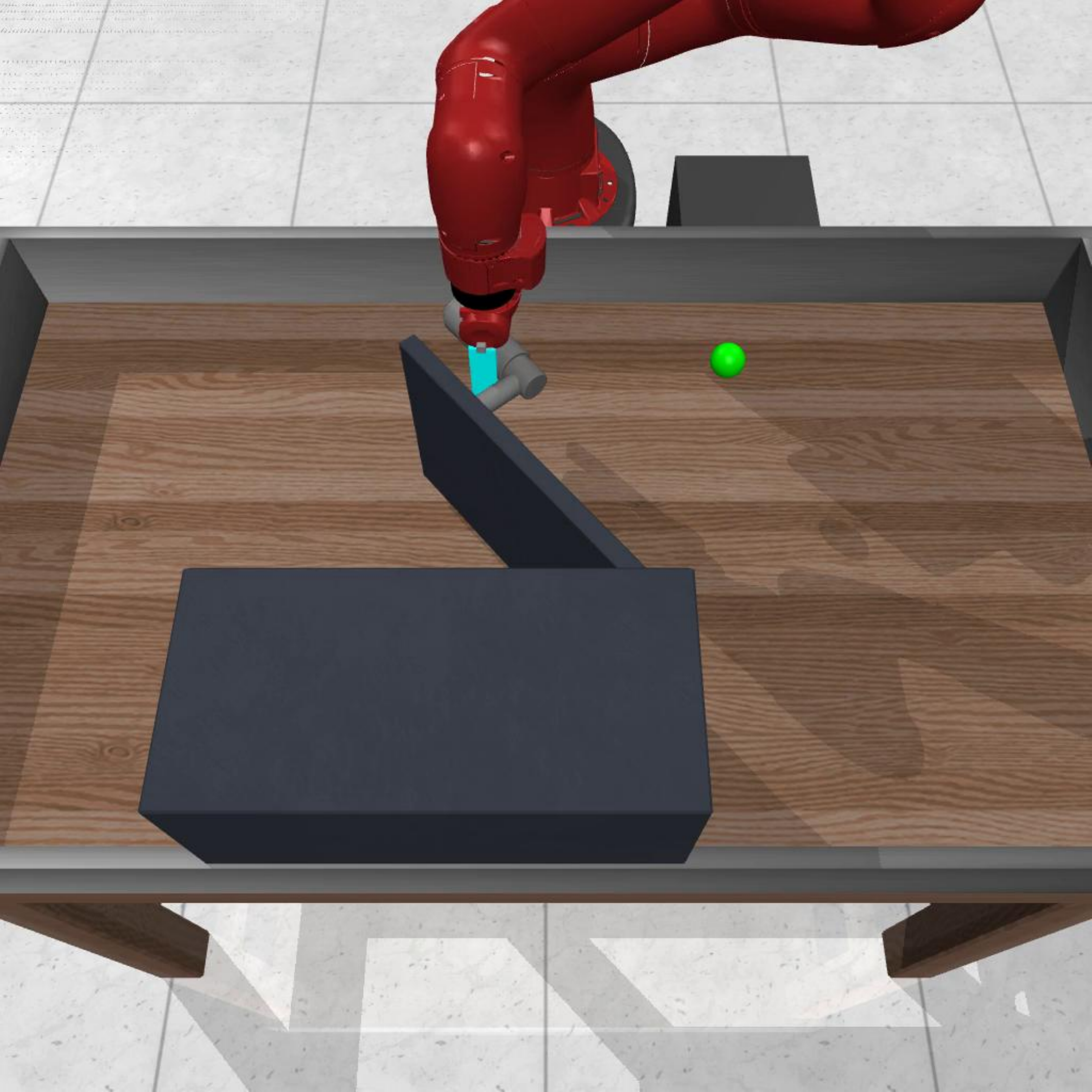}};
        \node[fill=black, opacity=.6, text opacity=1, above]at(img7.south){\scriptsize \textcolor{white}{door open}};
        \node[inner sep=0pt, right, 
        ](img8)at([xshift=.1cm]img7.east){\includegraphics[width=2cm]{imgs/mw-hammer.pdf}};
        \node[fill=black, opacity=.6, text opacity=1, above]at(img8.south){\scriptsize \textcolor{white}{hammer}};
        \node[below right, fill=black]at(img1.north west){\scriptsize\textcolor{white}{\textbf{\textsf{a}}}};
        \node[below right, fill=black]at(img2.north west){\scriptsize\textcolor{white}{\textbf{\textsf{b}}}};
        \node[below right, fill=black]at(img3.north west){\scriptsize\textcolor{white}{\textbf{\textsf{c}}}};
        \node[below right, fill=black]at(img4.north west){\scriptsize\textcolor{white}{\textbf{\textsf{d}}}};
        \node[below right, fill=black]at(img5.north west){\scriptsize\textcolor{white}{\textbf{\textsf{e}}}};
        \node[below right, fill=black]at(img6.north west){\scriptsize\textcolor{white}{\textbf{\textsf{f}}}};
        \node[below right, fill=black]at(img7.north west){\scriptsize\textcolor{white}{\textbf{\textsf{g}}}};
        \node[below right, fill=black]at(img8.north west){\scriptsize\textcolor{white}{\textbf{\textsf{h}}}};
    \end{tikzpicture}
    }
    \vskip -.05in
    \caption{
    Scalability validation across three benchmarks.
    (a)--(d) Franka Kitchen tasks: kettle, slide cabinet, microwave, and hinge cabinet. 
    (e)--(f) LIBERO-long tea set, coffee heat task. 
    (g)--(h) MetaWorld tasks: door open, and hammer. 
    }
    \vskip -.2in
    \label{fig:efficiency-validation}
\end{figure}

%
\textbf{Efficiency and Scalability}: Skill-oriented human-guided feedback is more efficient than task-specific flat code feedback from LLMs.  
Table~\ref{tab:NoC} shows the average number of corrections (NoC) across Franka Kitchen (4 tasks) and MetaWorld (20 tasks).  
A lower NoC means the framework achieves success and aligns with user expectations more efficiently.  
Both DAHLIA and our LYRA variant w/o memory require more feedback iterations (4.75 and 5.00 in Franka Kitchen) compared to our LYRA framework (2.75).  
Although LYRA w/o memory has access to learned skills, the presence of irrelevant information in the prompt hinders reasoning and leads to more feedback iterations.
In MetaWorld, DAHLIA struggles to complete several tasks within 10 feedback rounds (e.g., hammer), showing the difficulty of fixing low-level execution errors through LLM feedback alone.
By comparison, our LYRA framework achieves the fewest corrections in both benchmarks (2.75 in Franka Kitchen and 2.55 in MetaWorld), yielding an average \textbf{42\% NoC efficiency improvement} over baselines.  
This gain comes not only from learning skills with human guidance, but also from our \textbf{hint} mechanism.  
Hints let users guide the agent using free-form text when retrieval fails or when irrelevant data is retrieved.  
By clarifying what the agent can do and which skills to apply, hints reduce unnecessary corrections compared to LLM feedback.
We further measured the average feedback time of three human users and LLMs over 90 Ravens trials. 
Human users and LLMs required 27.6 s and 17.6 s on average to complete one feedback loop, respectively. 
Humans were faster in reasoning and decision-making (7.5 s vs. 14.2 s), while the additional human time mainly came from manual text entry; in contrast, LLMs are faster in text generation or response output (3.4 s vs. 20.1 s).
Fig.~\ref{fig:efficiency-validation}(a)–(h) shows example snapshots of our framework’s performance on three benchmarks.
\begin{table}[t!]
    \centering
    \caption{Average number of corrections (NoC) for LLM-based flat code generation vs our human-guided skill code generation.}
    \vskip -.1in
    \resizebox{\linewidth}{!}{
    \begin{tikzpicture}
        \node[](table)at(0,0){
        \begin{tabular}{c | c c }
            \toprule
            \rowcolor{nature_tab_gray1}
            \textbf{Framework} & \textbf{Franka Kitchen (4)} & \textbf{MetaWorld (MT20)} \\
            \midrule
            \textbf{DAHLIA} & $4.75$  & $>10^\dagger$  \\
            \rowcolor{nature_tab_gray2}
            \midrule
            \textbf{LYRA w/o memory} & $5.00$ & $4.15$  \\
            \midrule
            \textbf{LYRA (Ours)} & $\bm{2.75}$ & $\bm{2.55}$  \\
            \bottomrule
        \end{tabular}
        };
        \node[below right, text width=8.5cm, align=left](footnote)at([yshift=.1cm]table.south west){\footnotesize $^\dagger$Some tasks may not be completed within 10 feedback rounds. E.g., in the hammer task, an incorrect grasp may cause the hammer to drop, which LLM-based closed-loop feedback systems may struggle to correct, often leading to task failure.};
    \end{tikzpicture}
     }       
    \label{tab:NoC}
    \vskip -.3in
\end{table}

\subsection{Can You Build a House?}
\begin{figure*}[ht!]
\vskip 0.05in
    \centering
    \resizebox{.95\textwidth}{!}{
    \begin{tikzpicture}
        \node[rectangle, fill=ffgreen_pv!60, rounded corners=2pt, minimum width=4cm, minimum height=1cm](0a)at(0,0){\textsf{build\_house}};

        \node[rectangle, fill=fflightgreen!60, rounded corners=2pt, minimum width=4cm, minimum height=1cm](1a)at([xshift=-3cm, yshift=-2cm]0a.south){\textsf{build\_house\_base}};
        \node[rectangle, fill=fflightgreen!60, rounded corners=2pt, minimum width=4cm, minimum height=1cm](1b)at([xshift=4cm, yshift=-2cm]0a.south){\textsf{identify\_roof\_tiles}};
        \node[rectangle, fill=fflightgreen!60, rounded corners=2pt, minimum width=4cm, minimum height=1cm, right](1c)at([xshift=.5cm]1b.east){\textsf{identify\_beam\_block}};
        \node[rectangle, fill=fflightgreen!60, rounded corners=2pt, minimum width=4cm, minimum height=1cm, right](1d)at([xshift=.5cm]1c.east){\textsf{identify\_roof\_base}};
        \node[rectangle, fill=ffyellow!60, rounded corners=2pt, minimum width=4cm, minimum height=1cm, left](1e)at([xshift=-6cm]1a.west){\textsf{get\_blocks\_by\_color}};

        \node[rectangle, fill=fflightgreen!60, rounded corners=2pt, minimum width=3cm, minimum height=1cm](2a)at([xshift=-5cm, yshift=-2cm]1a.south){\textsf{assemble\_roof}};
        \node[rectangle, fill=ffyellow!60, rounded corners=2pt, minimum width=4cm, minimum height=1cm](2b)at([xshift=2cm, yshift=-2cm]1a.south){\textsf{build\_structure\_from\_blocks}};
        \node[rectangle, fill=fforange_pv!60, rounded corners=2pt, minimum width=4cm, minimum height=1cm](2c)at([xshift=1cm, yshift=-2cm]1c.south){\textsf{get\_object\_color}};
        \node[rectangle, fill=fforange_pv!60, rounded corners=2pt, minimum width=4cm, minimum height=1cm](2d)at([yshift=-2cm]1e.south){\textsf{get\_objects}};

        \node[rectangle, fill=fflightgreen!60, rounded corners=2pt, minimum width=3cm, minimum height=1cm](3a)at([xshift=1cm, yshift=-2cm]2a.south){\textsf{place\_roof\_tiles}};
        \node[rectangle, fill=fflightgreen!60, rounded corners=2pt, minimum width=4cm, minimum height=1cm](3b)at([xshift=-1cm, yshift=-2cm]2b.south){\textsf{make\_line\_of\_blocks\_next\_to}};
        \node[rectangle, fill=ffyellow!60, rounded corners=2pt, minimum width=3cm, minimum height=1cm, right](3c)at([xshift=.5cm]3b.east){\textsf{make\_line\_with\_blocks}};

        \node[rectangle, fill=ffyellow!60, rounded corners=2pt, minimum width=4cm, minimum height=1cm](4a)at([xshift=-1cm, yshift=-2cm]3b.south){\textsf{stack\_blocks}};
        \node[rectangle, fill=ffyellow!60, rounded corners=2pt, minimum width=4cm, minimum height=1cm, right](4b)at([xshift=1cm]4a.east){\textsf{move\_block\_next\_to\_reference}};

        \node[rectangle, fill=fforange_pv!60, rounded corners=2pt, minimum width=4cm, minimum height=1cm](5a)at([xshift=-6cm, yshift=-2cm]4a.south){\textsf{get\_object\_pose}};
        \node[rectangle, fill=fforange_pv!60, rounded corners=2pt, minimum width=4cm, minimum height=1cm, right](5b)at([xshift=2cm]5a.east){\textsf{put\_first\_on\_second}};
        \node[rectangle, fill=fforange_pv!60, rounded corners=2pt, minimum width=4cm, minimum height=1cm, right](5c)at([xshift=7.5cm]5b.east){\textsf{get\_object\_size}};

        \draw[->, line width=1pt](0a) -- (1a);
        \draw[->, line width=1pt](0a) -- (1b);
        \draw[->, line width=1pt](0a) -- (1c);
        \draw[->, line width=1pt](0a) to[out=-10, in=150] (1d);
        \draw[->, line width=1pt](0a.west) to[out=180, in=30] (1e);
        \draw[->, line width=1pt](0a.west) to[out=180, in=90] (2a);
        \draw[->, line width=1pt](0a.west) to[out=180, in=120] ([xshift=.5cm]5a.north west);
        \draw[->, line width=1pt](0a) to[out=-90, in=100] (3c);

        \draw[->, line width=1pt](1a) to[out=180, in=140] (5a.north west);
        \draw[->, line width=1pt](1a) to[out=-135, in=135] (4a);
        \draw[->, line width=1pt](1a) to[out=-135, in=135] (3b);
        \draw[->, line width=1pt](1a) to[out=0, in=110] (5c);
        \draw[->, line width=1pt](1a) -- (2b);
        \draw[->, line width=1pt](1b) -- (2c);
        \draw[->, line width=1pt](1c) -- (2c);
        \draw[->, line width=1pt](1d) -- (2c);
        \draw[->, line width=1pt](1e) -- (2d);
        \draw[->, line width=1pt](1b) to[out=-70, in=95] (5c);
        \draw[->, line width=1pt](1c) to[out=-130, in=70] (5c);
        \draw[->, line width=1pt](1d) to[out=-100, in=60] (5c);

        \draw[->, line width=1pt](2a) to[out=-150, in=120] (5a);
        \draw[->, line width=1pt](2a) to[out=-130, in=170] (5b);
        \draw[->, line width=1pt](2a) -- (3a);
        \draw[->, line width=1pt](2b) -- (3c);
        \draw[->, line width=1pt](2b) to[out=0, in=120] (5c);

        \draw[->, line width=1pt](3a) -- (5a);
        \draw[->, line width=1pt](3a) to[out=-90] (5b);
        \draw[->, line width=1pt](3a) -- (4b);
        \draw[->, line width=1pt](3b) -- (4b);
        \draw[->, line width=1pt](3c) -- (4b);

        \draw[->, line width=1pt](4a) -- (5a);
        \draw[->, line width=1pt](4a) -- (5b);
        \draw[->, line width=1pt](4a) -- (5c);
        \draw[->, line width=1pt](4b) -- (5a);
        \draw[->, line width=1pt](4b) -- (5b);
        \draw[->, line width=1pt](4b) -- (5c);

        \node[fill=fflightgreen!60, minimum width=.5cm, minimum height=.5cm, label={[label distance=0]right:\Large \textbf{\textsf{skills learned specifically for building a house}}}](l1)at(-14.5cm, -14cm){};
        \node[fill=ffyellow!60, minimum width=.5cm, minimum height=.5cm, label={[label distance=0]right:\Large \textbf{\textsf{skills learned independently of building a house}}}, right](l2)at([xshift=12cm]l1.east){};
        \node[fill=fforange_pv!60, minimum width=.5cm, minimum height=.5cm, label={[label distance=0]right:\Large \textbf{\textsf{Core primitives}}}, right](l3)at([xshift=12cm]l2.east){};

        \node[inner sep=0, anchor=north west](img1) at(19cm, 0cm){\includegraphics[width=6cm]{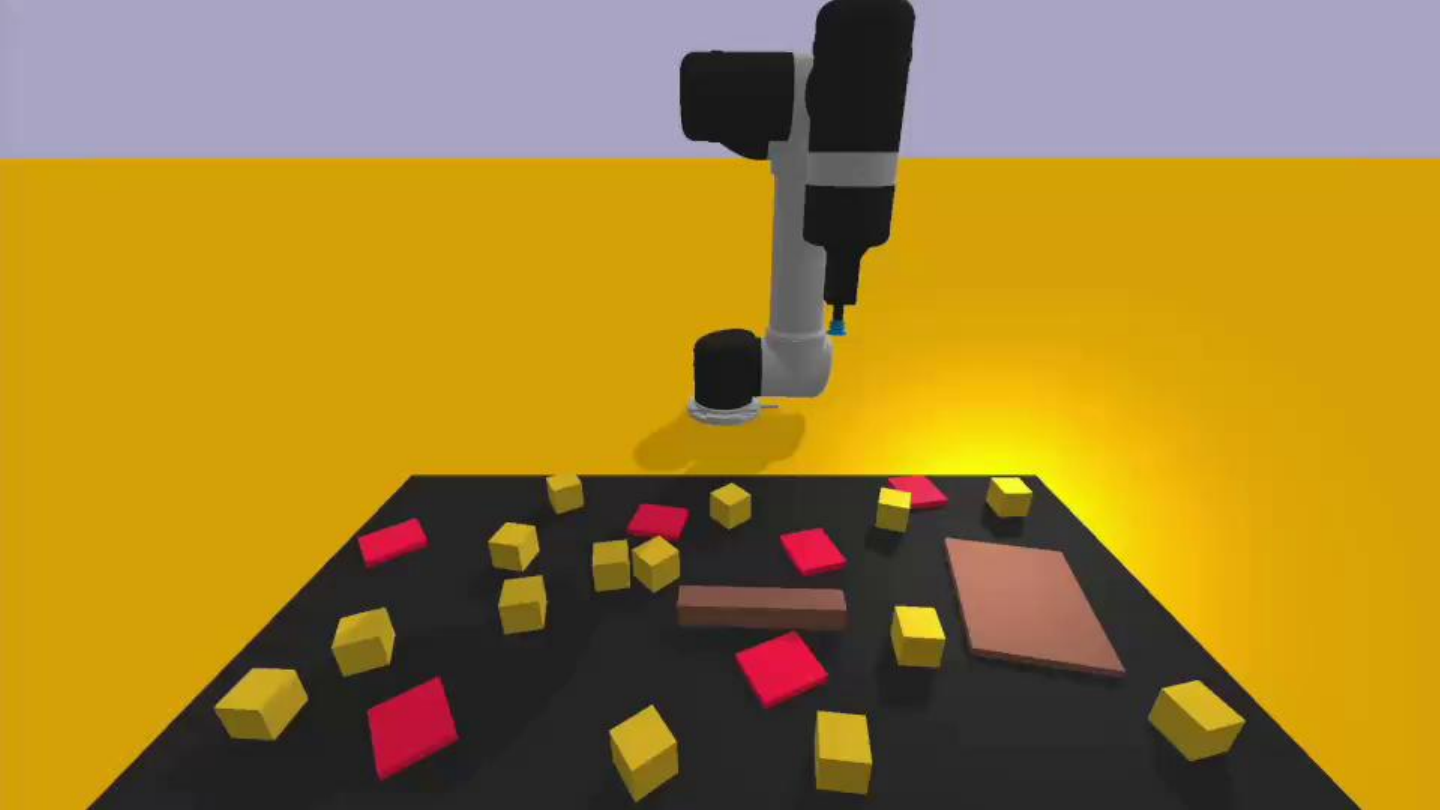}};
        \node[inner sep=0, below left, label={[label distance=-.5cm]-90:\large \textcolor{white}{\textsf{clear table}}}](img2) at([xshift=-.2cm, yshift=-.5cm]img1.south west){\includegraphics[width=5cm]{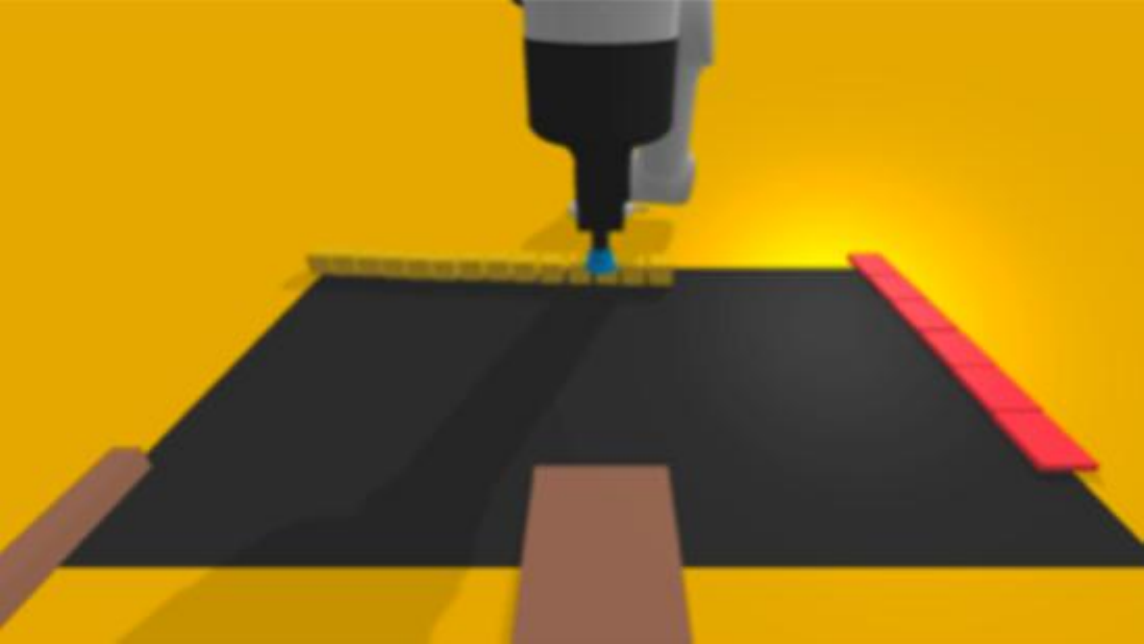}};
        \node[inner sep=0, below, label={[label distance=-.5cm]-90:\large \textcolor{white}{\textsf{build house base}}}](img3) at([yshift=-.5cm]img1.south){\includegraphics[width=5cm]{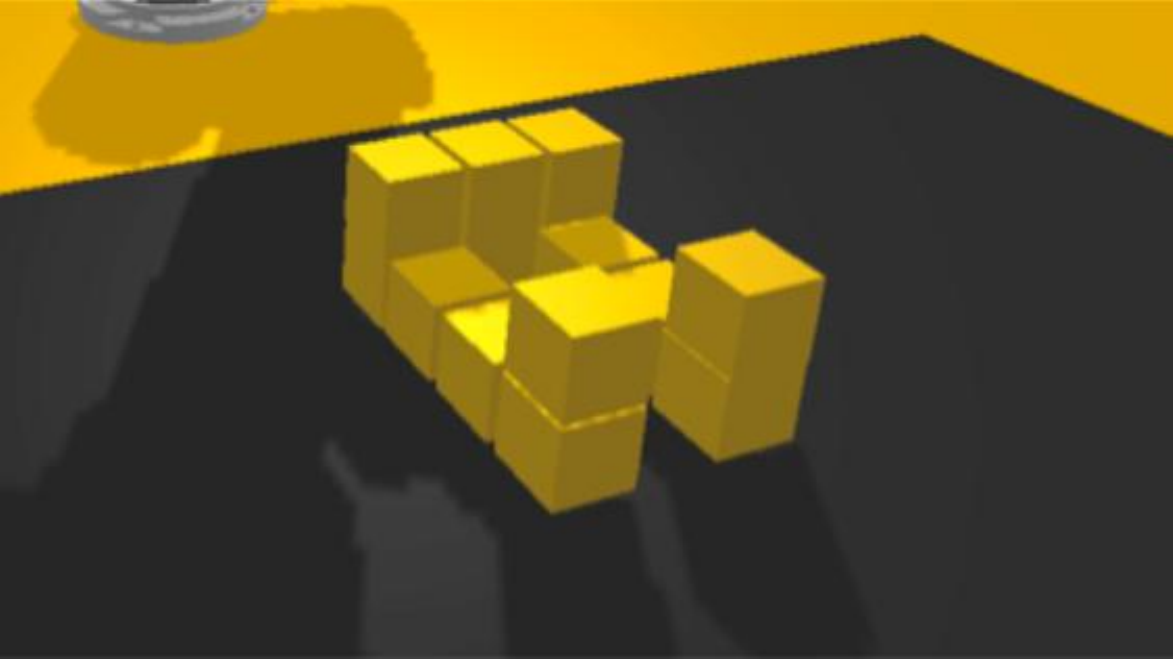}};
        \node[inner sep=0, below right, label={[label distance=-.5cm]-90:\large \textcolor{white}{\textsf{build house roof}}}](img4) at([xshift=.2cm, yshift=-.5cm]img1.south east){\includegraphics[width=5cm]{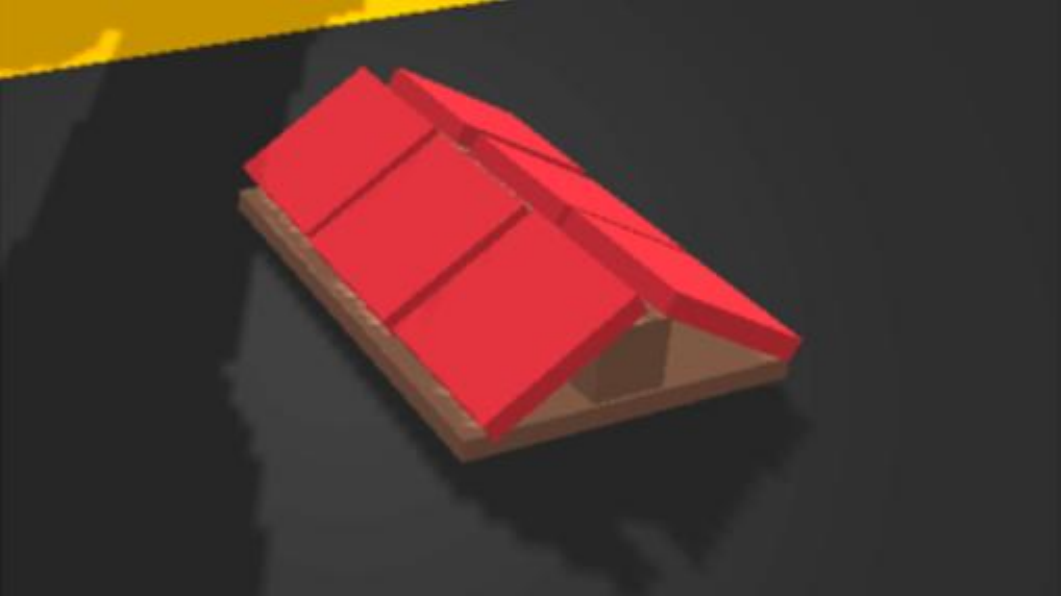}};
        \node[inner sep=0, below](img5) at([yshift=-.5cm]img2.south){\includegraphics[width=5cm]{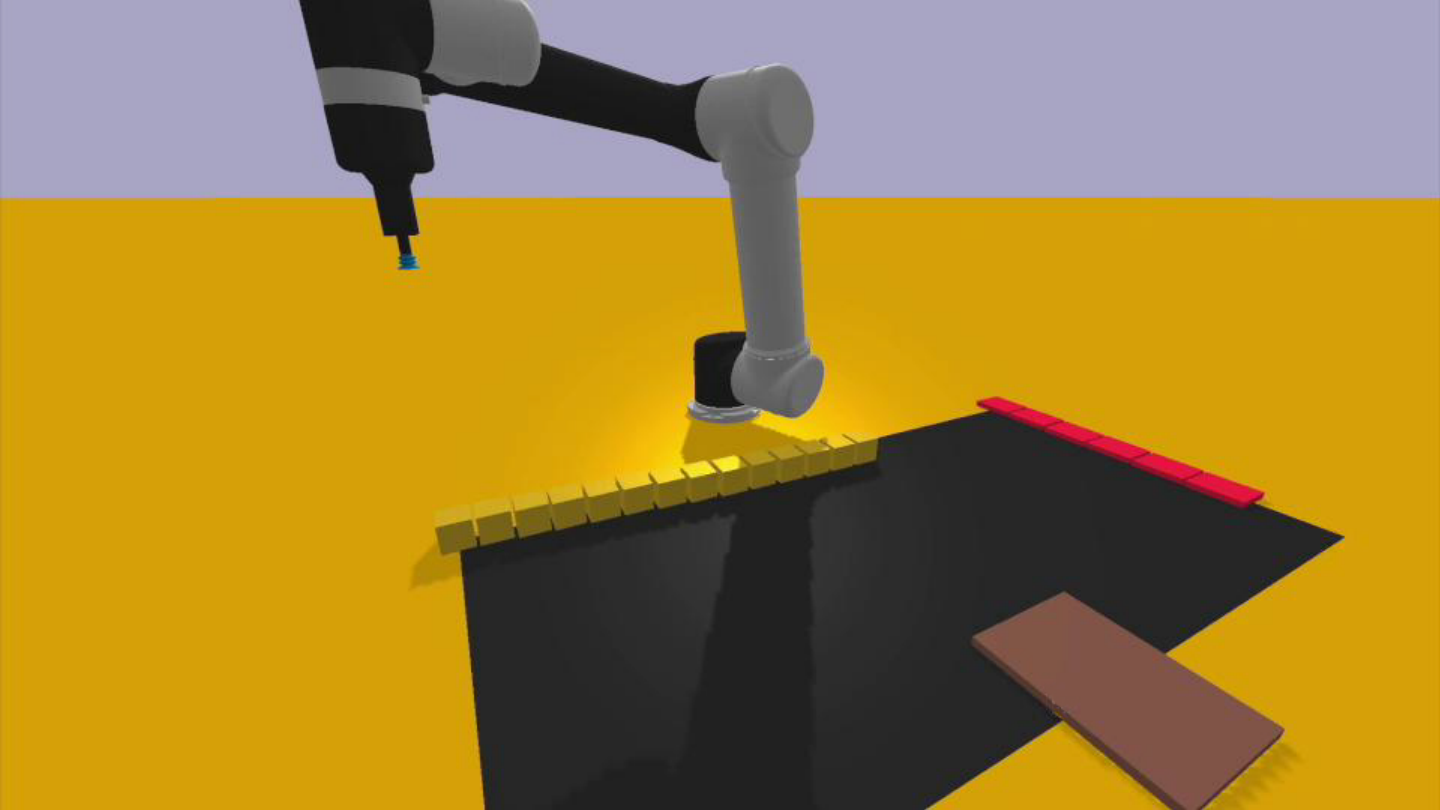}};
        \node[inner sep=0, below](img6) at([yshift=-.5cm]img3.south){\includegraphics[width=5cm]{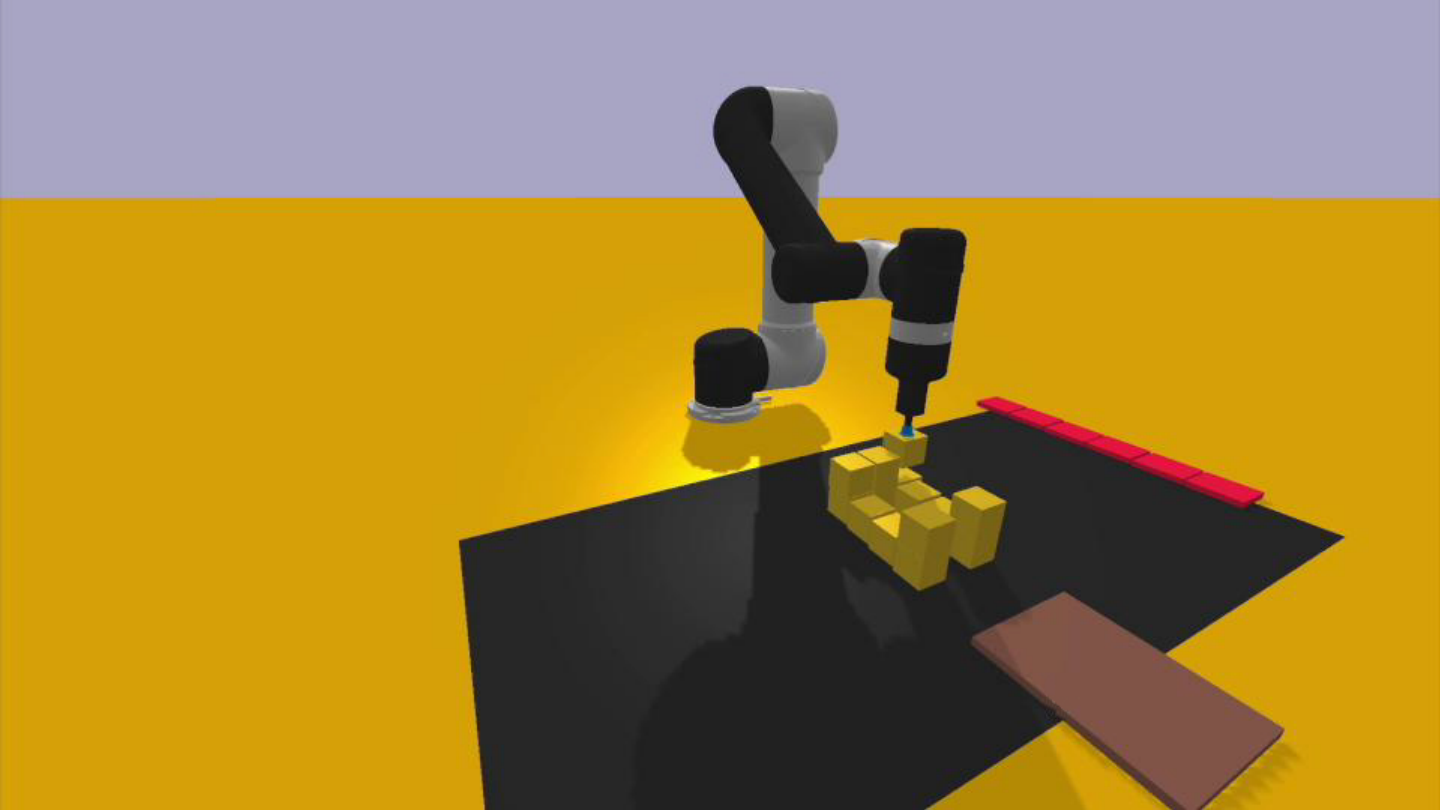}};
        \node[inner sep=0, below](img7) at([yshift=-.5cm]img4.south){\includegraphics[width=5cm]{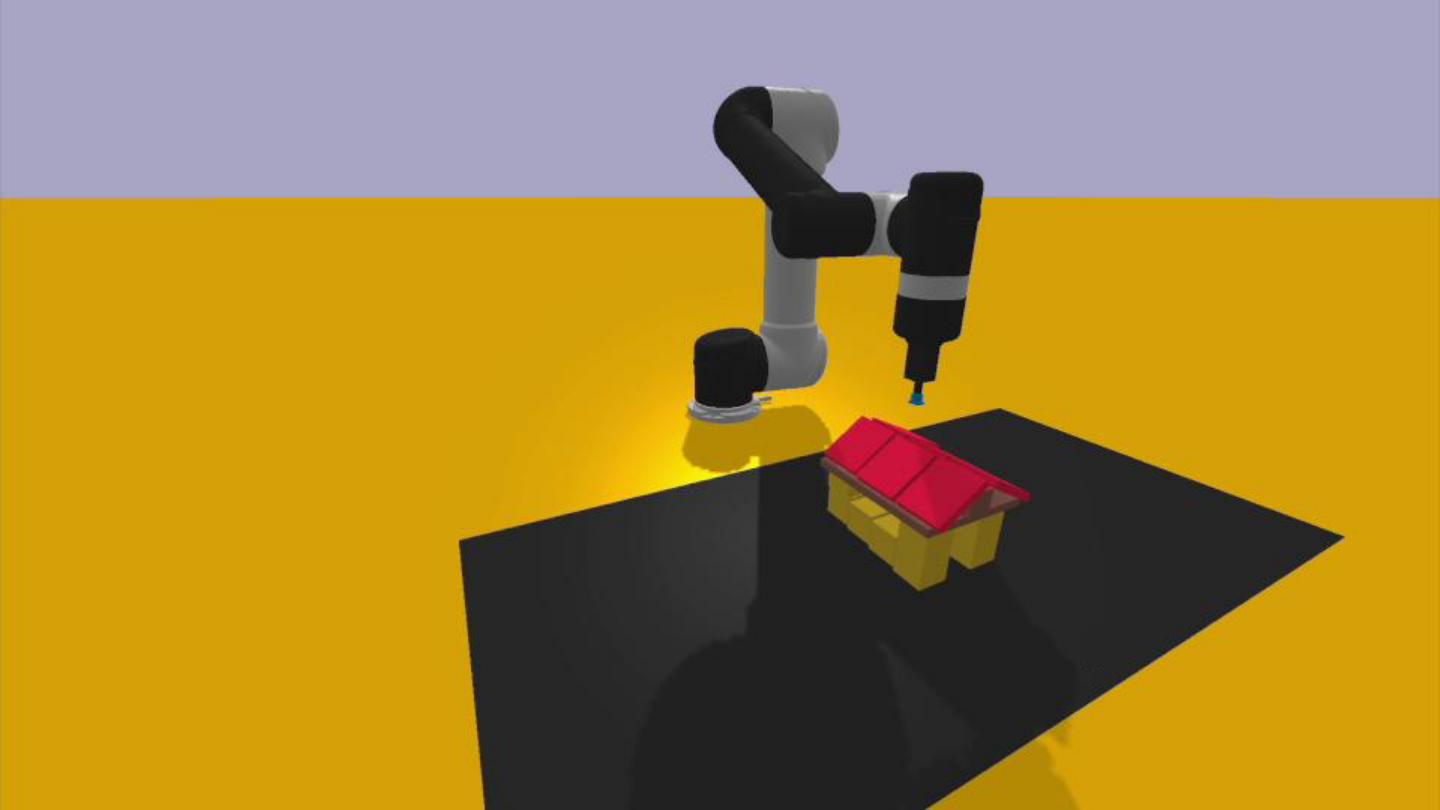}};
        \node[inner sep=0, below](img8) at([yshift=-.5cm]img6.south){\includegraphics[width=6cm]{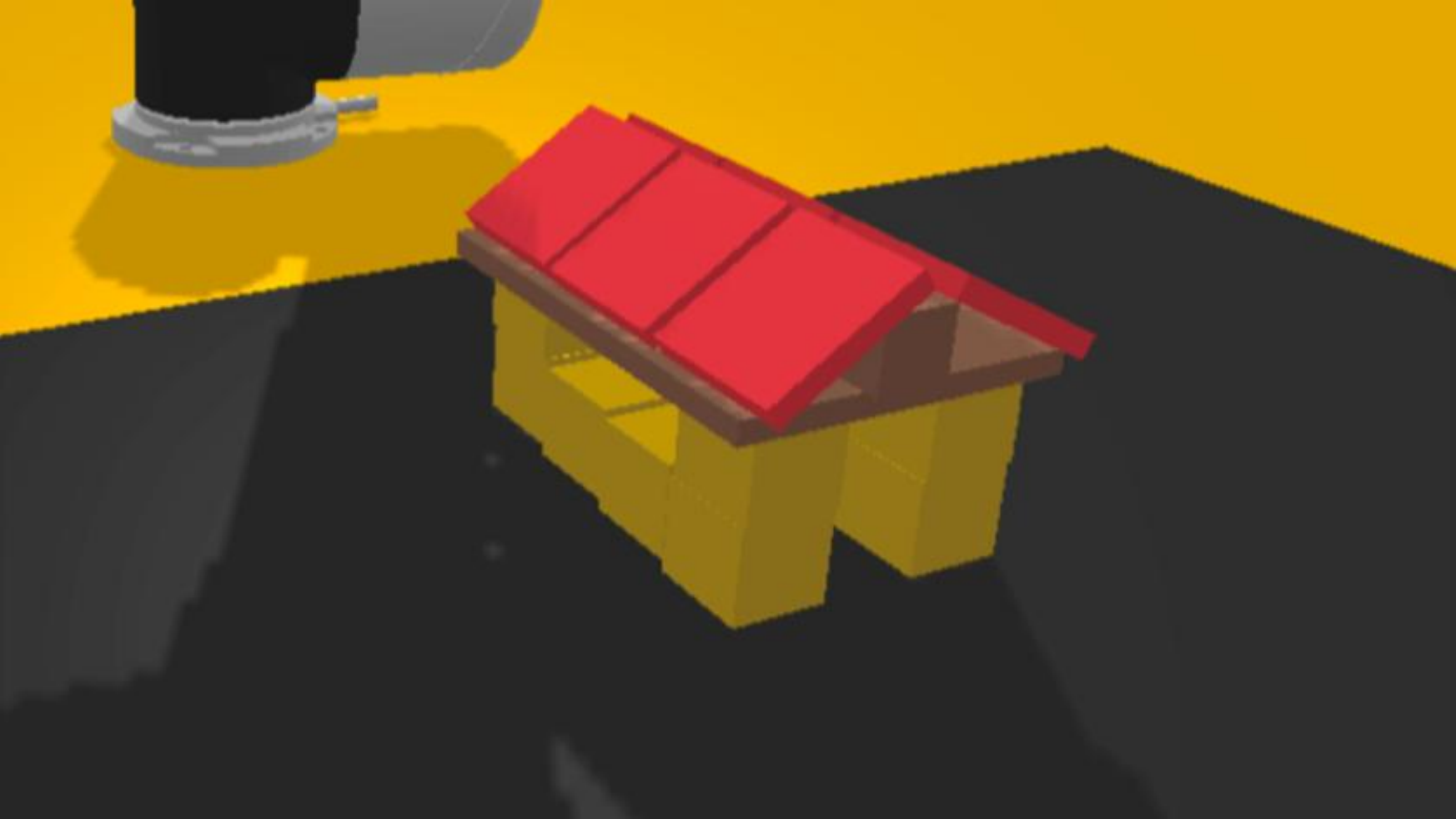}};
        \draw[->, line width=1pt](img1.west) -| (img2.north);
        \draw[->, line width=1pt](img1.south) -- (img3.north);
        \draw[->, line width=1pt](img1.east) -| (img4.north);
        \draw[->, line width=1pt](img2.south) -- (img5.north);
        \draw[->, line width=1pt](img3.south) -- (img6.north);
        \draw[->, line width=1pt](img4.south) -- (img7.north);
        \draw[->, line width=1pt](img5.east) -- (img6.west);
        \draw[->, line width=1pt](img6.east) -- (img7.west);
        \draw[->, line width=1pt](img7.south) |- (img8.east);

        \node[]at(-14cm, 0){\LARGE \textbf{\textsf{a}}};
        \node[below right]at(img1.north west){\LARGE \textcolor{white}{\textbf{\textsf{b}}}};
        \node[below right]at(img2.north west){\LARGE \textcolor{white}{\textbf{\textsf{c}}}};
        \node[below right]at(img3.north west){\LARGE \textcolor{white}{\textbf{\textsf{d}}}};
        \node[below right]at(img4.north west){\LARGE \textcolor{white}{\textbf{\textsf{e}}}};
        \node[below right]at(img5.north west){\LARGE \textcolor{white}{\textbf{\textsf{f}}}};
        \node[below right]at(img6.north west){\LARGE \textcolor{white}{\textbf{\textsf{g}}}};
        \node[below right]at(img7.north west){\LARGE \textcolor{white}{\textbf{\textsf{h}}}};
        \node[below right]at(img8.north west){\LARGE \textcolor{white}{\textbf{\textsf{i}}}};
        
    \end{tikzpicture}
    }
    \vskip -.1in
    \caption{
    Case study of ``build a house''. 
    (a) Skill tree, where green nodes are house-specific skills, yellow nodes are reusable skills learned from prior tasks, and orange nodes are core primitives. 
    (b)--(i) Task snapshots: (b) initial scene, (c)--(e) learned sub-skills for clearing the table, building the base, and constructing the roof, and (f)--(i) their reuse in the final task plan.}
    \vskip -.1in
    \label{fig:skill-tree}
\end{figure*}
To demonstrate how user-guided lifelong learning expands skill capabilities and enables challenging long-horizon tasks, we present the case study ``build a house''. 
As shown in Fig.~\ref{fig:skill-tree}(a), the task required 12 skills developed bottom-up from core primitives (orange), including 7 learned specifically for house building (light green) and others inherited from prior tasks (yellow). 
Starting from the chaotic initial scene (Fig.~\ref{fig:skill-tree}(b)), the task can be decomposed top-down into three subtasks: organizing the scene (Fig.~\ref{fig:skill-tree}(c)), building the base (Fig.~\ref{fig:skill-tree}(d)), and constructing the roof (Fig.~\ref{fig:skill-tree}(e)). 
Unlike LLM-only closed-loop methods that directly generate flat code without awareness of required skills, our framework acquires intermediate skills such as stacking blocks corner-to-corner or aligning them with precise distance and orientation, which can be further extended to build cubes and more complex structures. 
Through lifelong learning, users introduce increasingly complex tasks, and the agent expands its capability by reusing existing skills, either nesting them within new ones or modularizing them to develop extended functionality. 
Both learned skills and examples are stored in an external database, enabling relevant information to be retrieved and appended to the prompt. 
This prevents catastrophic forgetting and gradually builds a complex skill tree for task planning.
This integration of \textbf{top-down planning} and \textbf{bottom-up skill learning} relies on human verifiers’ global awareness of which skills are necessary, a capability current LLM verifiers lack, as they tend to decompose tasks in a single direction and often get stuck in sub-optimal solutions. 
To the best of our knowledge, our framework is the first to successfully accomplish the ``build a house'' task.
%

\subsection{Real-world Performance}
\begin{figure}[ht]
    \centering
    \resizebox{\linewidth}{!}{
    \begin{tikzpicture}
        \node[inner sep=0](img1) at(0, 0){\includegraphics[width=3cm]{imgs/build-house_rw_1.pdf}};
        \node[inner sep=0, right](img2) at([xshift=.1cm]img1.east){\includegraphics[width=3cm]{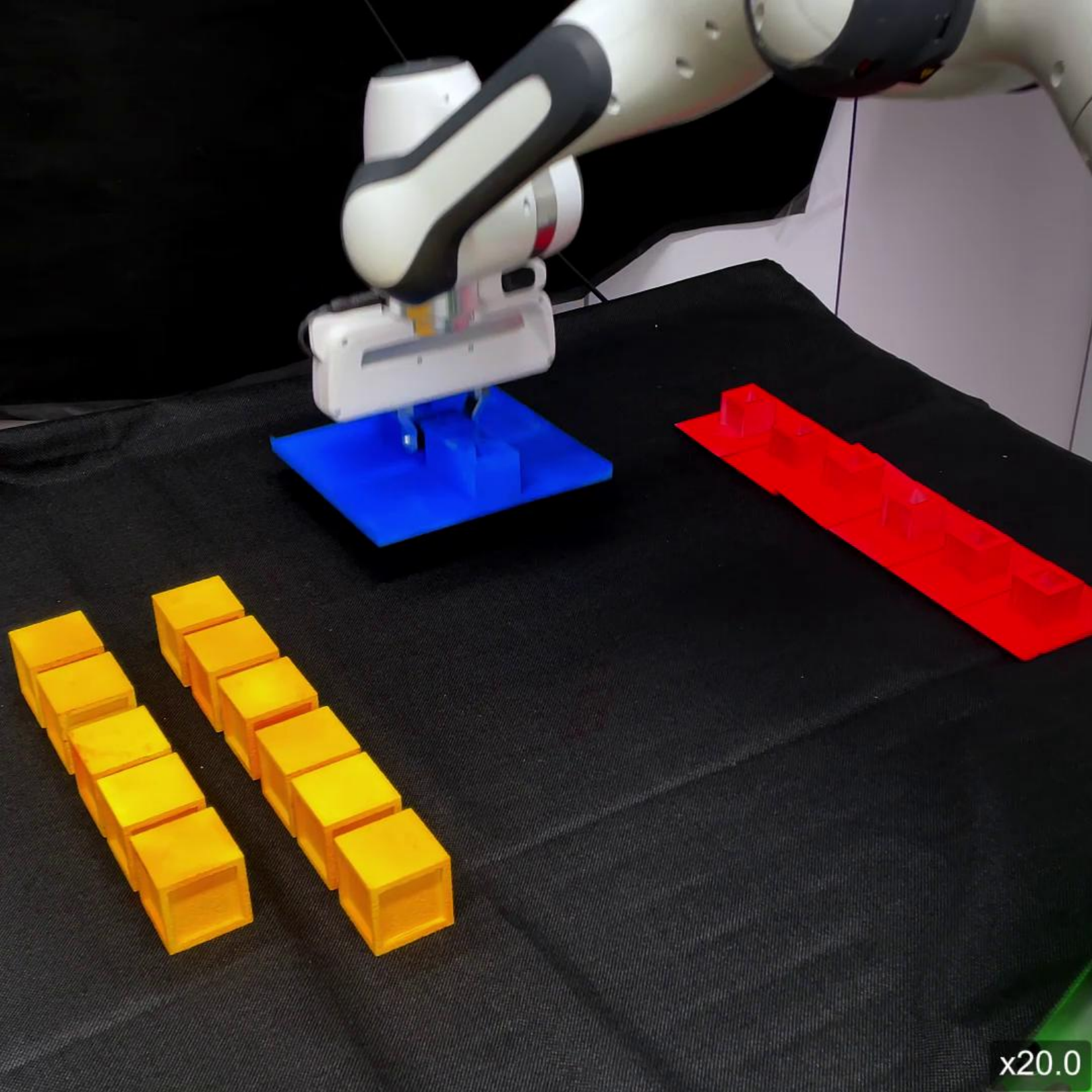}};
        \node[inner sep=0, right](img3) at([xshift=.1cm]img2.east){\includegraphics[width=3cm]{imgs/build-house_rw_6.pdf}};
        \node[inner sep=0, right](img4) at([xshift=.1cm]img3.east){\includegraphics[width=3cm]{imgs/build-house_rw_12.pdf}};
        \node[inner sep=0, right](img5) at([xshift=.1cm]img4.east){\includegraphics[width=3cm]{imgs/build-house-rw.pdf}};
        \node[inner sep=0, below](img6) at([yshift=-.1cm]img1.south){\includegraphics[width=3cm]{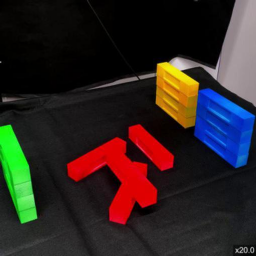}};
        \node[inner sep=0, right](img7) at([xshift=.1cm]img6.east){\includegraphics[width=3cm]{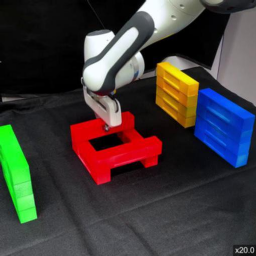}};
        \node[inner sep=0, right](img8) at([xshift=.1cm]img7.east){\includegraphics[width=3cm]{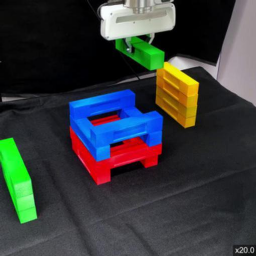}};
        \node[inner sep=0, right](img9) at([xshift=.1cm]img8.east){\includegraphics[width=3cm]{imgs/build-jenga-tower_12.pdf}};
        \node[inner sep=0, right](img10) at([xshift=.1cm]img9.east){\includegraphics[width=3cm]{imgs/build-jenga-tower-rw.pdf}};
        \node[inner sep=0, below](img11) at([yshift=-.1cm]img6.south){\includegraphics[width=3cm]{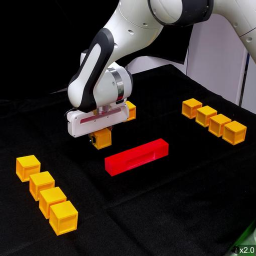}};
        \node[inner sep=0, right](img12) at([xshift=.1cm]img11.east){\includegraphics[width=3cm]{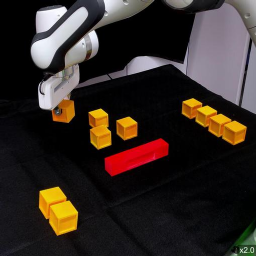}};
        \node[inner sep=0, right](img13) at([xshift=.1cm]img12.east){\includegraphics[width=3cm]{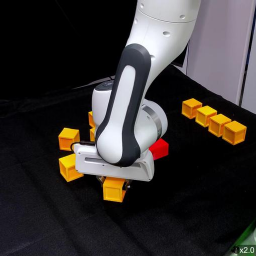}};
        \node[inner sep=0, right](img14) at([xshift=.1cm]img13.east){\includegraphics[width=3cm]{imgs/build-face_12.pdf}};
        \node[inner sep=0, right](img15) at([xshift=.1cm]img14.east){\includegraphics[width=3cm]{imgs/build-face-rw.pdf}};
        \node[inner sep=0, below](img16) at([yshift=-.1cm]img11.south){\includegraphics[width=3cm]{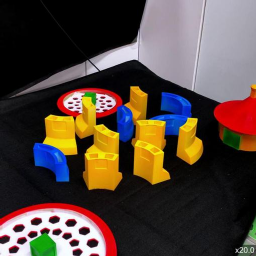}};
        \node[inner sep=0, right](img17) at([xshift=.1cm]img16.east){\includegraphics[width=3cm]{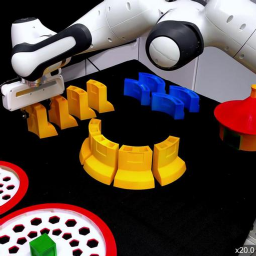}};
        \node[inner sep=0, right](img18) at([xshift=.1cm]img17.east){\includegraphics[width=3cm]{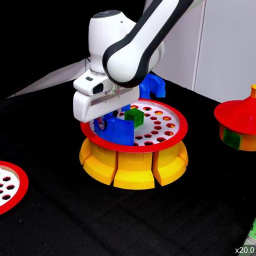}};
        \node[inner sep=0, right](img19) at([xshift=.1cm]img18.east){\includegraphics[width=3cm]{imgs/build-temple_12.pdf}};
        \node[inner sep=0, right](img20) at([xshift=.1cm]img19.east){\includegraphics[width=3cm]{imgs/build-temple-rw.pdf}};
        
        \node[below right]at(img1.north west){\Large \textcolor{white}{\textbf{\textsf{a}}}};
        \node[below right]at(img6.north west){\Large \textcolor{white}{\textbf{\textsf{b}}}};
        \node[below right]at(img11.north west){\Large \textcolor{white}{\textbf{\textsf{c}}}};
        \node[below right]at(img16.north west){\Large \textcolor{white}{\textbf{\textsf{d}}}};

        \node[below right](table)at([xshift=-.3cm]img16.south west){
        \resizebox{15.4cm}{!}{
        \begin{tabular}{c | c c c c }
            \toprule
            \rowcolor{nature_tab_gray1}
             & \textbf{build house} & \textbf{jenga tower} & \textbf{human face} & \textbf{build temple} \\
            \midrule
            \textbf{Success rate} & 7/10  & 7/10 & 9/10 & 6/10 \\
            \bottomrule
        \end{tabular}
        }
        };
    \end{tikzpicture}
    }
    \vskip -.1in
    \caption{Real-world demonstrations of 10 repeat trials. Snapshots of \textbf{(a)} building a house, \textbf{(b)} stacking a Jenga tower, \textbf{(c)} building a human face, and \textbf{(d)} generalization transfer in an unseen task: constructing a temple. 
    Full real-world results see supplementary video.
    }
    \label{fig:real-world-demo}
\end{figure}

The skills learned through our pipeline are embodiment-agnostic and can be deployed on heterogeneous robot arms, enabling smooth transfer to real-world settings. 
We provide APIs for the Franka FR3 with consistent naming and variables to match the simulation environment. 
As shown in Fig.~\ref{fig:real-world-demo}(a)–(d), our framework allows the agent to robustly perform challenging long-horizon real-world tasks, including building a house (0.7 SR), stacking a Jenga tower (0.7 SR), and building a human face pattern (0.9 SR). 
These tasks can last over 12 minutes, highlighting the framework’s effectiveness in real-world deployment. 
We also test generalization to unseen tasks, such as ``construct a temple''. 
Within 5 NoC, the agent reorganizes the scene, builds the temple base, layers, and head by retrieving and combining existing skills, and successfully completes the task with 0.6 SR of 10 trials. 
Other real-world demonstrations include human preference-aligned tower stacking, where our agent stacks blocks with strictly aligned edges to achieve a human-preferred aesthetic shape.
In contrast, baseline methods such as CaP fail to satisfy these requirements.
More details are provided in the supplementary video.

\section{RELATED WORKS}

\textbf{Code generation in embodied AI.}
LLM-based code generation has shown promise by translating human instructions into executable robot-control programs~\cite{zhang2025generative}.
CaP~\cite{liang2023code} enables language models to generate Python policies by composing perception and control primitives.
More recently, CaP-X~\cite{fu2026cap} provides a benchmark and engineering framework for evaluating and improving CaP agents with stronger perception, interaction, and sim-to-real components.
These methods improve training-free robot code generation, but they still mainly rely on task-level program synthesis and do not explicitly preserve human feedback as reusable lifelong skill knowledge.
Recent closed-loop methods further use LLMs or VLMs as task-level evaluators~\cite{zhang2023lohoravens,meng2025data}, but they often struggle with ultra-long-horizon tasks because feedback remains transient and is not grounded in persistent, compositional skills.

\textbf{Robotic manipulation with language feedback and planning.}
Language-conditioned systems often use LLMs for task decomposition and feedback while relying on pretrained policies or planners for execution~\cite{zhang2023lohoravens,huang2023inner}.
TiPToP~\cite{shen2026tiptop} represents a recent modular open-vocabulary planning system that combines foundation-model perception with task-and-motion planning, focusing on training-free language-conditioned manipulation.
Recent works~\cite{liang2024learning,arenas2024prompt} also use human corrections to improve code generation, but their feedback is mainly stored at the prompt or data level, making it difficult to preserve and reuse preferences across long-horizon tasks.
In contrast, LYRA focuses on human-guided lifelong code generation: user feedback is distilled into reusable parameterized skills, stored in external memory, and retrieved for later ultra-long-horizon planning.

\section{CONCLUSION}
We introduced LYRA, a human-guided framework for lifelong skill code generation. It turns user feedback into reusable skills, preserving knowledge for ultra-long-horizon manipulation. Skill inheritance and external memory help align learning with human preferences, and LYRA outperforms prior closed-loop methods across simulation benchmarks and real-world evaluations, including the challenging task of building a house. Our current approach focuses on relatively stable manipulation primitives and standard similarity-based retrieval. Future work will extend it to contact-rich and dynamic tasks and explore multimodal, graph-based, and collaborative retrieval involving humans and LLMs.

\section*{ACKNOWLEDGMENT}
This work was supported by the Beijing Academy of Artificial Intelligence (BAAI), 
the NVIDIA Academic Grant Program, 
the National Key Research and Development Program of China (2025YFE0217000), 
the 2026 Innovation Project (General Program) of the National Key Laboratory for Novel Software Technology under Grant ZZKT2026A36, 
and the Fundamental Research Funds for the Central Universities (KG202514).

\ifCLASSOPTIONcaptionsoff
  \newpage
\fi


\begin{thebibliography}{10}
\providecommand{\url}[1]{#1}
\csname url@samestyle\endcsname
\providecommand{\newblock}{\relax}
\providecommand{\bibinfo}[2]{#2}
\providecommand{\BIBentrySTDinterwordspacing}{\spaceskip=0pt\relax}
\providecommand{\BIBentryALTinterwordstretchfactor}{4}
\providecommand{\BIBentryALTinterwordspacing}{\spaceskip=\fontdimen2\font plus
\BIBentryALTinterwordstretchfactor\fontdimen3\font minus
  \fontdimen4\font\relax}
\providecommand{\BIBforeignlanguage}[2]{{%
\expandafter\ifx\csname l@#1\endcsname\relax
\typeout{** WARNING: IEEEtran.bst: No hyphenation pattern has been}%
\typeout{** loaded for the language `#1'. Using the pattern for}%
\typeout{** the default language instead.}%
\else
\language=\csname l@#1\endcsname
\fi
#2}}
\providecommand{\BIBdecl}{\relax}
\BIBdecl

\bibitem{yao2023bridging}
X.~Yao, H.~Zhou, O.~Mees, Y.~Meng, T.~Xiao, Y.~Bisk, J.~Oh, E.~Johns,
  M.~Shridhar, D.~Shah \emph{et~al.}, ``Bridging language and action: A survey
  of language-conditioned robot manipulation,'' \emph{The International Journal
  of Robotics Research}, p. 02783649261468360, 2026.

\bibitem{meng2025preserving}
Y.~Meng, Z.~Bing, X.~Yao, K.~Chen, K.~Huang, Y.~Gao, F.~Sun, and A.~Knoll,
  ``Preserving and combining knowledge in robotic lifelong reinforcement
  learning,'' \emph{Nature Machine Intelligence}, pp. 1--14, 2025.

\bibitem{meng2025pretrained}
Y.~Meng, X.~Yao, K.~Chen, Y.~Wu, L.~Zhang, Z.~Bing, and A.~Knoll, ``Pretrained
  bayesian non-parametric knowledge prior in robotic long-horizon reinforcement
  learning,'' \emph{arXiv preprint arXiv:2503.21975}, 2025.

\bibitem{zhang2023lohoravens}
S.~Zhang, P.~Wicke, L.~K. {\c{S}}enel, L.~Figueredo, A.~Naceri, S.~Haddadin,
  B.~Plank, and H.~Sch{\"u}tze, ``Lohoravens: A long-horizon
  language-conditioned benchmark for robotic tabletop manipulation,''
  \emph{arXiv preprint arXiv:2310.12020}, 2023.

\bibitem{huang2023inner}
W.~Huang, F.~Xia, T.~Xiao, H.~Chan, J.~Liang, P.~Florence, A.~Zeng, J.~Tompson,
  I.~Mordatch, Y.~Chebotar \emph{et~al.}, ``Inner monologue: Embodied reasoning
  through planning with language models,'' in \emph{Conference on Robot
  Learning}.\hskip 1em plus 0.5em minus 0.4em\relax PMLR, 2023, pp. 1769--1782.

\bibitem{black2024pi_0}
K.~Black, N.~Brown, D.~Driess, A.~Esmail, M.~Equi, C.~Finn, N.~Fusai, L.~Groom,
  K.~Hausman, B.~Ichter \emph{et~al.}, ``$\pi_0$: A vision-language-action flow
  model for general robot control,'' \emph{arXiv preprint arXiv:2410.24164},
  2024.

\bibitem{zhang2025generative}
K.~Zhang, P.~Yun, J.~Cen, J.~Cai, D.~Zhu, H.~Yuan, C.~Zhao, T.~Feng, M.~Y.
  Wang, Q.~Chen \emph{et~al.}, ``Generative artificial intelligence in robotic
  manipulation: A survey,'' \emph{CoRR}, 2025.

\bibitem{liang2023code}
J.~Liang, W.~Huang, F.~Xia, P.~Xu, K.~Hausman, B.~Ichter, P.~Florence, and
  A.~Zeng, ``Code as policies: Language model programs for embodied control,''
  in \emph{2023 IEEE International Conference on Robotics and Automation
  (ICRA)}.\hskip 1em plus 0.5em minus 0.4em\relax IEEE, 2023, pp. 9493--9500.

\bibitem{fu2026cap}
M.~Fu, J.~Yu, K.~El-Refai, E.~Kou, H.~Xue, H.~Huang, W.~Xiao, G.~Wang, F.-F.
  Li, G.~Shi \emph{et~al.}, ``Cap-x: A framework for benchmarking and improving
  coding agents for robot manipulation,'' \emph{arXiv preprint
  arXiv:2603.22435}, 2026.

\bibitem{liang2024learning}
J.~Liang, F.~Xia, W.~Yu, A.~Zeng, M.~G. Arenas, M.~Attarian, M.~Bauz{\'a},
  M.~Bennice, A.~Bewley, A.~Dostmohamed \emph{et~al.}, ``Learning to learn
  faster from human feedback with language model predictive control,''
  \emph{CoRR}, 2024.

\bibitem{zhi2025closed}
P.~Zhi, Z.~Zhang, Y.~Zhao, M.~Han, Z.~Zhang, Z.~Li, Z.~Jiao, B.~Jia, and
  S.~Huang, ``Closed-loop open-vocabulary mobile manipulation with gpt-4v,'' in
  \emph{2025 IEEE International Conference on Robotics and Automation
  (ICRA)}.\hskip 1em plus 0.5em minus 0.4em\relax IEEE, 2025, pp. 4761--4767.

\bibitem{meng2025data}
Y.~Meng, X.~Yao, H.~Ye, Y.~Zhou, S.~Zhang, Z.~Bing, and A.~Knoll,
  ``Data-agnostic robotic long-horizon manipulation with vision-language-guided
  closed-loop feedback,'' \emph{arXiv preprint arXiv:2503.21969}, 2025.

\bibitem{arenas2024prompt}
M.~G. Arenas, T.~Xiao, S.~Singh, V.~Jain, A.~Ren, Q.~Vuong, J.~Varley,
  A.~Herzog, I.~Leal, S.~Kirmani \emph{et~al.}, ``How to prompt your robot: A
  promptbook for manipulation skills with code as policies,'' in \emph{2024
  IEEE International Conference on Robotics and Automation (ICRA)}.\hskip 1em
  plus 0.5em minus 0.4em\relax IEEE, 2024, pp. 4340--4348.

\bibitem{zhao2024expel}
A.~Zhao, D.~Huang, Q.~Xu, M.~Lin, Y.-J. Liu, and G.~Huang, ``Expel: Llm agents
  are experiential learners,'' in \emph{Proceedings of the AAAI Conference on
  Artificial Intelligence}, vol.~38, 2024, pp. 19\,632--19\,642.

\bibitem{zha2024distilling}
L.~Zha, Y.~Cui, L.-H. Lin, M.~Kwon, M.~G. Arenas, A.~Zeng, F.~Xia, and
  D.~Sadigh, ``Distilling and retrieving generalizable knowledge for robot
  manipulation via language corrections,'' in \emph{2024 IEEE international
  conference on robotics and automation (ICRA)}.\hskip 1em plus 0.5em minus
  0.4em\relax IEEE, 2024, pp. 15\,172--15\,179.

\bibitem{wang2023voyager}
G.~Wang, Y.~Xie, Y.~Jiang, A.~Mandlekar, C.~Xiao, Y.~Zhu, L.~Fan, and
  A.~Anandkumar, ``Voyager: An open-ended embodied agent with large language
  models,'' \emph{arXiv preprint arXiv:2305.16291}, 2023.

\bibitem{mitchell2006discipline}
T.~M. Mitchell, \emph{The discipline of machine learning}.\hskip 1em plus 0.5em
  minus 0.4em\relax Carnegie Mellon University, School of Computer Science,
  Machine Learning~…, 2006, vol.~9.

\bibitem{neelakantan2022text}
A.~Neelakantan, T.~Xu, R.~Puri, A.~Radford, J.~M. Han, J.~Tworek, Q.~Yuan,
  N.~Tezak, J.~W. Kim, C.~Hallacy \emph{et~al.}, ``Text and code embeddings by
  contrastive pre-training,'' \emph{arXiv preprint arXiv:2201.10005}, 2022.

\bibitem{kim2024openvla}
M.~J. Kim, K.~Pertsch, S.~Karamcheti, T.~Xiao, A.~Balakrishna, S.~Nair,
  R.~Rafailov, E.~P. Foster, P.~R. Sanketi, Q.~Vuong \emph{et~al.}, ``Openvla:
  An open-source vision-language-action model,'' in \emph{8th Annual Conference
  on Robot Learning}, 2024.

\bibitem{zeng2021transporter}
A.~Zeng, P.~Florence, J.~Tompson, S.~Welker, J.~Chien, M.~Attarian,
  T.~Armstrong, I.~Krasin, D.~Duong, V.~Sindhwani \emph{et~al.}, ``Transporter
  networks: Rearranging the visual world for robotic manipulation,'' in
  \emph{Conference on Robot Learning}.\hskip 1em plus 0.5em minus 0.4em\relax
  PMLR, 2021, pp. 726--747.

\bibitem{gupta2020relay}
A.~Gupta, V.~Kumar, C.~Lynch, S.~Levine, and K.~Hausman, ``Relay policy
  learning: Solving long-horizon tasks via imitation and reinforcement
  learning,'' in \emph{Conference on Robot Learning}.\hskip 1em plus 0.5em
  minus 0.4em\relax PMLR, 2020, pp. 1025--1037.

\bibitem{yu2020meta}
T.~Yu, D.~Quillen, Z.~He, R.~Julian, K.~Hausman, C.~Finn, and S.~Levine,
  ``Meta-world: A benchmark and evaluation for multi-task and meta
  reinforcement learning,'' in \emph{Conference on robot learning}.\hskip 1em
  plus 0.5em minus 0.4em\relax PMLR, 2020, pp. 1094--1100.

\bibitem{NEURIPS2023_8c3c6668}
B.~Liu, Y.~Zhu, C.~Gao, Y.~Feng, Q.~Liu, Y.~Zhu, and P.~Stone, ``Libero:
  Benchmarking knowledge transfer for lifelong robot learning,'' in
  \emph{Advances in Neural Information Processing Systems}, A.~Oh, T.~Naumann,
  A.~Globerson, K.~Saenko, M.~Hardt, and S.~Levine, Eds., vol.~36.\hskip 1em
  plus 0.5em minus 0.4em\relax Curran Associates, Inc., 2023, pp.
  44\,776--44\,791.

\bibitem{ren2024grounded}
T.~Ren, S.~Liu, A.~Zeng, J.~Lin, K.~Li, H.~Cao, J.~Chen, X.~Huang, Y.~Chen,
  F.~Yan \emph{et~al.}, ``Grounded sam: Assembling open-world models for
  diverse visual tasks,'' \emph{arXiv preprint arXiv:2401.14159}, 2024.

\bibitem{shen2026tiptop}
W.~Shen, N.~Kumar, S.~Chintalapudi, J.~Wang, C.~Watson, E.~Hu, J.~Cao,
  D.~Jayaraman, L.~P. Kaelbling, and T.~Lozano-P{\'e}rez, ``Tiptop: A modular
  open-vocabulary planning system for robotic manipulation,'' \emph{arXiv
  preprint arXiv:2603.09971}, 2026.

\end{thebibliography}


\end{document}